\documentclass{elsarticle}

\date{15 October 2019}
\usepackage{hyperref}

\usepackage{graphicx}
\usepackage{color}
\usepackage{times}
\usepackage{xcolor}
\usepackage{soul}
\usepackage[utf8]{inputenc}
\usepackage[small]{caption}
\usepackage{subcaption}

\usepackage{booktabs}
\usepackage{bm}
\usepackage{mathtools}
\usepackage[export]{adjustbox}
\usepackage{amsmath,amssymb}
\usepackage[Algorithm]{algorithm}
\usepackage[noend]{algpseudocode}
\usepackage{tabularx}
\usepackage{url}
\usepackage{todonotes}

\def\SynTeam{SynTeam}
\def\cplex{STCPSolver}
\newtheorem{mydef}{Definition}

\journal{Knowledge-Based Systems}

\begin{document}

\begin{frontmatter}

\title{Synergistic Team Composition: A Computational Approach to Foster Diversity in Teams}

\author[iiia,ntu]{Ewa Andrejczuk}
\ead{ewaa@ntu.edu.sg}
\author[iiia]{Filippo Bistaffa}
\ead{filippo.bistaffa@iiia.csic.es}
\author[iiia]{Christian Blum\corref{cor1}}
\ead{christian.blum@iiia.csic.es}
\author[iiia]{Juan A. Rodriguez-Aguilar}
\ead{jar@iiia.csic.es}
\author[iiia]{Carles Sierra}
\ead{sierra@iiia.csic.es}
\cortext[cor1]{Corresponding author}

\address[iiia]{Artificial Intelligence Research Institute (IIIA-CSIC), Campus of the UAB, Bellaterra, Catalonia, Spain}
\address[ntu]{ST Engineering - NTU Corporate Laboratory, Nanyang Technological University, Singapore}



\begin{abstract}
Co-operative learning in heterogeneous teams refers to learning methods in which teams are organised both to accomplish academic tasks and for individuals to gain knowledge. Competencies, personality and the gender of team members are key factors that influence team performance. Here, we introduce a team composition problem, the so-called \emph{synergistic team composition problem} (STCP), which incorporates such key factors when arranging teams. Thus, the goal of the STCP is to partition a set of individuals into a set of \emph{synergistic teams}: teams that are diverse in personality and gender and whose members cover all required competencies to complete a task. Furthermore, the STCP requires that all teams are balanced in that they are expected to exhibit similar performances when completing the task. 
We propose two efficient algorithms to solve the STCP. Our first algorithm is based on a linear programming formulation and is appropriate to solve small instances of the problem. Our second algorithm is an anytime heuristic that is effective for large instances of the STCP. 
Finally, we thoroughly study the computational properties of both algorithms in an educational context when grouping students in a classroom into teams using actual-world data.
%
%
\end{abstract}

\begin{keyword}
\texttt{team composition}\sep \texttt{exact algorithms} \sep \texttt{heuristic algorithms} \sep \texttt{optimisation} \sep \texttt{coalition formation}
\end{keyword}

\end{frontmatter}



\section{Introduction}

Active learning refers to a broad range of teaching techniques that engage students to participate in all learning activities in the classes. Typically, active learning strategies involve a substantial amount of students working together within teams. Research shows that students learn better when using active learning compared to the traditional schooling methods \cite{vosniadou2003children}. They do not only acquire and retain the information better but also are more content with their classes \cite{Barkley}. 


Nevertheless, not all teams facilitate learning. For team-based learning to be effective, every team composed in the classroom needs to be heterogeneous, i.e. diverse in individuals' characteristics. \textcolor{black}{Furthermore, having some significantly weaker teams and some significantly stronger teams is undesirable. Hence, the distribution of teams in a classroom must be \emph{balanced} in the sense that all teams are more or less equally strong.}

Even though much research in the industrial, organisational, and educational psychology fields investigated what are the predictors of team success, to the best of our knowledge, there are no computational models to build teams for a given task that are broadly used in the classrooms. Frequently studied individual characteristics that influence team performance are competencies, personality traits, and gender~\cite{Arnold,Mount,West,White}. \cite{Arnold,White} show a positive correlation between certain personality traits and team composition. \cite{Mount,West} show that in order to increase team performance, team members should be heterogeneous in their individual characteristics. 

Some of those characteristics were also acknowledged by multiagent systems (MAS) research. 
The most studied characteristic in MAS research are competencies  \cite{alberola2016artificial,Chen2015,FarhangianPPS15, Okimoto,Rangapuram2015}. However, in these works agents' competencies are generaly represented as True/False characteristics, that is, an agent has or does not have a required competence. This is an oversimplified approach to model agents' competencies since it disregards any competence grade. In reality, competencies are non-binary because individuals are characterized by different grades of competencies. Unfortunately, MAS research has typically ignored significant psychology findings (with the exception of recent, preliminary works such as \cite{alberola2016artificial} and~\cite{FarhangianPPS15}). 

To the best of our knowledge, neither the current MAS literature nor the current psychology literature has considered team composition based on competencies, personality and gender of individuals at the same time.

In this paper, we focus on the following team composition problem that is commonly encountered in education. 
We consider a \emph{complex task that needs to be performed by multiple student teams of even size}~\cite{acuna}. The task requires each team to have at least one student with a minimum level of competence for each competence from a given set of competencies. There is a pool of students with varying competencies, genders, and personalities. The objective is to partition students into teams so that each team is even in size and balanced in competencies as well as personalities, and gender.  We term those teams as \emph{synergistic teams}.  

In this context, the paper makes the following contributions:

\begin{itemize}
    \item We identify and formally define a new type of real-life problem, the so-called synergistic team composition problem (STCP). The goal of the STCP is to partition a set of individuals into a set of \emph{synergistic teams}: teams that are diverse in personality and gender and whose members cover all required competencies to complete a task. Furthermore, the STCP requires that all teams are balanced in that they are expected to exhibit similar performances when completing the task.
    
    \item We introduce two different algorithms to tackle the STCP: (i) \cplex, an algorithm that employs a reformulation of the problem which is then solved to optimality by an off-the-shelf integer linear programming (ILP) solver; and (ii) \SynTeam, an anytime heuristic that can produce solutions of high quality within a limited computation time. 
    
    \item We perform an exhaustive computational comparison of \cplex\ and \SynTeam\ over realistic settings in education, considering actual-world data. Overall, our analysis indicates that \cplex\ is efficient for rather small problem instances, whereas \SynTeam\ is able to cope with larger problem instances. First, we notice that the runtime of the optimal algorithm greatly increases with a growing team size and a growing number of students, which causes the algorithm not to be applicable to larger instances of the problem. 
    This is not the case for \SynTeam, which is capable of composing teams for larger problem instances while providing good quality approximate solutions (whose values range, in the worst case, between 75\% and 95\% of the value of an optimal solution). Second, we compare the anytime performance of both algorithms. We observe that \SynTeam\ outperforms \cplex\ for large team sizes (beyond 3), whereas the opposite occurs for small team sizes.
    We also compared our optimal approach (i.e., STCPSolver) to ODP-IP~\cite{michalak2016hybrid}, the state-of-the-art algorithm to solve the \emph{coalition structure generation} problem. Results show that STCPSolver outperforms ODP-IP both in terms of runtime (since ODP-IP cannot exploit the presence of cardinality constraints to reduce the space of feasible solutions) and scalability (due to ODP-IP's exponential memory requirements).
\end{itemize}


\noindent
\textbf{Outline.} The remainder of this paper is structured as follows. Section~\ref{sec:related}  provides an overview of the related work. 
Section~\ref{sec:model} introduces the basic definitions used in this paper.  
Section~\ref{sec:problem} introduces the key notions used to measure the synergistic value of a team and formally defines the synergistic team composition problem. Sections~\ref{sec:optimal} and \ref{sec:approxAlg} describe \cplex\ and \SynTeam, the two algorithms that we introduce in this work. Then, Section~\ref{sec:results} discusses our empirical comparison of the proposed algorithms over synthetically-generated instances of the STCP. Section~\ref{sec:eduteams} briefly introduces a web application that is freely available and offers team composition as its main functionality. Finally, Section~\ref{sec:discuss} discusses both the conclusions and directions for future research.

\section{Related work} \label{sec:related}

In this section we review relevant related work. 
Sections \ref{relatedEd} and \ref{relatedOP} go through related work in the education literature and the organisational psychology literature respectively. Sections \ref{relatedMAS} and \ref{sec:cfrel} revise related work in the computer science literature; section \ref{relatedMAS} revises the multiagent systems literature, whereas section \ref{sec:cfrel} discusses relevant work in the coalition formation literature.

\subsection{Relation with the education literature} \label{relatedEd}

There are many works that advise on how to \emph{handcraft} heterogeneous teams with the purpose of increasing team-based learning and improving team performance, for instance, \cite{Wilde2013}  or \cite{michaelsen2008team}. 

\cite{Wilde2013} offers a \emph{manual} method to divide a classroom based on students' personalities and genders. In this paper, we extend the method in \cite{Wilde2013} by adding competencies and propose an algorithm to compose teams in an \emph{automatic} way. 

\cite{michaelsen2008team} advises beginning a team composition process by simply asking questions to a group of students. These questions are used to gather information about those competencies that are important for the successful completion of a given task.  Students respond to each question either orally or with a show of hands. Then, students are lined up based on the number of required competencies that they have, derived from the answers to the questions. Ties are broken randomly. Finally, teams are built by asking students to count off down the line. For instance, if teams of five students are required, the count is as follows: $1, 2, 3, 4, 5, 1, 2, \ldots $. Hereby, the number associated to a student indicates the team to which he/she is assigned.

Some authors have tried to automatise the team composition process. That is, they have aimed at composing a set of teams so that all teams are as similar as possible with regard to the mean values of multiple attributes \cite{cutshall2007indiana, dear2000applying,desrosiers2005design,layton2010design,rubin2015forming}. As opposed to our approach, none of these works imposes heterogeneity in a direct way when composing teams. They are rather limited to studying a set of fixed constraints (such as avoiding clustering particular majors, ensuring that no international student and no female are isolated on a team, etc). Additionally, compared to our approach where we compose teams for particular tasks, they do not explicitly consider the notion of the task when composing balanced teams.
\textcolor{black}{This is also the case in the work by Agrawal et al. \cite{Agrawal:2014:grouping}, though the authors diverge from the above-mentioned approaches to team composition. Their team composition approach focuses on grouping students so that, in the end, the value gained by less capable students through collaboration is maximised. Despite the novelty of their team composition (grouping) problem, Agrawal et al. only consider students capabilities (disregarding the findings of the organisational psychology literature about other individual attributes, i.e. personality and gender, that we include), consider that students count on abilities for a single competence (instead of multiple competencies as we do), and are not concerned about yielding balanced team compositions, which is our main goal.} 

To the best of our knowledge, the only available web tool supporting team composition is the Comprehensive Assessment of Team Member Effectiveness (CATME)\footnote{\url{http://www.catme.org}} that composes teams based on individual students' responses to an online survey. Teachers define student surveys by selecting the desired students' characteristics from a given inventory \cite{layton2010design}.  The application calculates a ``question score'' for each characteristic that informs how well each team's distribution of that characteristic satisfies the teacher’s aims. The application also measures a global ``compliance score'' for each composed team characterizing how well the team satisfies the teacher's wishes. The higher these values the better the team.
Their team composition algorithm starts by randomly distributing students across teams of a pre-specified size. Next, the algorithm calculates both question and compliance scores. Then, it iteratively changes the teams with the purpose of maximising the minimum compliance score of all teams. This work is similar to our approach, however, there are also substantial differences. In addition to the differences discussed above, the authors do not analyze their solutions' quality.  They assume that the groupings produced by their algorithm are near-optimal. The analysis performed by \cite{henry2013forming} shows, however, that it is implausible the CATME method achieves near-optimal results.

\subsection{Relation with the organisational psychology literature} \label{relatedOP}

As far as we are concerned, there are no methods in the organisational psychology literature that would provide a complete guideline on how to compose teams. Instead, the researchers in this field study how individual characteristics influence team performance. 

The most studied individual characteristic that is associated with team performance is cognitive ability. \cite{devine2001smarter} define it as the ``capacity to understand complex ideas, learn from experience, reason, solve problems, and adapt'' \cite[p.507]{devine2001smarter}. It is a very wide concept that---in addition to competencies, broadly used in multiagent systems research---covers many other characteristics such as experience, gender or even age. \cite{Bell2007} and \cite{devine2001smarter} discovered the positive correlation between team performance and the average team values of cognitive ability. \cite{devine2001smarter} also showed that the variance of team members' cognitive ability was not a good predictor of team performance. Additionally, these authors observed that the average value is two times more informative for the prediction of team performance than the lowest and the highest member scores. \cite{woolley2015collective} suggested the existence of collective intelligence in teams that can predict team performance. This collective intelligence is not strongly correlated with the maximum or average intelligence of team members. Instead, it is positively correlated with increasing equality in conversational turn-taking, the mean social sensitivity of group members, and gender balance \cite{woolley2010evidence}. 

The organisational psychology literature, in addition to cognitive ability, has examined the impact of personality traits on team performance~\cite{west2012, White}. The most popular questionnaires to determine personality include: (1) the Five Factor Model (known as ``Big Five'' or FFM), which uses five traits to define individual personality~\cite{Costa}; (2) Belbin theory~\cite{belbin}, which provides a theory of nine different personality role types; and 
(3) the Myers-Briggs Type Indicator (MBTI) questionnaire that uses four traits to specify psychological preferences concerning the way people perceive the world and make decisions~\cite{Myers}. 

Concerning FFM, \cite{Bell2007} found that for each personality trait examined individually (i.e. extraversion, agreeableness, conscientiousness, emotional stability, openness to experience), team means were associated with team performance. The study in~\cite{prewett2009team} confirmed these findings for all traits except for the Openness to Experience trait which was not considered. However, the sizes of studied samples were small and it is unclear whether these findings are statistically significant \cite{peeters2006personality}.
\cite{mohammed2003personality} reported contradictory findings after studying student teams. Each team was asked to improve processes based on problems encountered in organisations. The researchers measured team orientation, extraversion, agreeableness, conscientiousness, and emotional stability of each team using the team average and team variability. Interestingly, they did not find any meaningful connection between team performance and any of these personality traits (when examined individually).

According to Belbin, there are nine team roles that should be covered in every team~\cite{belbin}. These roles are: completer--finisher, coordinator, implementer, monitor evaluator, plant, resource investigator, shaper,  specialist and teamworker.
Although some studies with very limited sample sizes (such as 10 teams in~\cite{senior1997team}) reported support for the theory, studies based on a larger number of samples did not find the relation between the Belbin roles and team performance \cite{partington1999belbin,batenburg2013belbin,van2008belbin}.

Finally, the MBTI has four binary dimensions, that is:
intuition vs sensing (N--S), thinking vs feeling (T--F), extraversion vs introversion (E--I), and perceiving vs judging (P--J). Within this questionnaire, every individual can be categorised into one of the sixteen possible four-letter combinations, where each letter represents one personality dimension. This approach is easy to interpret by non-psychologists. Reliance on dichotomous preference scores rather than on continuous scores, however, excessively restricts the level of statistical analysis~\cite{devito}.

\subsection{Relation with the multiagent systems literature} \label{relatedMAS}
\label{subsec:mas}

To our knowledge, the only computational model in the context of team composition that takes both personality and competencies into account was presented in~\cite{farhangian2015agent}. In particular, the influence of personality on different strategies for allocating tasks is studied in this paper. However, there are substantial differences with our work. Firstly, instead of proposing an algorithm for the composition of teams both based on personality and competence, they only describe a model to evaluate teams. Secondly, they give no importance to gender balance. And finally, they do not evaluate their algorithm with real data (only via agent-based simulation).

We separate the remaining literature that is relevant to this article into the following two categories: works that deal with agent competencies (individual and social capabilities of agents), and works that consider agent personality (individual behaviour models).

\textbf{Competencies.}
Various previous works have focused on the competency dimension. However, in contrast to our work, in which competencies are graded, the majority of works assume agents to have multiple binary skills (either an agent has a required skill or not). In \cite{Crawford} and \cite{Okimoto}, for instance, one k-robust team is composed for a single task, based on the agents' capabilities. Hereby, a team is called $k$-robust if by removing any $k$ members from the team, the completion of the task is not compromised. In \cite{Anagnostopoulos12onlineteam}, each task requires a specific set of competencies. Moreover, tasks arrive sequentially over time. The team composition algorithm, whose focus is on balancing the workload of the agents across teams, builds teams based on competencies and communication cost.

\textbf{Personality.}
There are, to our knowledge, two works in the literature that consider agents' personality to compose teams, namely~\cite{alberola2016artificial} and~\cite{FarhangianPPS15}. In \cite{alberola2016artificial}, Belbin's theory is used to obtain human predominant \emph{roles} (see Section~\ref{relatedOP}). As discussed in subsection~\ref{relatedOP}, these roles do not tend to be related to team performance. Additionally, gender is not considered for the composition of heterogeneous teams.

In~\cite{FarhangianPPS15}, Farhangian et al.~make use of the classical MBTI personality test (see Section~\ref{relatedOP}). Their aim is to build the best possible team around a selected leader. In other words, they compose the \emph{best} possible team for a particular task. However, gender balance is again not considered. Finally, although real data was considered in~\cite{FarhangianPPS15}, the resulting teams' performance was not validated. Instead, Bayesian theory was used to predict the success probability in a variety of team composition conditions.

\subsection{Relation with the coalition formation literature}\label{sec:cfrel}

The STCP can be seen as a \emph{coalition structure generation} (CSG) problem~\cite{michalak2016hybrid} over the entire set of students with a characteristic function that assigns a synergistic value to every \emph{feasible} coalition (i.e., with the desired number of students), and $-\infty$ to every \emph{unfeasible} coalition (since the characteristic function has to be defined for every possible subset of agents in the standard definition of CSG).
Solving the STCP requires to compute the coalition structure (team partition) with the largest total value, i.e., the optimal solution to the CSG problem.
In principle, state-of-the-art CSG approaches such as ODP-IP~\cite{michalak2016hybrid} could be used to solve the STCP problem. Unfortunately, these approaches are not able to exploit the presence of cardinality constraints to reduce the space of feasible solution, and hence, are limited to problem instances of up to 25 agents, due to their exponential memory requirements, as shown by our experiments in Section~\ref{ssec:CompResults}.

On the other hand, given a STCP we can also define a constrained coalition formation (CCF) \cite{Rahwan:2011:CCF} game $\mathcal{G} = \langle A,\mathcal{P}_m(A),s\rangle$, where $\mathcal{P}_m(A)$ is the set of feasible coalition structures.
More precisely, the STCP poses a specific type of CCF game, namely, a \emph{basic} CCF game \cite{Rahwan:2011:CCF}. 
Intuitively, basic CCF game express constraints in the form of: (1) allowed sizes of coalitions that can be formed; and (2) subsets of agents whose presence in any coalition is permitted or not.
On the one hand, a STCP naturally defines constraints on the size of coalitions.
On the other hand, expressing a STCP as a CCF problem would require to define one positive constraint per feasible team, 
while the set of negative constraints would be empty.
As a consequence, the number of positive constraints quickly becomes very large (i.e., $> 3000$ in our case), hence making the use of the approach by Rahwan \emph{et al.}~\cite{Rahwan:2011:CCF} impossible.
 
 \section{Team Composition Model}\label{sec:model}

There are three diversity dimensions of students used in our model: gender, personality, and competencies. 
We measure personality using the theory of personality called Post-Jungian \cite{Wilde2013} which is a reduced variant of the Myers-Briggs Type Indicator (MBTI) \cite{Myers}. The numbers are obtained from the answers of a short questionnaire of 20 quick questions (much shorter than the common 93 questions of the Boolean MBTI). This is very efficient in terms of time and effort for both teachers and students, as completing the test takes only a few minutes (see \cite[p.21]{Wilde2013} for details). Douglass J. Wilde claimed that this numerical method is a coherent extension of the psychological dimensions of MBTI \cite{Wilde2009}. The test is based on the personality model proposed by C. G. Jung \cite{PT} containing two sets of functions and attitudes:  
\begin{enumerate}
\item Sensing -- Intuition (SN),
\item Thinking -- Feeling (TF)
\item Extroversion -- Introversion (EI),
\item Perception -- Judgment (PJ).
\end{enumerate}
The numerical values along each dimension (SN, TF, EI, PJ) are the result of combining the answers to the questionnaire mentioned above where each question can be answered by selecting one out of five possible answers; Each possible answer has a value (in a scale from -1 to +1). Each personality trait is assessed by five questions. The values of the answers are added up and divided by 5 (the number of questions) to give the final value along each personality dimension. This method seems promising as---within one decade---Prof. Wilde multiplied the number of teams of Stanford that were awarded prizes by the Lincoln Foundation \cite{Wilde2009} by three. Accordingly, the definition of a personality profile in our context is as follows.

\begin{mydef}
A \emph{personality profile} is a tuple $\langle sn, \mathit{tf}, ei, pj \rangle \in [-1, 1]^4$ of personality traits.
\end{mydef}

A competence is understood as the knowledge, skills and attitudes that enable a student to successfully solve tasks and face challenges  \cite{roe2002competences}. Moreover, a student possesses every competence with a certain level. Let $C = \{c_1, \dots , c_k\}$ be the set of competencies.

\begin{mydef}
A \emph{student} is represented as a tuple $\langle id, g, \emph{{\bf p}}, l \rangle$ such that:
\begin{itemize}
\item $id$ is the student's identifier;
\item $g \in \{man, {\mathit woman}\}$ stands for the student's gender;
\item $\emph{\bf{p}}$ is a personality profile tuple; 
\item $l: C \to{[0,1]}$  gives the students competence levels, that is, $l(c)$ is the student \emph{competence level} for competence $c$. We assume that when a student does not have a competence (or we do not know about it), $l(c)=0$.
\end{itemize}
Henceforth, the set of considered students is denoted by $A =\{a_1,\ldots, a_n\}$.
\label{AgentDef}
\end{mydef}


The notion of a team is defined as follows, in a straightforward way, as a group of two or more students.

\begin{mydef}[Team] A \emph{team} is any subset of $A$ with at least two students. We denote by $\mathcal{K}_A$ $ = (2^A \setminus \{\emptyset\})\setminus \{\{a_i\}| a_i \in A\}$ the set of all possible teams from students in $A$. 
\end{mydef}

$w(K)$ and $m(K)$ are the number of women and men respectively in team $K$. Students are organised in teams to solve tasks. We understand a task as an instance of a \emph{task type}. A task type not only determines the competences that are required to successfully solve any instance, but also specifies the competence levels and the relative importance of competences. Task types thus differ in requiring different competence levels. A specific task type may require, for instance, a high level of creativity (e.g. to design a city brochure), while another one may require analitycal competences (e.g. to solve mathematical equations). This is formalized as follows.

\begin{mydef}
A task type $\tau$ is a tuple $\langle \lambda, 
{\{(c_{i},l_{i}, w_{i})\}_{i \in I_{\tau}}} \rangle$ where:
\begin{itemize}
\item  $I_{\tau}$ is the index set of the required competencies.
\item $\lambda \in [0,1]$ is the importance given to proficiency; the higher the value of $\lambda$, the higher the team proficiency importance.
\item $c_{i} \in C$ is a competence required to perform the task;
\item $l_{i} \in [0,1]$ is the required competence level for $c_i$; 
\item $w_{i} \in [0,1]$ is the importance of competence $c_i$ for the success in solving an instance of task type $\tau$; and $\sum_{i \in I_{\tau}} w_i = 1$.
\end{itemize}
\label{TaskTypeDef}
\end{mydef}

Tasks are instances of task types plus a required number of students.

\begin{mydef}A \emph{task} $t$ is a tuple $\langle \tau, m \rangle$ such that $\tau$ is a task type and $m$ is the required number of students, where $m\geq 2$.
\end{mydef} 

We note by $T$ the set of tasks and by $\mathcal{T}$ the set of task types. We will note by $C_{\tau} =\{c_{i} | i \in I_{\tau}\}$ the set of competencies required by task type $\tau$.

Given a team and a task, we must consider how to assign responsibilities for the competencies within the team. 
This \emph{competence assignment} is defined as follows.

\begin{mydef}Given a task type $\tau$ and a team $K \in \mathcal{K_A}$, a competence assignment is a function $\eta_{\tau}: K \to 2^{C_{\tau}}$ satisfying that 
$C_{\tau} = \bigcup_{a \in K} \eta_{\tau}(a)$. We note by $\Theta_{\tau}^{K}$ the set of competence  assignments for task type $\tau$ and team $K$.
\label{TaskDef}
\end{mydef}

The list of students assigned to each competence is defined as follows.

\begin{mydef}
Given a task type $\tau$, a team $K$, and competence assignment $\eta_{\tau}$, the set $\delta(c_{i},K,\eta_{\tau}) = \{a \in K | c_{i} \in \eta_{\tau}(a)\}$ stands for those students in team $K$ responsible for competence $c_{i}$.
\end{mydef}


In team-based learning, it is a key requirement that students share responsibilities in order to achieve a successful performance. Hence, our objectives are: (a) to distribute responsibilities in a balanced way across a team; and (b) to have each team member  responsible of at least one competence. This is especially important in an education context, where no one should be cornered within a team. We shall refer to such an assignment as a \emph{balanced competence assignment}. Note that we will be concerned with this particular assignment in this paper.
Hereafter, we note by $\bar{\Theta}_{\tau}^{K}$ the set of balanced competence  assignments for task type $\tau$ and team $K$, where $\bar{\Theta}_{\tau}^{K} \subseteq \Theta_{\tau}^{K}$.

\section{The Problem of Composing Synergistic Teams}\label{sec:problem}


Next, we present our computational model to compose and evaluate teams. First, we introduce 
a way of measuring \emph{proficiency}, namely the degree of matching between a competence assignment and the competences of the members of a team. Thereafter, we provide a measure of 
\emph{congeniality}, namely of the diversity of personalities of the members in a team. Then, the \emph{synergistic} value of a team results from combining both proficiency and congeniality values.


\subsection{How to assess the proficiency value of a team}
\label{ssec:proficiency}

Our goal is to calculate the  \emph{proficiency degree} of a team for a particular task from a competence assignment. With this aim, our measure of proficiency will adhere to the following principle: the closer the competence levels of the team members in a competence assignment to the competence levels required by the task, the larger the proficiency degree of the team. In this way, we pursue to avoid both \emph{under-proficient} and \emph{over-proficient} competence assignments, since they involve under-qualified and over-qualified teams. On the one hand, students in under-proficient teams  may get frustrated because of their lack of knowledge to undertake their assignments. On the other hand, as argued in  \cite{bashur2011impact}, students in over-qualified teams are bound to lose attention and motivation because of the lack of challenge in their assignments. 

Our formal definitions of under-proficiency degree and over-proficiency degree are based on measuring the distance between what is required (in terms of competence levels) by a task and what a team offers to perform the task (according to a competence assignment).



\begin{mydef}[Under-proficiency degree]
The under-proficiency degree of a team $K$ to perform a task of type $\tau$ according to a competence assignment $\eta_{\tau}$ is:
\begin{equation*}
u(\eta_{\tau})=
\sum_{i \in I_{\tau}} w_{i} \cdot \frac{\sum_{a \in \delta(c_{i},K,\eta_{\tau})} |\min(l^{a}(c_{i}) - l_{i},0)|}{ |\delta(c_{i},K,\eta_{\tau})| +1}
\end{equation*}
\end{mydef}

\begin{mydef}[Over-proficiency degree]
The over-proficiency degree of a team $K$ to perform a task of type $\tau$ according to a competence assignment $\eta_{\tau}$ is:
\begin{equation*}
o(\eta_{\tau})=
\sum_{i \in I_{\tau}} w_i \cdot \frac{\sum_{a \in \delta(c_{i},K,\eta_{\tau})} \max(l^{a}(c_{i}) - l_{i},0)}{|\delta(c_{i},K,\eta_{\tau})|+1}
\end{equation*}
\end{mydef}




We combine the under-proficiency and over-proficiency degrees of a team as a weighted average to finally obtain the proficiency degree of a team as follows:


\begin{mydef}
\label{v_value}
The proficiency degree of a team $K$ to perform a task of type $\tau$ following a competence assignment $\eta_{\tau}$, and considering an under-proficiency penalty $\upsilon \in [0,1]$ is:
\begin{equation}
\label{eq:uprof}
u_{\mathit{prof}}(K,\tau) = \underset{\eta_{\tau} \in \bar{\Theta}_{\tau}^{K}}\max(1-(\upsilon \cdot u(\eta_{\tau})+(1-\upsilon) \cdot o(\eta_{\tau})).
\end{equation}
\end{mydef}


Definition \ref{v_value} is restricted to the set of balanced competence assignments $\bar{\Theta}_{\tau}^{K}$, which are the relevant competence assignments in education scenarios, as discussed above.\footnote{A more general definition of proficiency could be readily obtained by considering the set of all competence assignments $\Theta_{\tau}^{K}$ instead. However, we propose this definition for the sake of simplicity.} Furthermore, It is worth noticing that function $u_{\mathit{prof}}(K,\tau)$ in Definition \ref{v_value} is well defined for any team, task type and competence assignment. Indeed, for any task type $\tau$, team $K$, and $\eta \in \Theta_{\tau}^{K}$, we observe that $u(\eta_\tau) + o(\eta_\tau) \in [0,1)$ and $0 \le u_{\mathit{prof}}(K,\tau) < 1$. This is true since no student can be over-proficient and under-proficient at the same time.

From equation \ref{eq:uprof} we observe that the larger the value of importance of the proficiency penalty ($\upsilon$), the larger the importance of the over-proficiency degree. And the other way around, the lower the proficiency penalty, the less important the under-proficiency degree. Hence, setting large values to the proficiency penalty guarantees that competence assignments that make a team under-competent (unable to cope with competence requirements) are penalised. Analogously, small proficiency penalties are meant to penalise over-competent teams. The correct setting of the proficiency penalty parameter 
will depend on each task type. On the one hand, if our objective is to foster 
 \emph{effective} teams, then we must set the proficiency penalty to a large value to penalise more under-proficiency.


In order to computer $u_{\mathit{prof}}(K,\tau)$ we must solve a constrained optimisation problem: find the balanced competence assignment with minimum \emph{cost} (in terms of under- and over-proficiency). This problem can be formulated and solved as a minimum cost flow problem. More precisely, given a team $K \in \mathcal{K_A}$ and a task type $\tau$, we derive the balanced competence assignment $\eta$ that maximises equation \ref{eq:uprof} by solving a minimum cost assignment problem, which in turn can be expressed as an integer linear program (ILP) as follows. The ILP employs a binary variable $x_{ij}$ to encode the decision of whether student $a_i \in K$ is tasked with competence $c_j \in C_{\tau}$, where $C_{\tau}$ is the set of competencies required by task $\tau$. Hereby, the cost of assigning a student $a_i$ to a competence $c_j$, denoted by $p_{ij}$, is defined as follows:

\begin{equation*}
p_{ij} :=
\begin{cases}
(l^{a_i}(c_{\mathit{j}}) - l_{\mathit{j}}) \cdot (1-\upsilon) \cdot w_{\mathit{j}} & \text{if  } l^{a_i}(c_{\mathit{j}} - l_{\mathit{j}}) \geq 0\\
-(l^{a_i}(c_{\mathit{j}}) - l_{\mathit{j}}) \cdot \upsilon \cdot w_{\mathit{j}} & \text{if  } l^{a_i}(c_{\mathit{j}} - l_{\mathit{j}}) < 0
\end{cases}
\label{costeq}
\end{equation*}
\noindent
where $v \in [0,1]$ is the penalty applied to the under-proficiency of team $K$ (see Section~\ref{ssec:proficiency} for a detailed introduction of this term) and $w_{\mathit{j}} \in [0,1]$ weighs the importance of competence $c_j$ to succeed in completing a task of type $\tau$ (see definition~\ref{TaskTypeDef}). 

The above-mentioned minimum cost assignment problem can then be expressed in the following way as an ILP model.
\begin{equation}
\operatorname{min} \sum_{a_i \in K} \sum_{c_j \in C_{\tau}} x_{ij}\cdot p_{ij}
\label{eq:obj_assign_ilp}
\end{equation}
subject to: 
\begin{equation}
\sum_{c_j \in C_{\tau}} x_{ij} \leq \left\lceil \frac{|C_{\tau}|}{|K|} \right\rceil \quad \forall \; a_i \in K
\label{eq:const1_assign_ilp}
\end{equation}
\begin{equation}
\sum_{c_j \in C_{\tau}} x_{ij} \geq 1 \quad \forall \; a_i \in K
\label{eq:const2_assign_ilp}
\end{equation}
\begin{equation}
\sum_{a_i \in K} x_{ij} = 1 \quad \forall \; c_j \in C_{\tau}
\label{eq:const3_assign_ilp}
\end{equation}
Constraint~(\ref{eq:const1_assign_ilp}) makes sure that each student is assigned to at most $\left\lceil \frac{|C_{\tau}|}{|K|} \right\rceil$ competencies, while constraint~(\ref{eq:const2_assign_ilp}) makes sure that each student is assigned to at least one competence. Note that constraints~(\ref{eq:const2_assign_ilp}) are only used if $|C_{\tau}| \geq |K|$. Finally, constraint~(\ref{eq:const3_assign_ilp}) ensures that each competence has exactly one student assigned to it.

The solution to the ILP above allows us to build the balanced competence assignment required to compute $u_{prof}$ in equation \ref{eq:uprof} as follows: for each student $a_i$ in team $K$, $\eta_\tau(a_i) = \{c_j \in C_\tau | x_{ij} = 1\}$.

At this point we have learned how to compute the proficiency value for a team given a particular competence assignment.
However, as argued in the introduction, the degree of proficiency alone is not enough for a team to succeed. Next we introduce a function to measure the \emph{congeniality} within a team from the personalities and genders of its team members. Thus, our congeniality measure does not consider any competence assignment, hence differing from our above-defined proficiency measure.


\subsection{How to assess the congeniality value of a team}
\label{ssec:congeniality}


Recent studies in organisational psychology have proven the existence of a trade-off between the creative productivity caused by ``meta-cognitive conflict'' and ``harmony'' ---good feeling--- in a team \cite{Bradley}.
On the one hand, meta-cognitive conflict stems from the different views of the world that people exhibit based on opposing personality and gender. On the other hand, harmony originates in agreements between people with similar personalities \cite{Wilde2013}. 

 
Based on such observations, in \cite{Wilde2009}, Wilde  proposes several heuristics to target the composition of successful teams.
Along these lines, here we propose to build cognitively diverse teams by employing psychological functions (the SN and TF pairs), psychological attitudes (PJ and EI), and gender.
With the aim of mathematically formalising Wilde's heuristics, we introduce a novel function to measure \emph{congeniality}, $u_{con}$, based on the following objectives: 
 

\begin{enumerate}
    \item the more diverse a team (in terms of the sensing-intuition (SN) and thinking-feeling (TF) personality dimensions of its team members), the larget its congeniality value $u_{con}$;
    \item $u_{con}$ values more teams with at least one member that is extrovert, thinking and judging (with positive EI, TF and PJ personality dimensions), namely exhibiting an ETJ personality;
    \item $u_{con}$ prefers teams that with at least one introvert member (with negative EI personality dimension); and
    \item the more the gender blanace in a team, the larger its congeniality value $u_{con}$.
\end{enumerate}



\begin{mydef}
Given a team $K$ and a task type $\tau$, we define the congeniality degree of the team to perform the task as:


\begin{equation}
u_{con}(K) = u_{SNTF}(K) + u_{ETJ}(K) + u_{I}(K) + u_{gender}(K),
\label{eq:congeniality_utility}
\end{equation}

with:

\begin{enumerate}
\item $u_{SNTF}(K) = \sigma(K,SN) \cdot \sigma(K,TF)$ measures team diversity, where $\sigma(K,SN)$ and $\sigma(K,TF)$ are the standard deviations over the distributions of the SN and TF personality traits for the members of team $K$.
the SN and TF personality trait distributions of the members of team $K$. The larger the values of those deviations, the larger the personality diversity with respect to the SN and TF dimensions, and the larger their product.\footnote{
    Other diversity measures could be used. A possibility would be to understand students as charged particles that distribute in the space as to minimise the overall energy (maximum entropy point). This analogy is appropriate as what is needed in a truly diverse team is that everybody is far from one another as repelling particles are. 
    The repelling force between two particles is proportional to $1/d^2$ where d is the distance between the particles. So given $n$ particles/students, the values (in the dimensions SN, TJ, or EI) that give the minimum energy are: $\arg \min_{f \in F} \sum_{i,j \in A} \frac{1}{(f(i) - f(j))^2}$. Where $f \in F$ is a function that assigns values to students in a particular dimension (SN, TJ or EI).}

\item $u_{ETJ}(K) = \max \{0, \max \{(0,\alpha, \alpha, \alpha) \cdot \mathbf{p^a}| a \in K\}\}$ represents the utility of ETJ personalities, where the importance of each dimension, TF, EI and PJ (the second, third and fourth dimensions of a personality profile) is considered equal and bounded by $\alpha$.

\item $u_{I}(K) = \max \{0, \max \{(0,0, -\beta, 0) \cdot \mathbf{p^a}|a \in K\}\}$  measures the utility of an introvert student, being $\beta$ a value to measure the relevance of introvert students.


\item $u_{gender}(K) = \gamma \cdot \sin(\pi \cdot g(K))$ measures the preference over gender balance. Function $g(K) = \frac{w(K)}{w(K) + m(K)}$ yields the ratio of women in a team considering the number of women ($w(K)$) and men ($m(K)$). The $\gamma$ parameter ($\gamma \le 1$) weighs the importance of gender balance. A team $K$ is perfectly gender-balanced iff $w(K) = m(K)$, and hence $g(K) = 1/2$ and $\sin{(\pi \cdot g(K))} = 1$. Observe that when the number of women and men is equal, it follows that $g(K) = 1/2$ and $\sin{(\pi \cdot g(K))} = 1$. In this case, we say that a team is perfectly balanced.

\end{enumerate}
\end{mydef}

Given the above definition, we now discuss how to choose the values of parameters $\alpha$, $\beta$ and $\gamma$, as they affect the congeniality degree of a team.

If all factors in $u_{\mathit{con}}(K)$ are equally important, then the values of $\alpha$, $\beta$ and $\gamma$ are interdependent and will ultimately depend on the shape of the distribution of the personality traits. Next we analyse two extreme cases and give the actual values that have been used in the experiments.

\begin{itemize}
    \item \emph{Distribution with maximal variance}. The maximum value of $u_{ETJ}(K)$ will be $3\alpha$ in case there is a student $a$ such that $p^a = (k,1,1,1)$, where $k \in [-1,1]$ . 
    If $u_{SNTF}(K)$ and $u_{ETJ}(K)$ have to have the same importance then we need to equate the maximum value of $\sigma(K,SN) \cdot \sigma(K,TF)$ to $3\alpha$. The distribution over an interval $[a,b]$ that has maximal variance corresponds to the distribution with the elements evenly situated at the extremes of the interval with $\sigma^2 \le ((b-a)/2)^2$. That distribution would correspond to teams whose students have values on dimensions SN and TJ at the extremes of the interval $[-1,1]$. Regarding our particular case, which considers the $[-1,1]$ for personality traits ($b=1$, $a=-1$), the value of $\sigma$ for that distribution would be $1$, and hence that would imply that $3\alpha = \sigma(K,SN) \cdot \sigma(K,TF) \le 1$ and thus $\alpha \le 1/3 = 0.33$.
    
    
    \item \emph{Uniform distribution}. If the distributions for SN and TF values follow a uniform distribution, then the variance of each distribution is $\sigma^2 \le \frac{(b-a)^2}{12}$ and thus $\sigma(K,SN) \cdot \sigma(K,TF)= \frac{(b-a)^2}{12}=3\alpha$ which implies that $\alpha \le 0.11(1).$
\end{itemize}

These cases represent extreme situation, whereas real-world scenarios usually lie in the middle.
Thus, we define $0.11(1) \le \alpha \le 0.33(3)$.
Then, in order to make $u_{SNTF}(K)$ and $u_{I}(K)$ equally important, it follows that $\beta \approx  3\alpha$.
Finally, by setting $\gamma \approx 3\alpha$ we make the gender factor equally important to the rest of factors in equation \ref{eq:congeniality_utility}.


\subsection{Evaluating synergistic teams}\label{ssec:synergy}

We now define our performance measure to evaluate each team.
Specifically, a team $K$ is effective when it is both \emph{proficient} and \emph{congenial}. This means that a team counts on the required competences required to perform a task, and also that it shows a balance of gender and personalities so that students will work well together.
The \emph{synergistic value} of a team results from aggregating its proficiency and congeniality values as follows:

\begin{mydef}
\label{synergistic}
Given a team $K$, its synergistic value to perform an instance of $\tau$ is:
\begin{equation}\label{eqn:synergistic-value}
s(K,\tau) = \lambda \cdot u_{\mathit{prof}}(K,\tau) + (1-\lambda) \cdot u_{con}(K),
\end{equation}
where $\lambda \in [0,1]$ weighs the importance of proficiency.
\end{mydef}

As the value of $\lambda$ determines the relative importance of the congeniality and proficiency factors, its definition depends on the task type.
As an example, congeniality is more important to solve tasks that require a high level of creativity (e.g., tasks tackled for the first time), hence, in this case, $\lambda<0.5$.
On the other hand, proficiency is crucial to complete tasks appearing in, for instance, sport competitions or disaster management, which require the fast operation of teams. In those cases, we must set  $\lambda$ to a value greater than $0.5$

\subsection{Problem definition}
\label{ssec:synergisticProblem}

Given a set of students $A$, we aim at partitioning $A$ into teams so that each team is balanced (in terms of competencies, gender and personality) and team sizes are even.
Henceforth, we shall refer to balanced (i.e., both congenial and proficient) teams as \emph{synergistic teams}.
Any partition of $A$ into teams is denoted as a team partition.
Furthermore, we are interested in forming team partitions whose teams are constrained by size $m$ as follows, as this constraint usually applies in educational contexts.

\begin{mydef}\label{def:constrained}
Given a set of students $A$, a team partition $P_m$ of $A$ is constrained by size $m$, $2 \leq m \leq |A|$, iff for every team $K \in P_m$, $m \leq |K| \leq m+1$\footnote{Since $|K| / m$ is not necessarily a natural number, we allow $m \leq |K| \leq m+1$. In practice, we want partitions whose teams differ in size by at most one student.} holds.
\end{mydef}
Henceforth, we will focus on the set ${\mathcal P}_m(A)$ of team partitions of set $A$ constrained by some size $m$.

As mentioned above, our objective is to compute a partition whose teams are as good as possible. Thus, we want to disregard unbalanced partitions composed of some teams that perform  well and some other teams that perform badly. Therefore, we target at partitions whose teams display homogeneous behaviours (similar performances). This leads to the definition of a measure for the overall performance of a partition that prefers homogeneous teams. Thus, our definition below defines the synergistic value of a team partition as the Bernoulli-Nash product of the teams' synergistic values. In this way, this function ensures that we give larger values to \emph{fair} partitions \cite{nash} (containing homogeneous teams), unlike other functions like, e.g., the addition.



\begin{mydef}
Given a team partition $P_m$ and task type $\tau$, the synergistic value of $P_m$ is
\begin{equation}
S(P_m, \tau) = \prod_{K\in P_m} s(K,\tau).
\end{equation}
\end{mydef}
\noindent

Now we are ready to formally define the Synergistic Team Composition Problem (STCP) as the problem of  finding the partition with the largest synergistic value.

\begin{mydef}
Given a set of students $A$ and task type $\tau$, the \emph{synergistic team composition problem (STCP)} is the problem of finding a team partition constrained by size $m$, $P^*_m \in \mathcal{P}_m(A)$, that maximises  $S(P_m, \tau)$, namely
$$P^*_m = \underset{P_m \in \mathcal{P}_m(A)}{\arg \max} S(P_m,\tau).$$
\end{mydef}

\section{\textcolor{black}{A complete algorithm for the STCP}}\label{sec:optimal}

\textcolor{black}{We now propose a complete algorithm to solve the STCP.
As a first step, in Section \ref{ssec:linear} we show how we linearise the problem, allowing us to
model the STCP as an ILP.
Then, in Section \ref{ssec:optimalAlg} we detail a complete algorithm for the STCP that solves such an ILP.}

\subsection{Linearising the STCP}\label{ssec:linear}

Given a set $A$ of $n$ students, a task $t$ of type $\langle \tau, m \rangle$, we define the total number of teams $b = \left\lfloor  n/m\right\rfloor$.
Depending on the cardinality of $A$ and the desired team size $m$, the number of students in each team may vary.
Let $Q(n,m)$ denote a set of couples such that each $(x,y)\in Q(n,m)$ indicates that we consider $x$ teams of size $y$.\footnote{For simplicity, in our experiments in Section~\ref{sec:results} we consider that the number of students is a multiple of the desired team size, i.e., $n\bmod m = 0$. In this case, $Q(n,m)=\{(b, m)\}.$}
We refer to $Q(n,m)$ as the \emph{the quantity distribution of team sizes}.




Let $K_1,\ldots,K_q$ denote the complete set of feasible teams that can be generated on the basis of the students from $A$, and $s(K_1,\tau),\ldots,s(K_q,\tau)$ their synergistic values given a task $t=\langle \tau, m \rangle$. 
Finally, let $C$ be a matrix of size $n \times q$ such that $c_{ij}=1$ if student $a_i$ is part of team $K_j$, and $c_{ij}=0$ otherwise.

For each team $K_j$ ($j=1,\ldots,q$) we consider a binary decision 
variable $x_j$. The value of $x_j$ indicates whether team $K_j$ is selected or not as part of the optimal solution of the STCP. Solving the STCP, then, amounts to solving the following non-linear integer program:


\begin{equation}
\operatorname{max} \prod_{j=1}^{q} {s(K_j, \tau)}^{x_j} 
\label{eq:nlp}
\end{equation}
subject to: 
\begin{equation}
\sum_{j=1}^{q} x_j = b
\label{eq:sizeConstraint}
\end{equation}
\begin{equation}
\sum_{j=1}^{b} c_{ij} \cdot x_j = 1 \quad \forall 1 \leq i \leq n
\label{eq:agentXOR}
\end{equation}
\begin{equation}
x_j \in \{0,1\} \quad 1 \leq j \leq q
\label{eq:domain}
\end{equation}

Constraint \ref{eq:sizeConstraint} ensures that any valid solution consists of exactly $b$ teams, whereas constraint \ref{eq:agentXOR} enforces that each student belongs to exactly one of the selected teams. Notice that the objective function (see Equation~\ref{eq:nlp}) is non-linear. Nevertheless, it is rather easy to linearise this objective function by maximising the logarithm of \linebreak[4] $\prod_{j=1}^{q} {s(K_j,\tau)}^{x_j}$ instead. Thus, solving the non-linear integer program above is equivalent to solving the following binary linear program:
\begin{equation}
\operatorname{max} \sum_{j=1}^{q} x_j \cdot log(s(K_j,\tau)) 
\label{eq:lp}
\end{equation}
subject to: equations \ref{eq:sizeConstraint}, \ref{eq:agentXOR}, and \ref{eq:domain}. 





\subsection{Solving the ILP model} \label{ssec:optimalAlg}

\textcolor{black}{Algorithm \ref{alg:optimalSolving} presents the pseudo-code of our complete approach to solve the STCP by means of the above detailed ILP.}
First, we generate the input for this ILP (see lines~2~to~4).
Specifically, line~2 generates all possible teams of size $m$ as determined by the quantity distribution $Q(|A|,m)$. The best synergistic values of these teams are computed in lines~3~and~4. This involves solving an optimisation problem, as discussed at the end of Section \ref{ssec:proficiency}. We then generate the ILP according to equation~\ref{eq:lp} and solve it
with the aid of an off-the-shelf ILP solver such as, for example, CPLEX, Gurobi, or GLPK.
\textcolor{black}{If given sufficient time, the algorithm returns an optimal solution (that is, an optimal team partition) together with the competence assignments (line~7).}


\begin{algorithm}[h]
\caption{STCPSolver}
\label{alg:optimalSolving}
\begin{algorithmic}[1]
    \Require $A$ \Comment{The set of students}
    \Require $t=\langle \tau, m \rangle$ \Comment{Task}
     \Ensure $(P, \bm{\eta^*})$ \Comment{Best partition found and best assignments}
     \State $P \gets \emptyset$
    \State $[K_1,\ldots,K_q] \gets \mathit{GenerateTeams(A,Q(|A|, m))}$
     \For {$i \in [1..q]$}
     	\State $(s(K_i,\tau),\eta_{\tau}^i(K_i,\tau)) \gets \mathit{getBestSynergisticValue}(K_i,t)$
     \EndFor
     \State $ILP \gets \mathit{generateILP}([K_1,\ldots,K_q],[s(K_1,\tau),\ldots,s(K_q,\tau)],b)$
	\State $P \gets solve(ILP)$ \\
    \Return $(P,\{\eta_{\tau}^i(K_i,\tau)\}_{K_i \in P})$
\end{algorithmic}
\end{algorithm}


\noindent
We remark that generating the input for STCPSolver takes linear time with respect to the number of feasible teams $q$, which grows rapidly with increasing $m$ and $n$.


\textcolor{black}{\section{A heuristic algorithm for the STCP}} \label{sec:approxAlg}

In this section we present \emph{\SynTeam{}}, an algorithm based on local search. The pseudo-code of \SynTeam{} is provided in Algorithm~\ref{alg:synteam}. \SynTeam{} starts by generating an initial solution/partition (line~1). This is done by randomly ordering the set of students and assigning them, one after the other, in this order, to a number of teams whose sizes are determined by $Q(|A|,m)$; see Section~\ref{ssec:linear} for the definition of $Q(.,.)$. This initial solution is denoted by $(P,S(P,\tau),\bm{\eta})$, where $\bm{\eta}$ is the vector of balanced competence assignments used to compute the proficiency degrees of the teams in $P$. The assignment of students to competencies is done as described in Section~\ref{ssec:proficiency}. The main part of the algorithm consists in a local search procedure which makes use of two different neighbourhoods. This first one, which is applied by default, consists in randomly selecting two teams from the current solution. Then, the set of students contained in these two teams is redistributed in the optimal way into two (possibly new) teams and the resulting solution, together with the corresponding competence assignments, is stored in $(P', S(P',\tau),\bm{\eta}')$; see line~4. In addition, whenever the algorithm detects that $n_l$ not necessarily consecutive, non-improving iteration were performed, the second---more fine-grained---neighbourhood is applied to the current solution $(P, \bm{\eta})$ in the following way in line~6 of Algorithm~\ref{alg:synteam}.\footnote{Note that an iteration is called \emph{improving} in case the current solution is improved in line~4 of Algorithm~\ref{alg:synteam}.} The second neighbourhood tries to identify---in ascending order determined by team and student indexes---two students from different teams whose swap results in an improved solution. The first improved solution that is found in this way (if any) is stored as $(P', \bm{\eta}')$. Moreover, counter $c_l$ regarding the non-consecutive non-improving iterations is re-initialized. The algorithm terminates after a number of $n_r$ consecutive non-improving iterations. 

\begin{algorithm}[htbp]
\caption{\SynTeam}
\label{alg:synteam}
\begin{algorithmic}[1]
    \Require $A$ \Comment{List of students}
    \Require $t=\langle \tau, m \rangle$ \Comment{Task}
    \Require $n_r$ \Comment{Max.~number of consecutive non-improving~iterations}
    \Require $n_l$ \Comment{Number of non-improving iterations before student-swap}
     \Ensure $(P, \bm{\eta})$ \Comment{Best solution found}
     \State $(P, S(P,\tau),\bm{\eta}) \gets \mathit{GenerateRandomSolution}(A, Q(|A|, m))$
     \State $c_r \gets 1$, $c_l \gets 1$
     \While{$c_r \leq n_r$}
       \State $(P', S(P',\tau),\bm{\eta}') \gets \mathit{GenerateNeighbour}(P, \bm{\eta})$
       \If{$S(P',\tau) \leq S(P,\tau)$ {\bf and} $c_l = n_l$}
           \State $(P', S(P',\tau), \bm{\eta}') \gets \mathit{ApplyImprovingSwap}(P, \bm{\eta})$
           \State $c_l \gets 1$
       \EndIf
       \If{$S(P',\tau) > S(P,\tau)$}
           \State $(P, S(P,\tau),\bm{\eta}) \gets (P', S(P',\tau),\bm{\eta}')$
           \State $c_r \gets 1$, $c_l \gets 1$
       \Else
           \State $c_r \gets c_r + 1$, $c_l \gets c_l + 1$
       \EndIf
     \EndWhile
    \Return $(P, \bm{\eta})$
\end{algorithmic}
\end{algorithm}

\section{Computational Results} \label{sec:results}
\noindent
In this section we conduct a comprehensive experimental evaluation in order to compare the two STCP solvers proposed in this work: (1) the optimal solver (\cplex{}), and (2) the \SynTeam{} solver which is based on local search. In particular, we compare the two approaches regarding their run-times, as team sizes and the number of students increase. Moreover, we study the quality of the solutions provided by \SynTeam{} in comparison to the optimal solutions. Finally, we also examine the anytime performance of \SynTeam\ with respect to \cplex.

\subsection{Computational Scenario}
\label{subsec:empiricalSettings}

The empirical evaluation is done within the following scenario:

\begin{table}[t!]
\caption{Specification of the four task types, designed by education professionals.}
  \label{tab:task-types}
\begin{subtable}[t]{0.45\textwidth}
\centering
\subcaption{Task type \texttt{body rythm}}
    \label{fig:task-types:body_rythm}
\scalebox{0.72}{
\begin{tabular}{lcc} \hline
	Competence           & Req.~level & Importance \\ \hline 
\textsc{bodily\_kinesthetic} & advanced & very important  \\
\textsc{musical}      & intermediate  & fairly important  \\
\textsc{linguistic}                & intermediate & slightly important \\
\textsc{interpersonal}         & advanced & very important \\
\textsc{visual\_spatial}       & novice & slightly important \\ \hline
\end{tabular}}
\end{subtable}
\begin{subtable}[t]{0.45\textwidth}
\centering
\subcaption{Task type \texttt{entrepreneur}}
    \label{fig:task-types:entrepreneur}
\scalebox{0.72}{
\begin{tabular}{lcc} \hline
	Competence           & Req.~level & Importance \\ \hline 
\textsc{linguistic}                & advanced & fairly important \\
\textsc{logic\_mathematics}     & intermediate  & very important  \\
\textsc{visual\_spatial}       & novice & slightly important \\
\textsc{musical}      & novice & slightly important \\
\textsc{interpersonal}         & advanced & very important  \\
\textsc{intrapersonal}         & intermediate  & important  \\ \hline
\end{tabular}}
\end{subtable}
\begin{subtable}[t]{0.48\textwidth}
\centering
\subcaption{Task type \texttt{arts design}}
    \label{fig:task-types:arts_design}
\scalebox{0.72}{
\begin{tabular}{lcc} \hline
	Competence           & Req.~level & Importance \\ \hline 
\textsc{linguistic}                & novice & slightly important \\
\textsc{visual\_spatial}       & advanced & very important \\
\textsc{intrapersonal}         & intermediate  & fairly important  \\ \hline
\end{tabular}}
\end{subtable}
\begin{subtable}[t]{0.48\textwidth}
\centering
\subcaption{Task type \texttt{English}}
    \label{fig:task-types:english}
\scalebox{0.72}{
\begin{tabular}{lcc} \hline
	Competence           & Req.~level & Importance \\ \hline 
\textsc{linguistic}                & intermediate  & very important \\
\textsc{intrapersonal}         & novice & important \\
\textsc{interpersonal}         & advanced & very important \\ \hline
\end{tabular}}
\end{subtable}
\end{table}

\begin{itemize}
\item \textbf{ILP Solver.} CPLEX Optimization Studio v12.7.1 \cite{cplex} was used for solving the ILPs generated by \cplex. 

\item \textbf{Students.}  Actual-world data, in terms of 210 students, each one identified by an ID, gender information, the personality profile, and the competence levels regarding seven competencies, was used.

\textcolor{black}{
\item \textbf{Classroom size.}  The total number of students (n) in a classroom ranges from 10 to 100.
}

\item \textbf{Task type.} In contrast to our preliminary paper~\cite{AndEtAl2018:prima}, we consulted with professionals from the educational sector in order to define four diverse task types. These task types, together with their required competencies and importance levels, are provided in Table~\ref{tab:task-types}. As usual in educational contexts, these task types require a subset of seven competencies that directly stem from Gardner's multiple intelligences~\cite{gardner1987theory}\textcolor{black}{, which is widely-used in education scenarios}: \textsc{linguistic}, \textsc{logic\_mathematics}, \textsc{visual\_spatial}, \textsc{bodily\_kinesthetic}, \textsc{musical}, \textsc{intrapersonal}, and \textsc{interpersonal}. Notice that requirement levels and importance levels were set by educators to \emph{qualitative} values to ease their specification. We employed five qualitative values for requirement levels (\emph{fundamental awareness}, \emph{novice}, \emph{intermediate}, \emph{advanced}, \emph{expert}) and five qualitative values for importance degrees (\emph{unimportant}, \emph{slightly important}, \emph{important}, \emph{fairly important}, \emph{very important}). Thereafter, we evenly mapped the qualitative labels in Table~\ref{tab:task-types} to quantitative values within the $[0,1]$ interval.


\item \textbf{Task.} The team size ($m$) ranged from $2$ to $6$. \textcolor{black}{On the one hand, this limitation comes from the fact that larger team sizes are rather rare in an educational context (see e.g. \cite{Apedoe2012,pujolas:2001:ADA,pujolas:2003:AJA})\footnote{\textcolor{black}{In fact, this actual-world constraint is in line with studies in the organisation psychology literature showing an inverse relationship between the size of a team and its performance \cite{levine1990progress, Oyster,Bartol, moon2005detailed}. \cite{Oyster} observes that in the case of a team containing more than six people
there is a tendency to split the team into two, which brings about negative effects. The cause is twofold: high coordination costs and loss of motivation by team members.}}
On the other hand, the computational burden for \cplex{} when handling team sizes larger than $6$ was too high, making comparisons with optimal solutions too costly.}

\item \textbf{Team proficiency.}
An intermediate value of $\upsilon = 0.5$ was used for all computational tests, because the specific team proficiency value is rather irrelevant for the study of the algorithm properties. 


\item \textbf{Team Congeniality.}
With the aim of making the personality and gender requirements equally relevant, the importance values were set as follows: (1) $\alpha = 0.11$, (2) $\beta = 3 \cdot \alpha$, (3) $\gamma = 0.33$. Note that a detailed description of the meaning of these values can be found in Section~\ref{ssec:congeniality}

\item \textbf{Balance between Proficiency and Congeniality.}
The value of parameter $\lambda \in [0,1]$ determines the balance between team proficiency and team congeniality. 
The experiments are performed with $\lambda \in \{0.2, 0.5, 0.8\}$. However, for space reasons we do not report on the results for $\lambda = 0.5$. They can be obtained from the authors upon request.

\item \textbf{Number of iterations without improvement ($n_r$).} This value was fixed to $1.5\cdot b$, with the following rationale behind. The computation time requirements of \SynTeam{} can be expected to relate strongly to the number of teams in a solution ($b$): the larger $b$, the higher the computation time requirements. Moreover, after studying the evolution of \SynTeam{} over time, we decided to scale this value with 1.5.

\item
\textbf{Frequency of local search ($n_l$).} 
It was observed that, after performing $\approx \frac{n_r}{6}$ applications of the first neighbourhood without improvement, the probability of finding an improvement in this way was very low. Hence, the value of $n_l$ was set to $\frac{n_r}{6}$.
\end{itemize}

\subsection{Computational Results} \label{ssec:CompResults}

The experiments were performed on a cluster of machines with Intel\textsuperscript{\textregistered} Xeon\textsuperscript{\textregistered} CPU 5670 CPUs with 12 cores of 2933 MHz and a minimum of 40 Gigabytes of RAM. Moreover, IBM ILOG CPLEX v12.7.1 was used within \cplex\ and \SynTeam. Remember, in this context, that CPLEX is used internally by \SynTeam\ for the calculation of the optimal assignment of students to tasks for any given team and task type. CPLEX was run, in all cases, in one-threaded mode for ensuring a fair comparison.


\textcolor{black}{In order to generate problem instances (classrooms) we considered the following parameters: task type, balance between proficiency and congeniality ($\lambda$), total number of students ($n$), and team size ($m$). 
For each combination of these parameters, we generate 20 problem instances, each one by randomly selecting $n$ students out of the set of 210 available students. For instance, when considering team size 2, we have 46 different values for $n$, 3 values for $\lambda$, and 4 task types. Thus, we generate $11040 = 46 \cdot 3 \cdot 4 \cdot 20$ different problem instances. Overall, considering all team sizes and parameters' values, we generate 31440 problem instances.  Then, we solved all these problem instances with both \cplex\ and \SynTeam. The results will be shown as averages over the 20 random sets of students.
}

\begin{figure}[p] 
\centering
	\begin{subfigure}{0.49\textwidth}
		\includegraphics[width=\linewidth]{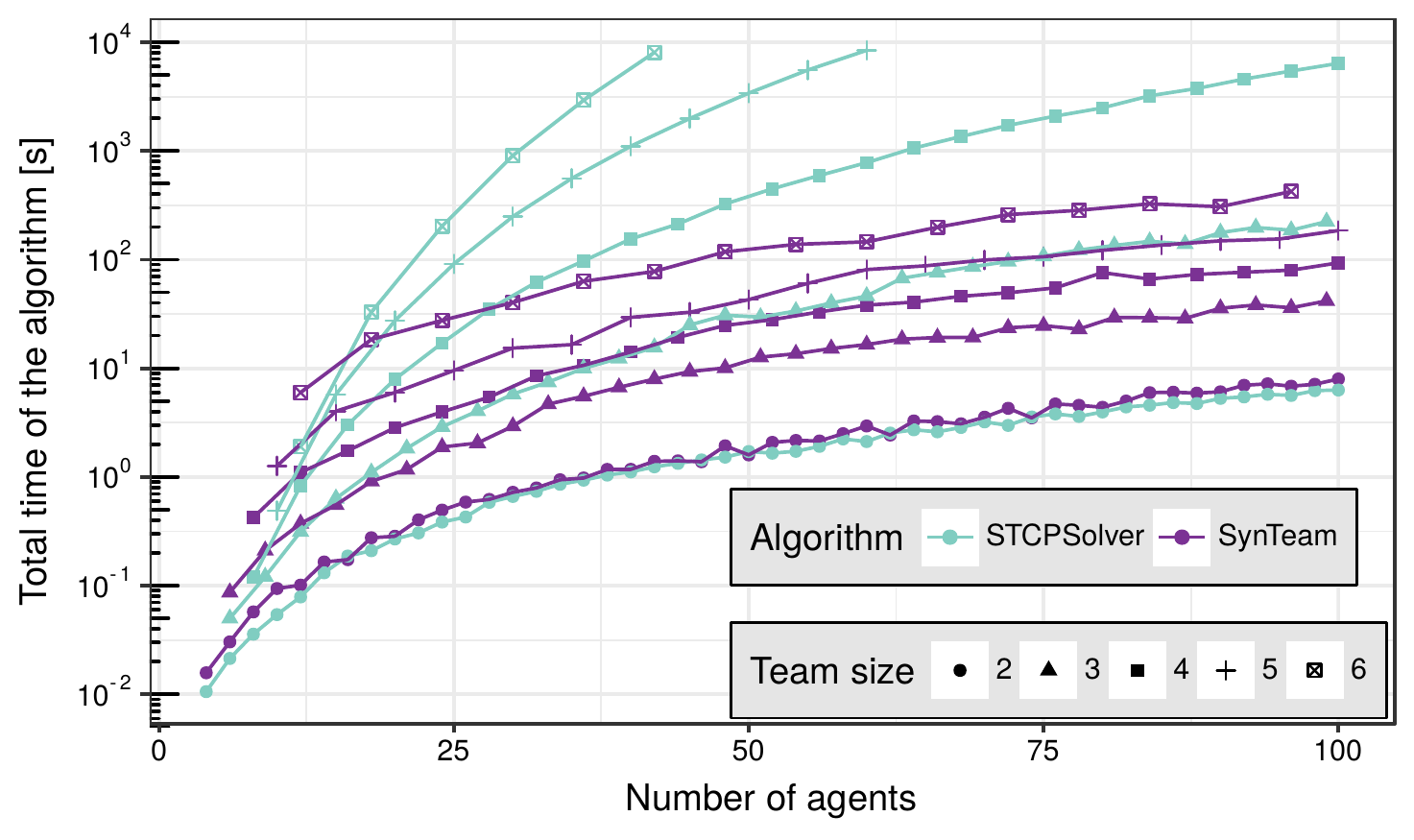}
		\caption{\texttt{body rythm}, $\lambda = 0.2$.}
		\label{figure:TotalTimes:body_rythm:lambda02}
	\end{subfigure}
	\begin{subfigure}{0.49\textwidth}
		\includegraphics[width=\linewidth]{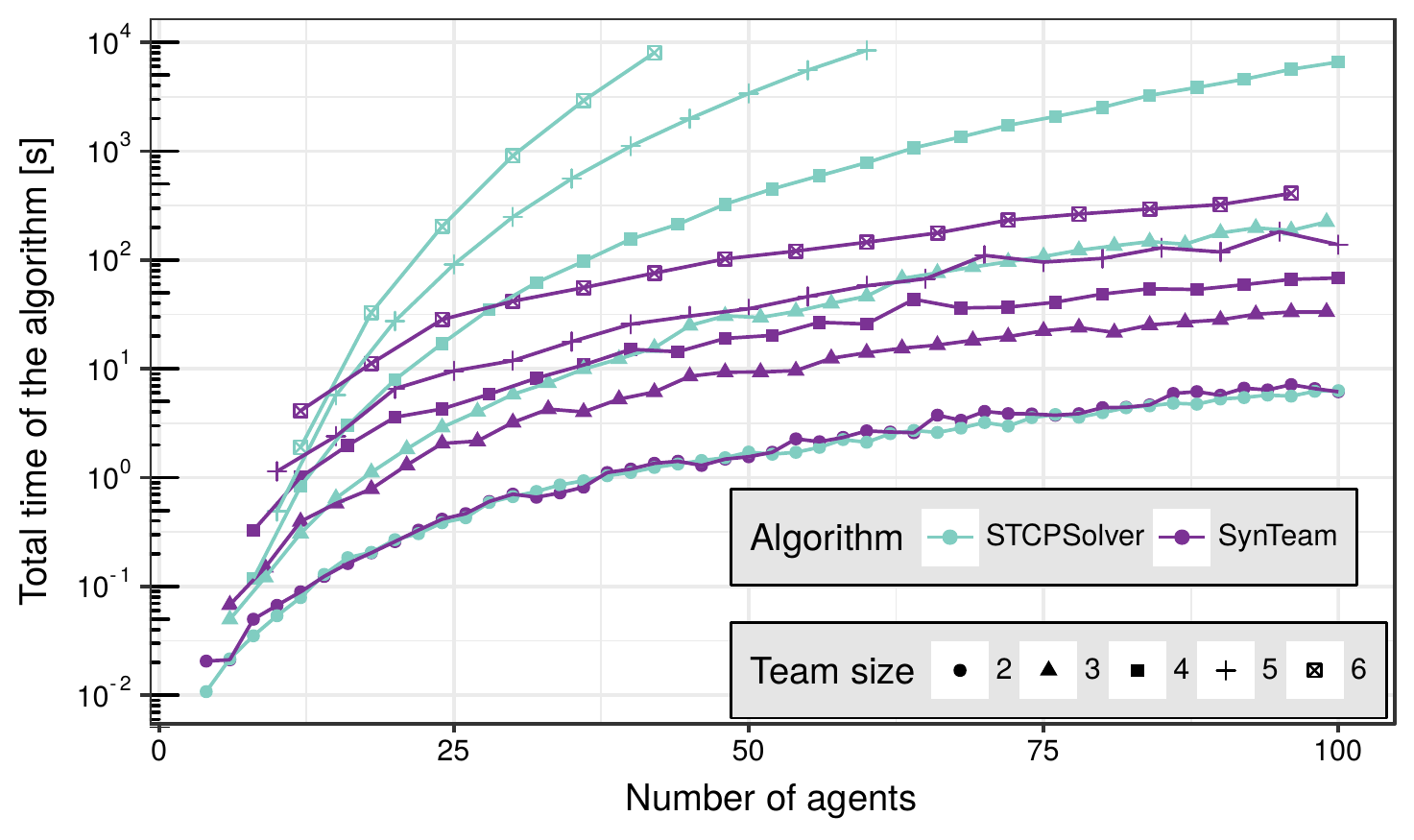}
		\caption{\texttt{body rythm}, $\lambda = 0.8$.}
		\label{figure:TotalTimes:body_rythm:lambda08}
	\end{subfigure}
	\begin{subfigure}{0.49\textwidth}
		\includegraphics[width=\linewidth]{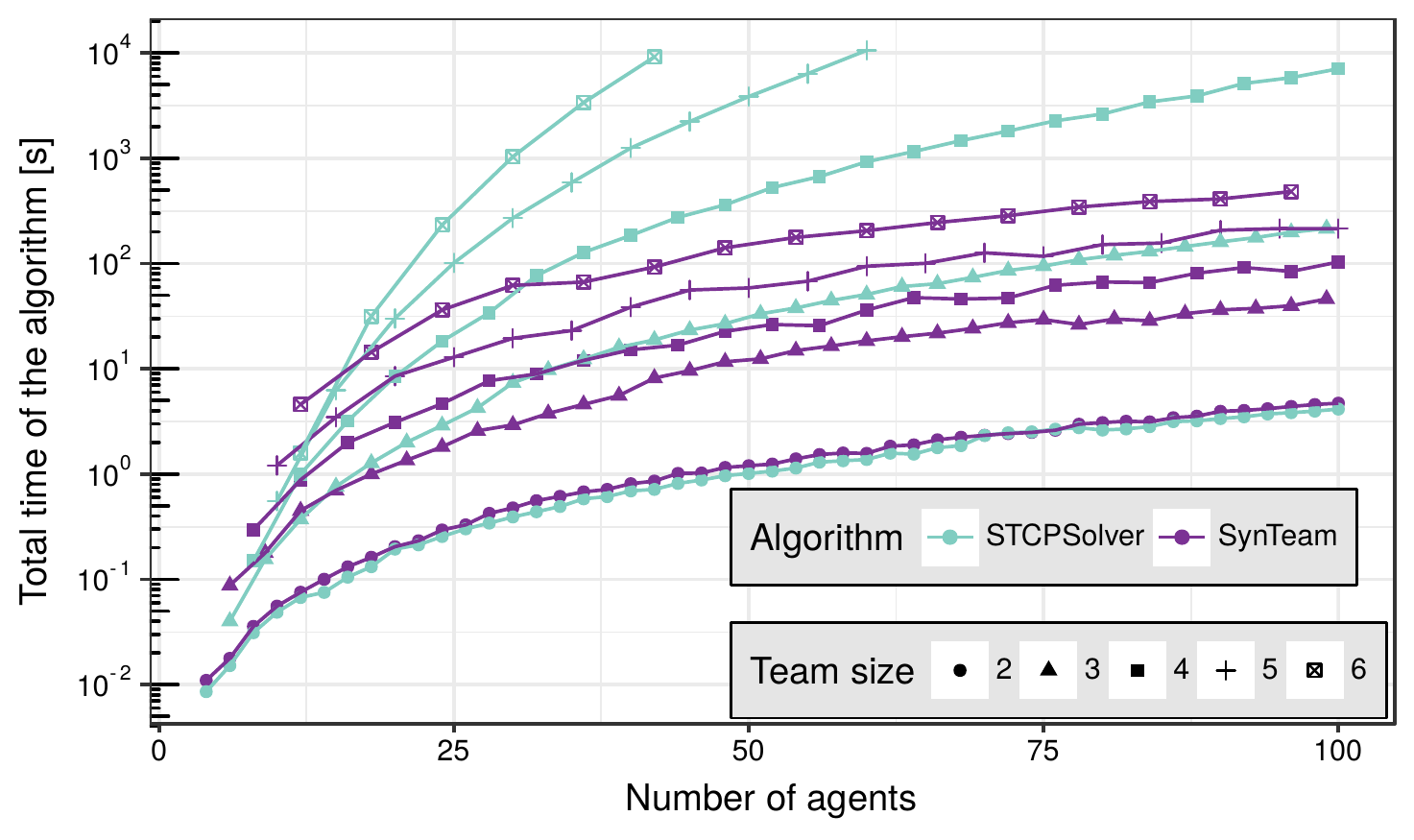}
		\caption{\texttt{entrepreneur}, $\lambda = 0.2$.}
		\label{figure:TotalTimes:entrepreneur:lambda02}
	\end{subfigure}
	\begin{subfigure}{0.49\textwidth}
		\includegraphics[width=\linewidth]{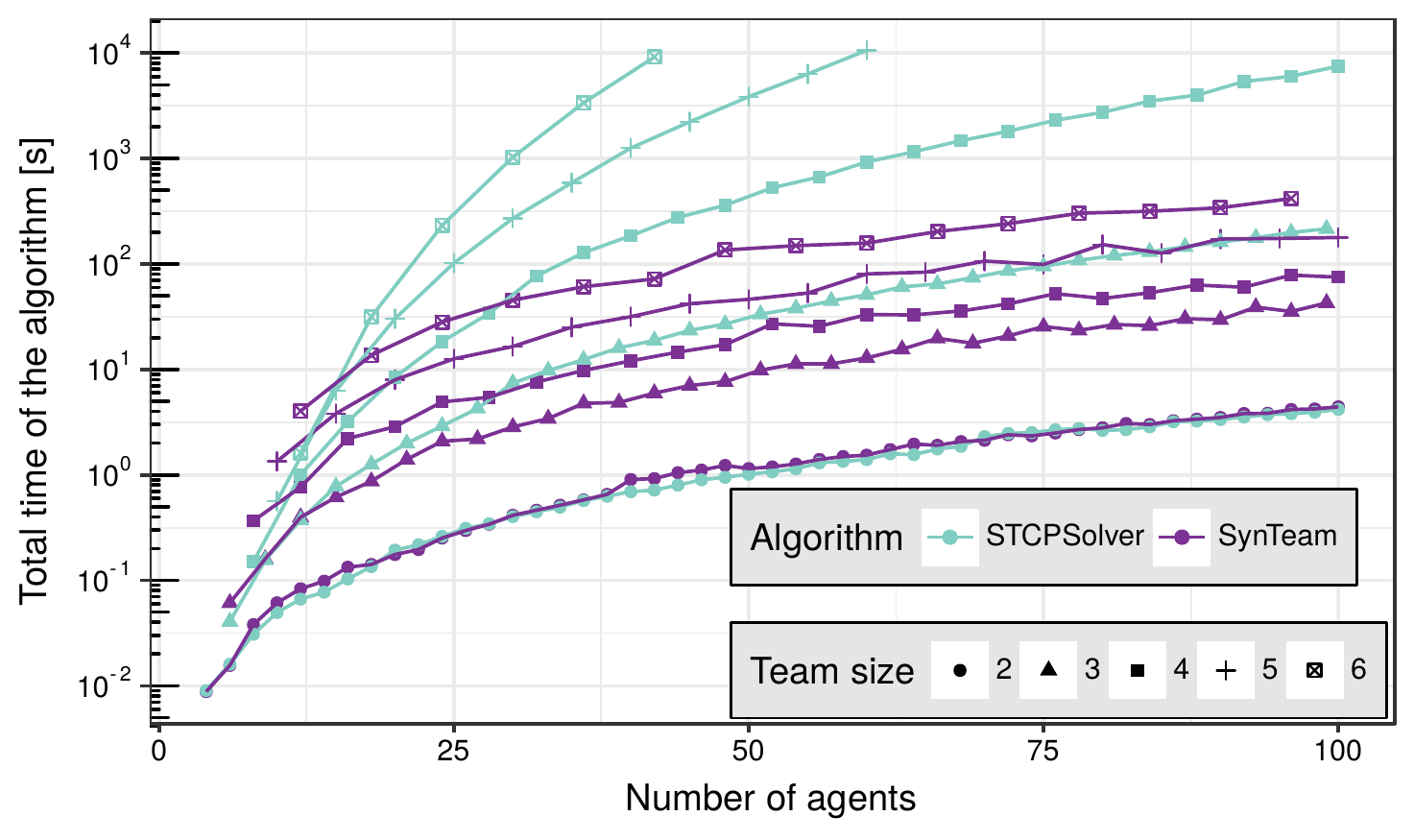}
		\caption{\texttt{entrepreneur}, $\lambda = 0.8$.}
		\label{figure:TotalTimes:entrepreneur:lambda08}
	\end{subfigure}
	\begin{subfigure}{0.49\textwidth}
		\includegraphics[width=\linewidth]{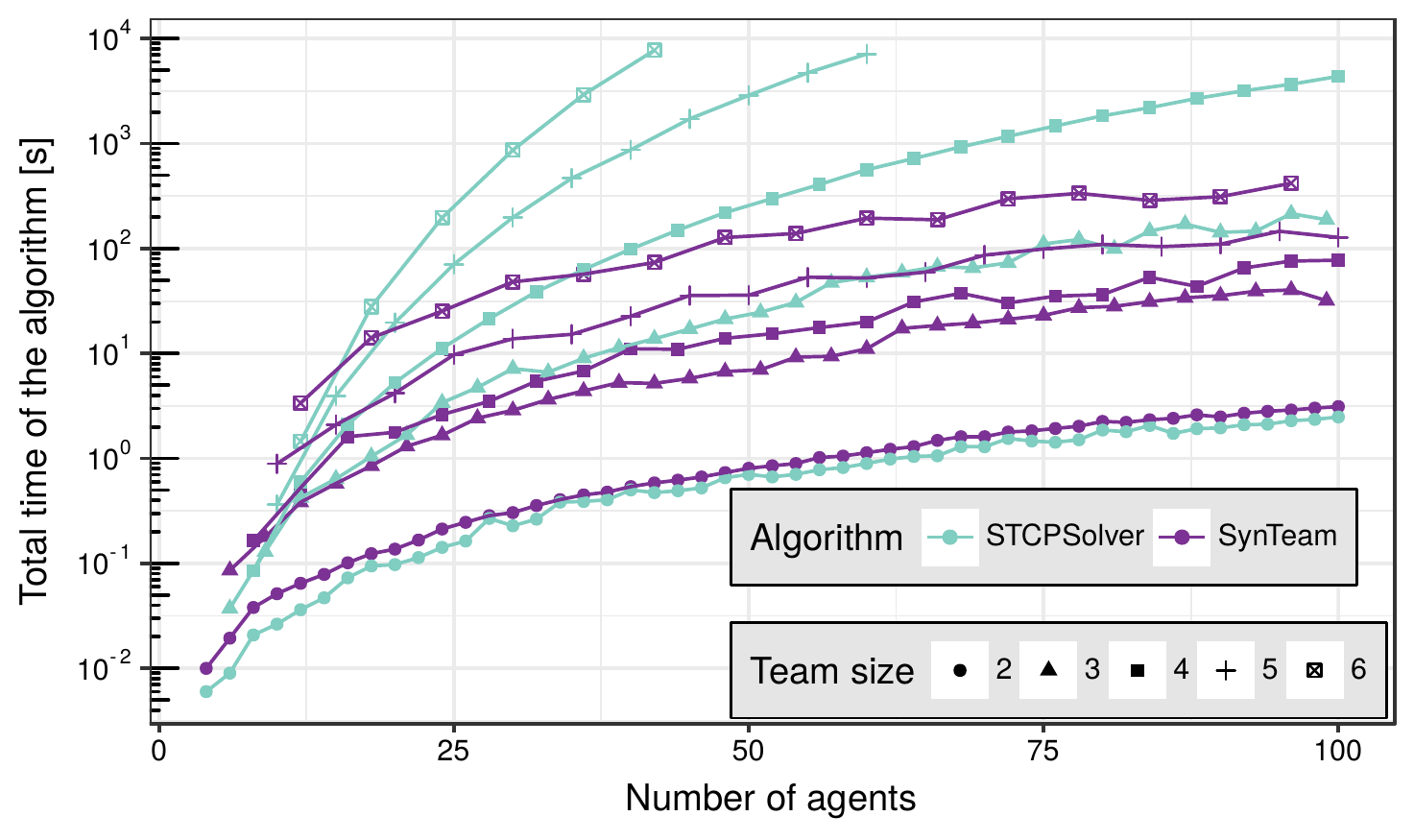}
		\caption{\texttt{arts design}, $\lambda = 0.2$.}
		\label{figure:TotalTimes:arts_design:lambda02}
	\end{subfigure}
	\begin{subfigure}{0.49\textwidth}
		\includegraphics[width=\linewidth]{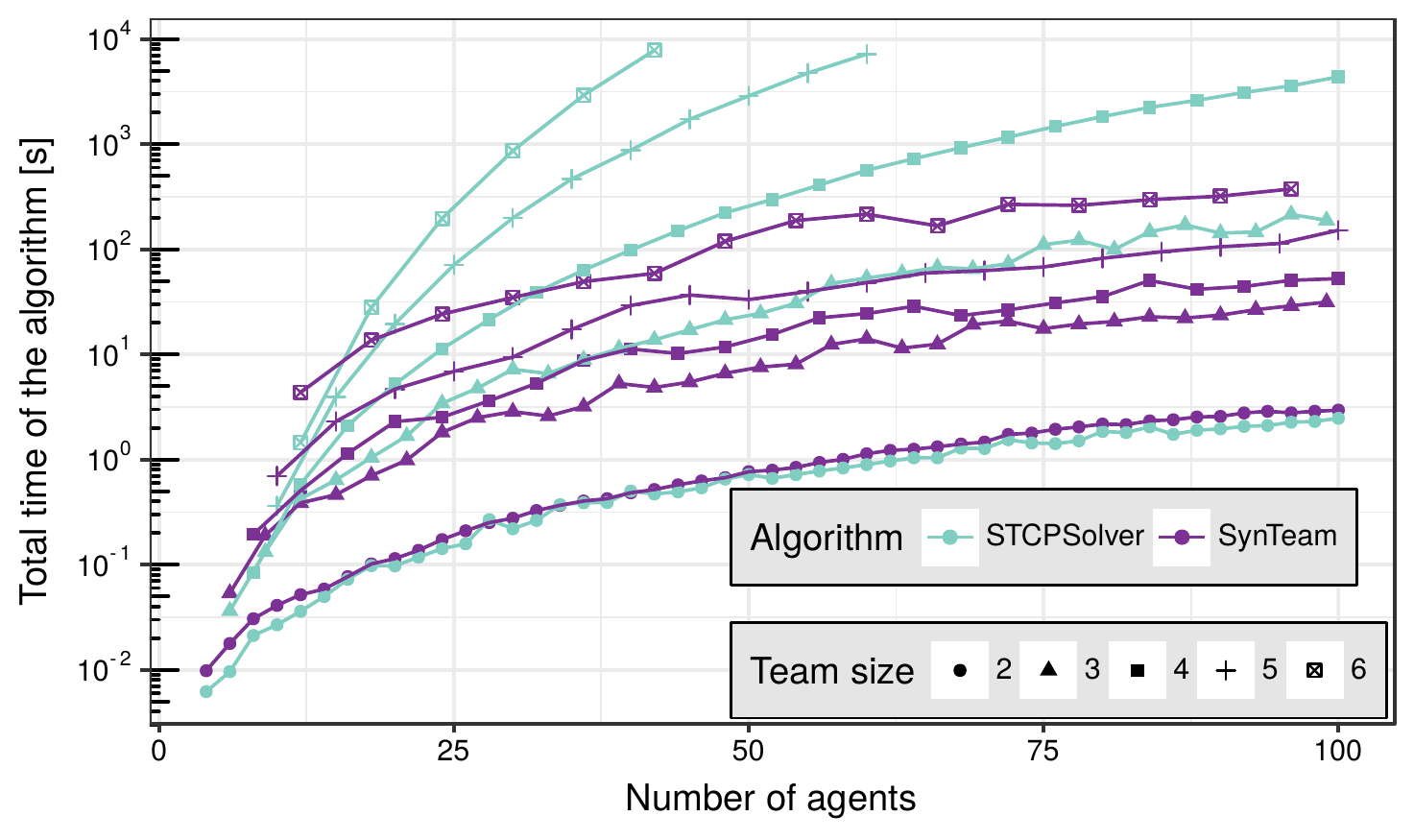}
		\caption{\texttt{arts design}, $\lambda = 0.8$.}
		\label{figure:TotalTimes:arts_design:lambda08}
	\end{subfigure}
	\begin{subfigure}{0.49\textwidth}
		\includegraphics[width=\linewidth]{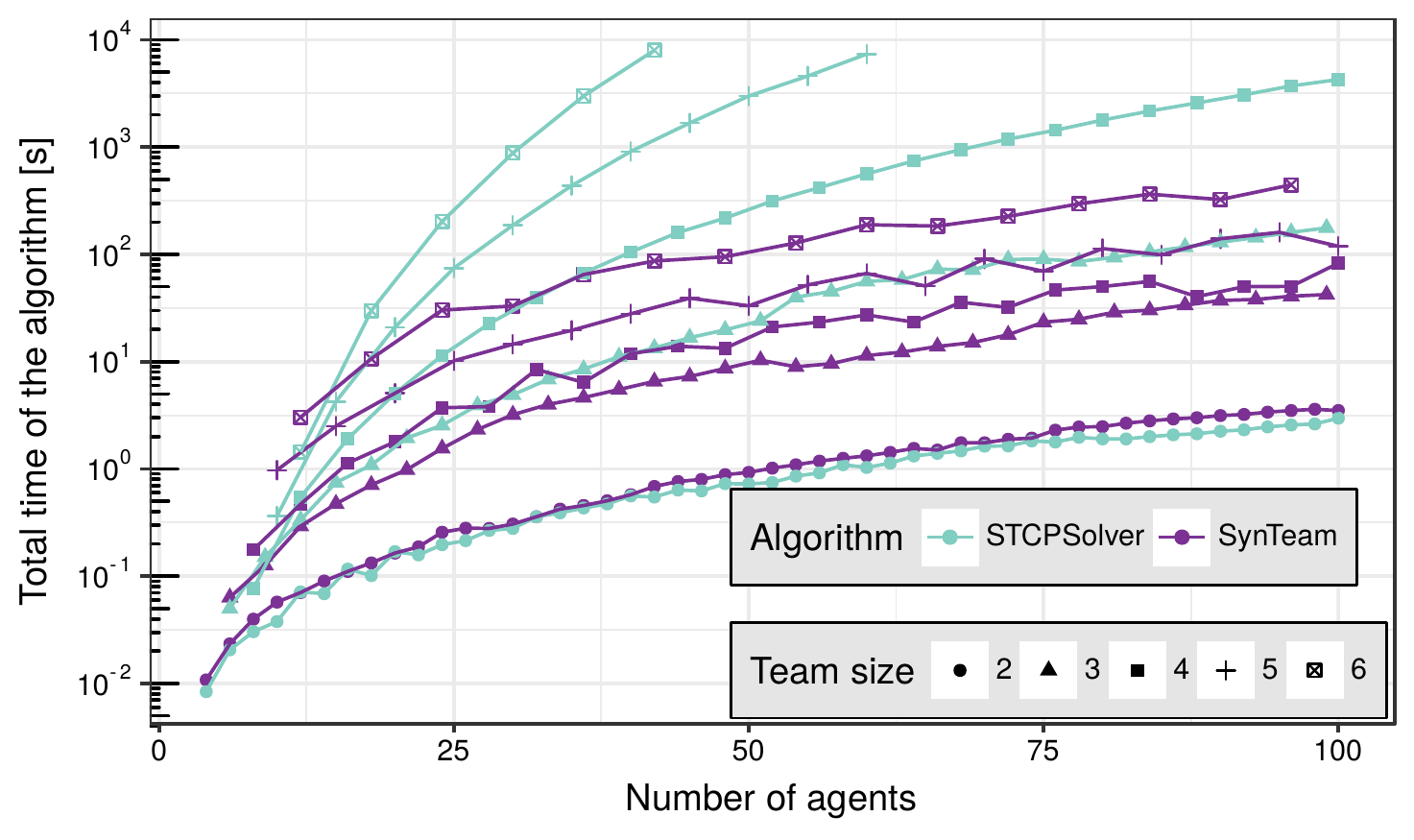}
		\caption{\texttt{English}, $\lambda = 0.2$.}
		\label{figure:TotalTimes:english:lambda02}
	\end{subfigure}
	\begin{subfigure}{0.49\textwidth}
		\includegraphics[width=\linewidth]{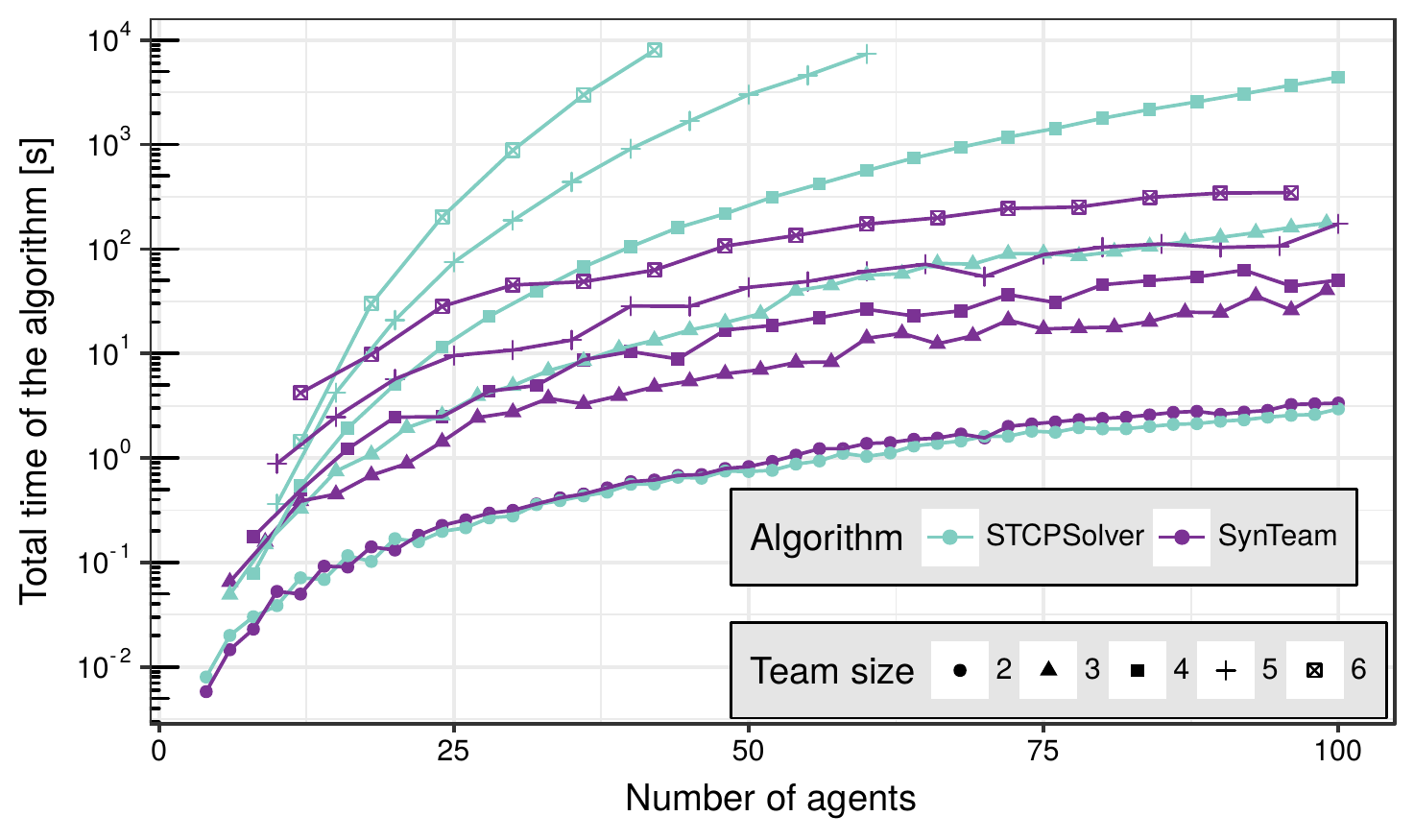}
		\caption{\texttt{English}, $\lambda = 0.8$.}
		\label{figure:TotalTimes:english:lambda08}
	\end{subfigure}
    \caption{Total time needed by the two techniques.}
    \label{figure:TotalTimes}
\end{figure}

\noindent
{\bf Runtime Analysis.}\label{ssec:runtime}
The graphics in Figure~\ref{figure:TotalTimes} show the performance of the algorithms in terms of the total running time (in seconds). Each graphic is dedicated to the results concerning one of the task types and one of the two considered settings for $\lambda$. Each data point represents the average over 20 random sets of students of size $n$. The total running time of \cplex\ consists of two components: the data generation time and the computation time. Hereby, the data generation time measures the time for generating all possible teams and calculating their synergistic value (lines 1-5 in Algorithm \ref{alg:optimalSolving}), while the computation time measures the time needed by CPLEX to load the corresponding ILP model and solve it to optimality. The graphics in Figure~\ref{figure:TotalTimes} show that, as the team size ($m$) increases, the total time needed by \cplex\ becomes prohibitively costly. In fact, the whole range of results could only be generated up to a team size of $m=4$. For $m=5$ the calculations were stopped after a total number of $60$ students, and for $m=6$ after $42$ students. This was done because for larger values of $n$ and $m$, the size of the corresponding ILP models was simply too large for CPLEX.\footnote{For instance, the ILP model for $n=48$ and $m=6$ consists of $12.271.512$ binary variables.} In general, it can be observed that the runtime of \cplex\ dramatically increases with the number of students ($n$) and team size ($m$). Note that for a team size of $m=6$ and for $n=42$ students, \SynTeam\ is at least two orders of magnitude faster than \cplex. On the other side, for $m=2$ the total time requirements of the two techniques is nearly equal. \\ 

\begin{figure}[p] 
\centering
	\begin{subfigure}{0.49\textwidth}
		\includegraphics[width=\linewidth]{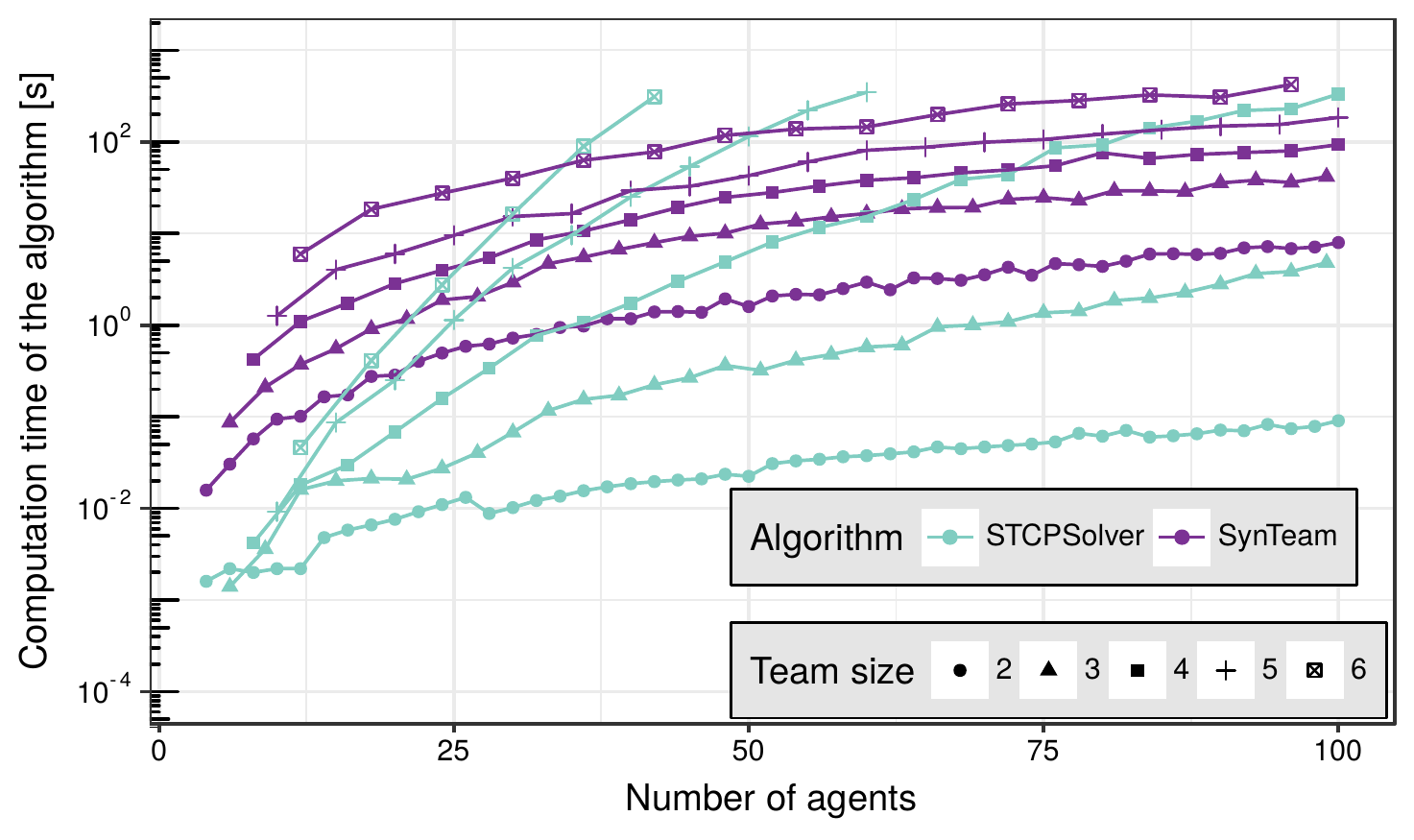}
		\caption{\texttt{body rythm}, $\lambda = 0.2$.}
		\label{figure:SecondTimes:body_rythm:lambda02}
	\end{subfigure}
	\begin{subfigure}{0.49\textwidth}
		\includegraphics[width=\linewidth]{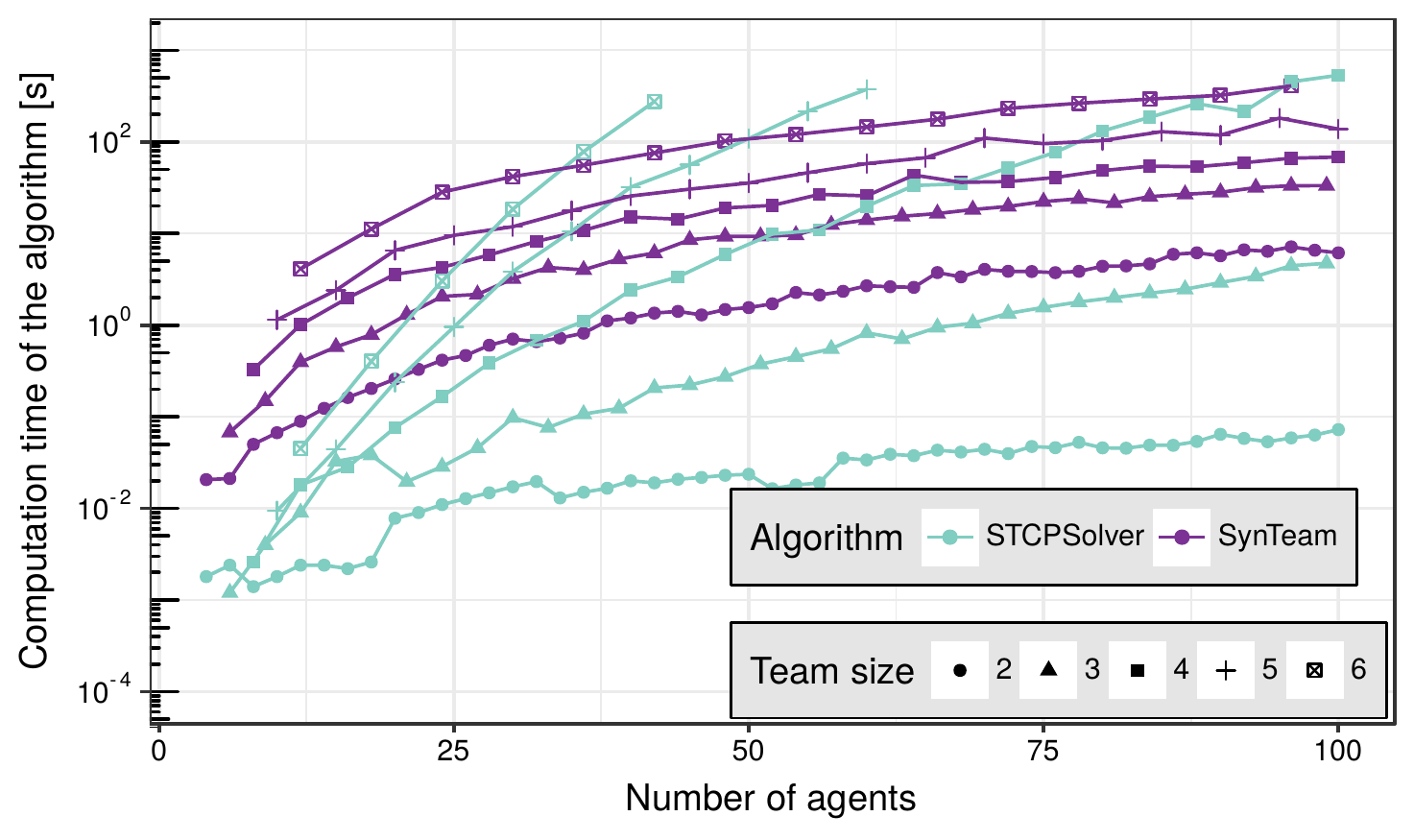}
		\caption{\texttt{body rythm}, $\lambda = 0.8$.}
		\label{figure:SecondTimes:body_rythm:lambda08}
	\end{subfigure}
	\begin{subfigure}{0.49\textwidth}
		\includegraphics[width=\linewidth]{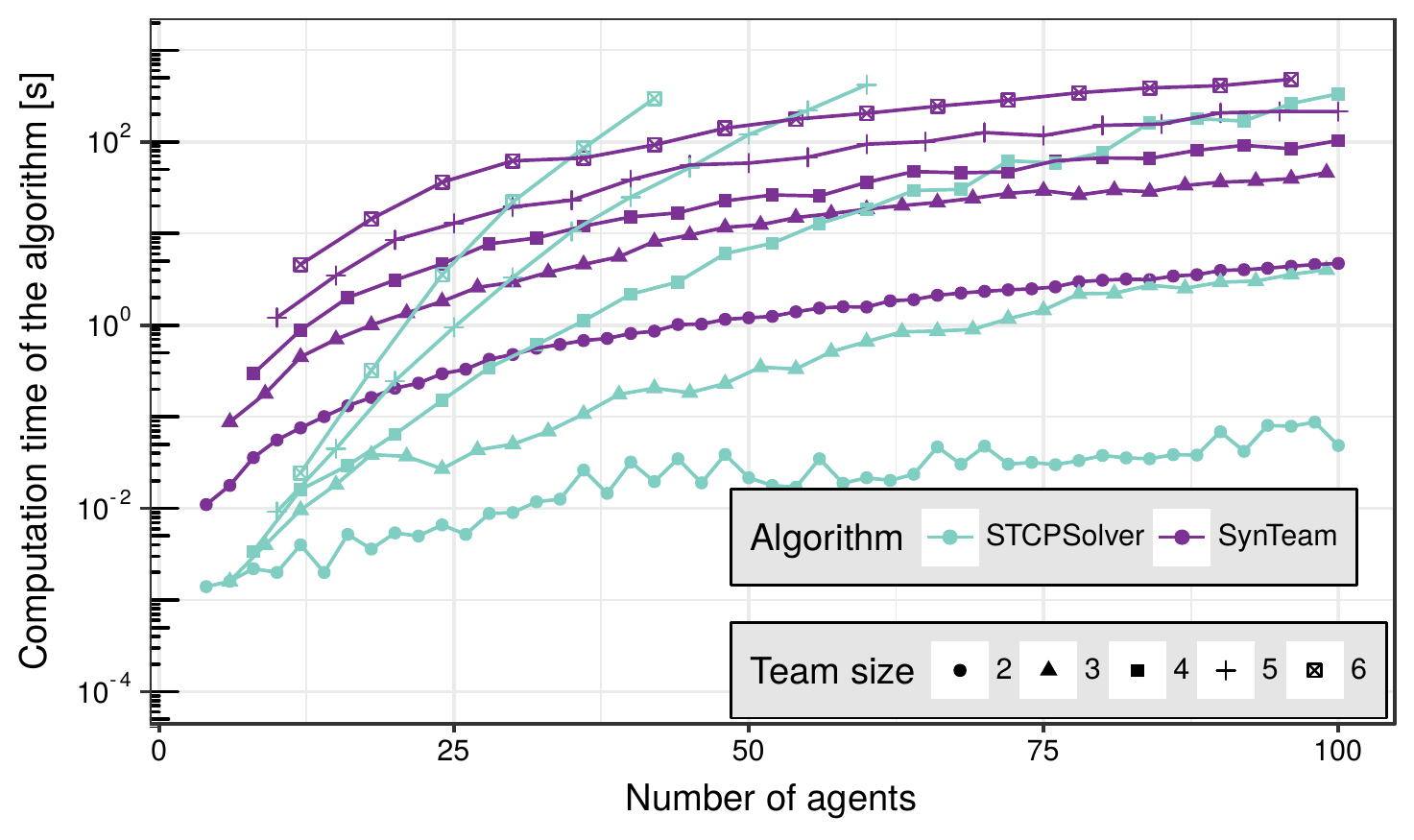}
		\caption{\texttt{entrepreneur}, $\lambda = 0.2$.}
		\label{figure:SecondTimes:entrepreneur:lambda02}
	\end{subfigure}
	\begin{subfigure}{0.49\textwidth}
		\includegraphics[width=\linewidth]{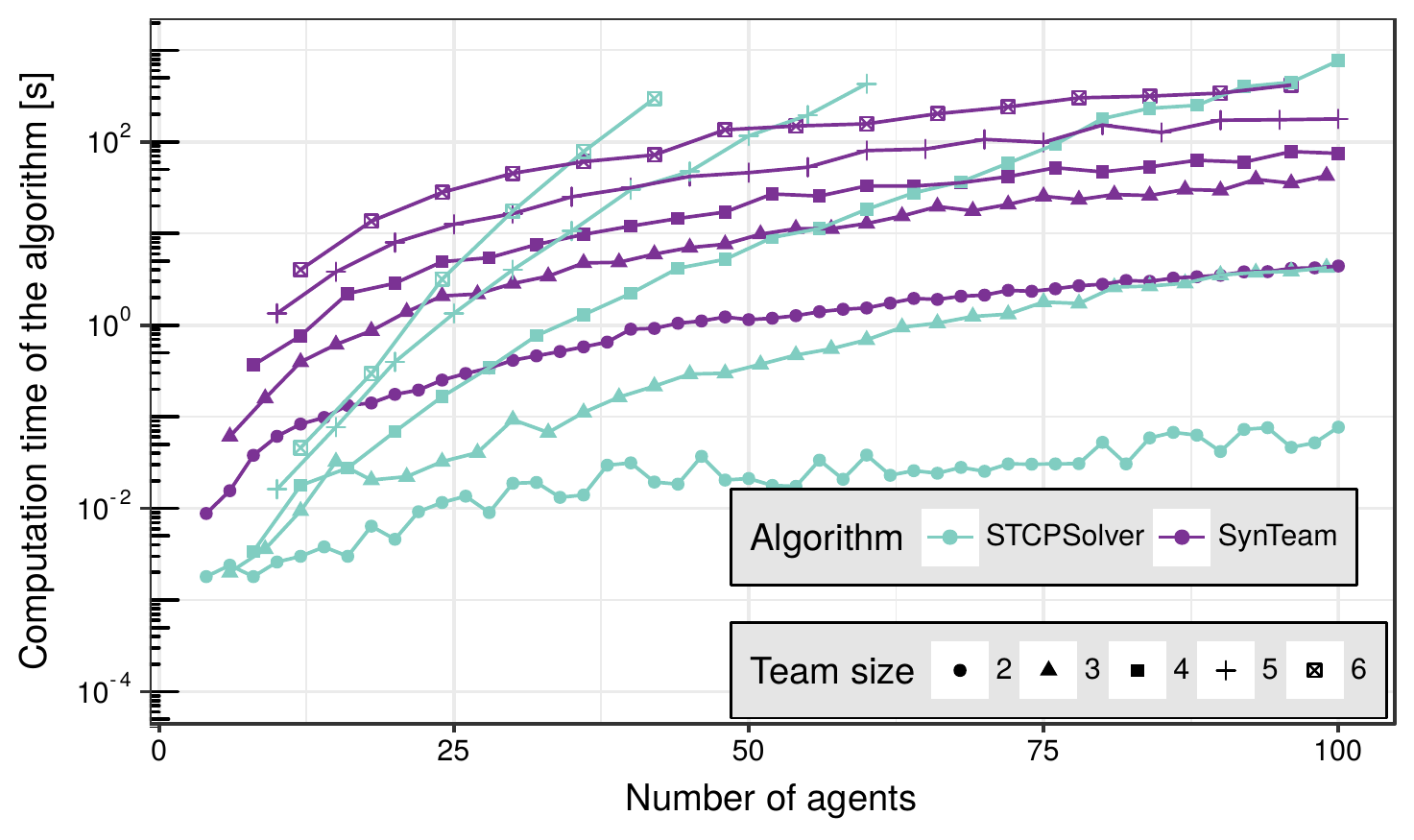}
		\caption{\texttt{entrepreneur}, $\lambda = 0.8$.}
		\label{figure:SecondTimes:entrepreneur:lambda08}
	\end{subfigure}
	\begin{subfigure}{0.49\textwidth}
		\includegraphics[width=\linewidth]{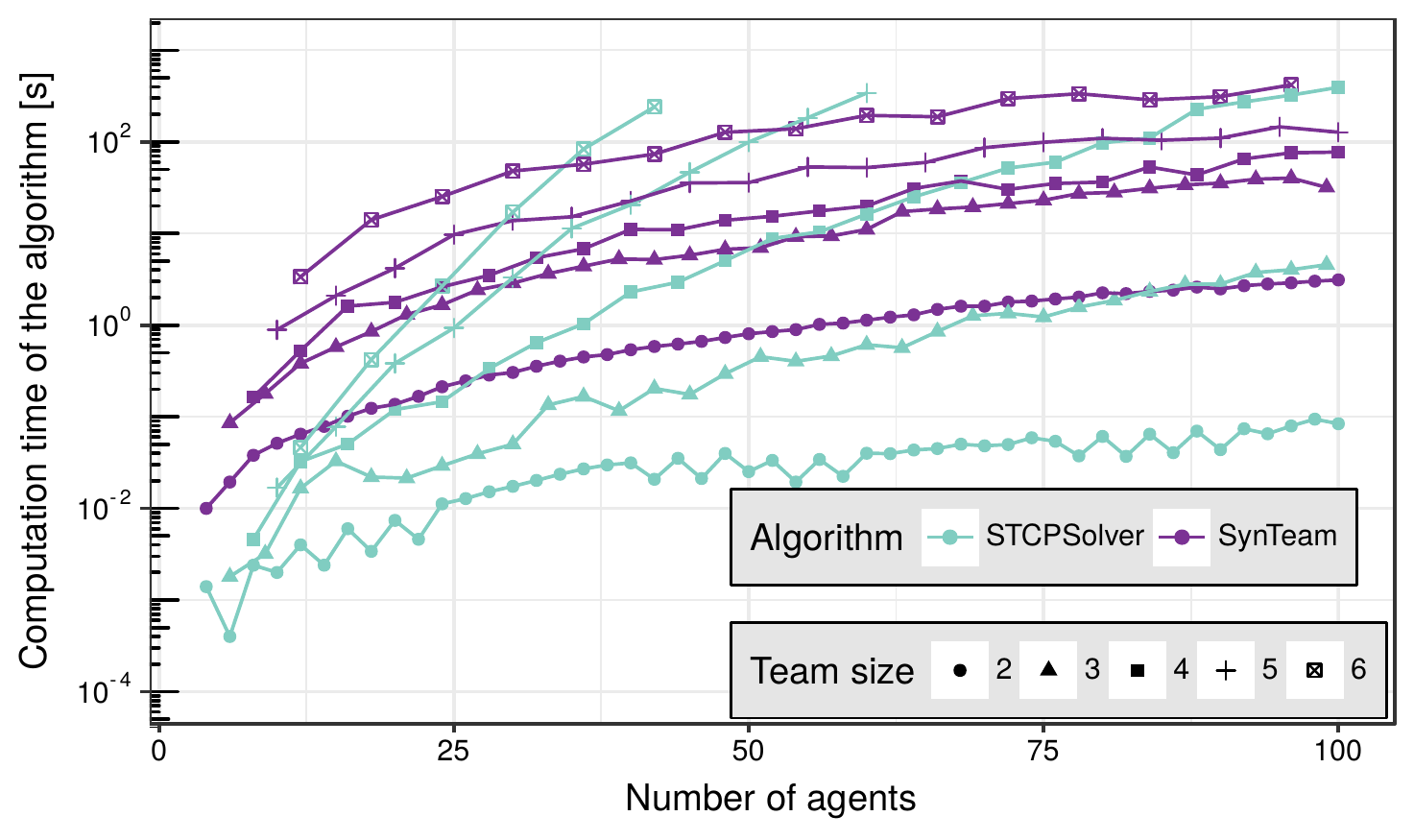}
		\caption{\texttt{arts design}, $\lambda = 0.2$.}
		\label{figure:SecondTimes:arts_design:lambda02}
	\end{subfigure}
	\begin{subfigure}{0.49\textwidth}
		\includegraphics[width=\linewidth]{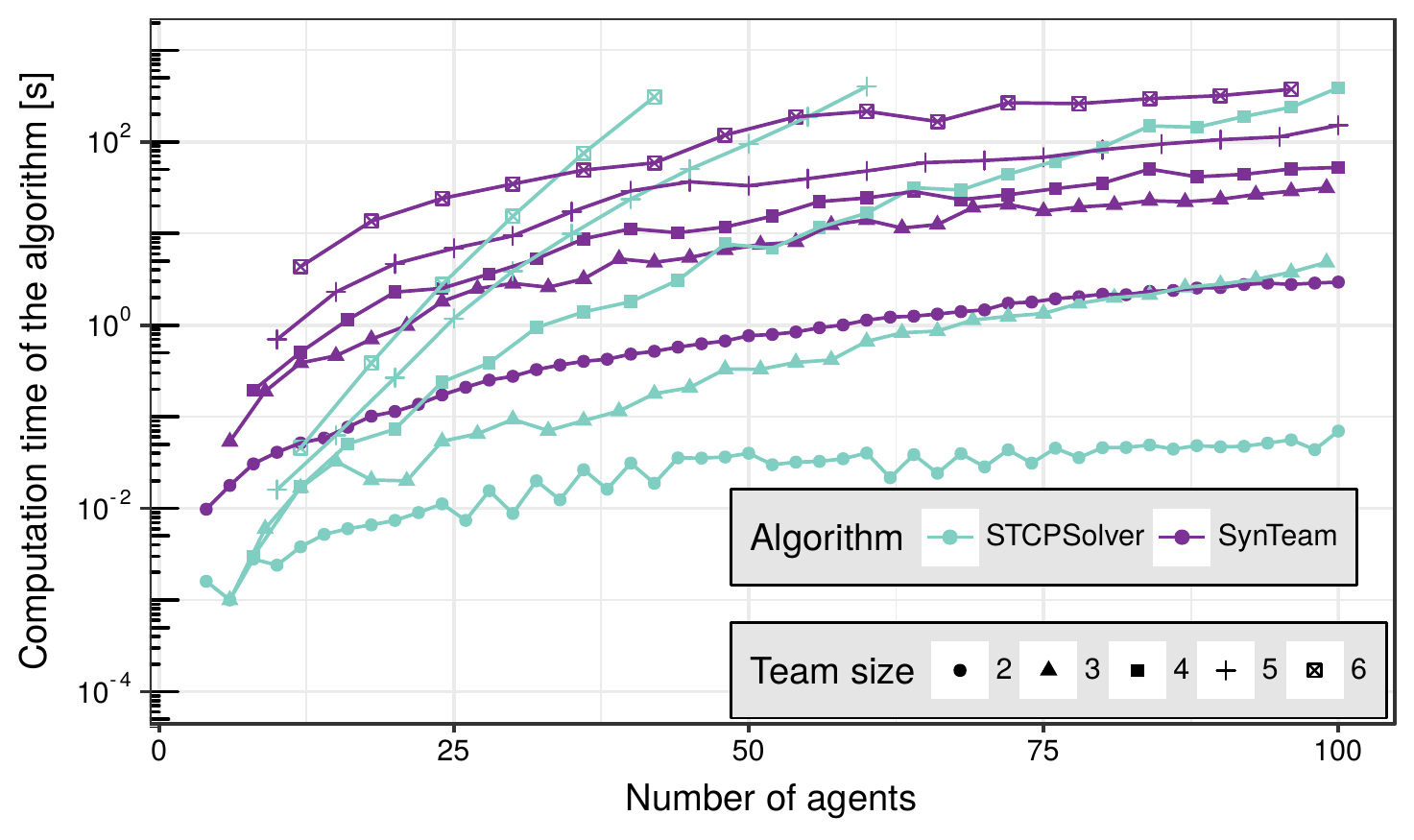}
		\caption{\texttt{arts design}, $\lambda = 0.8$.}
		\label{figure:SecondTimes:arts_design:lambda08}
	\end{subfigure}
	\begin{subfigure}{0.49\textwidth}
		\includegraphics[width=\linewidth]{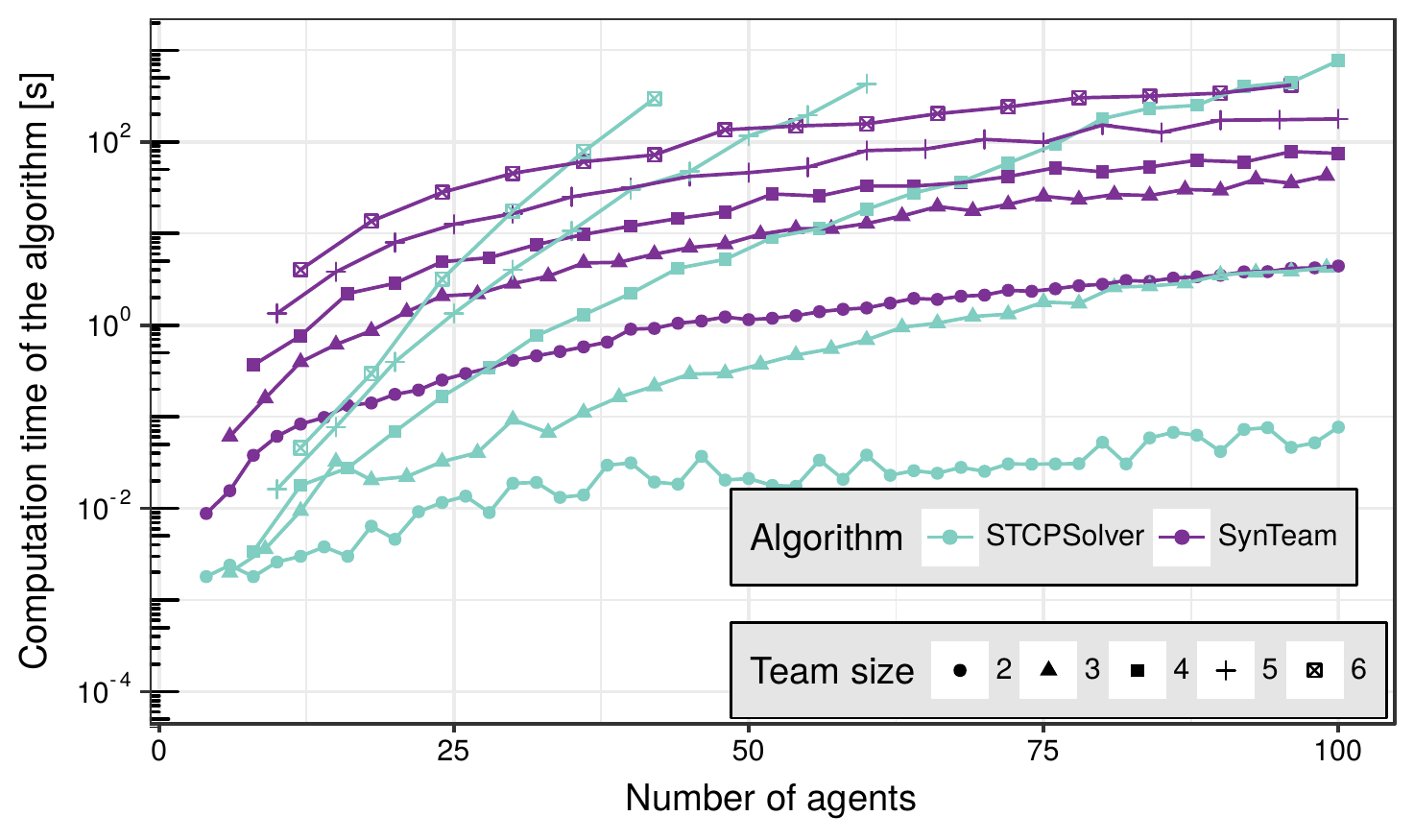}
		\caption{\texttt{English}, $\lambda = 0.2$.}
		\label{figure:SecondTimes:english:lambda02}
	\end{subfigure}
	\begin{subfigure}{0.49\textwidth}
		\includegraphics[width=\linewidth]{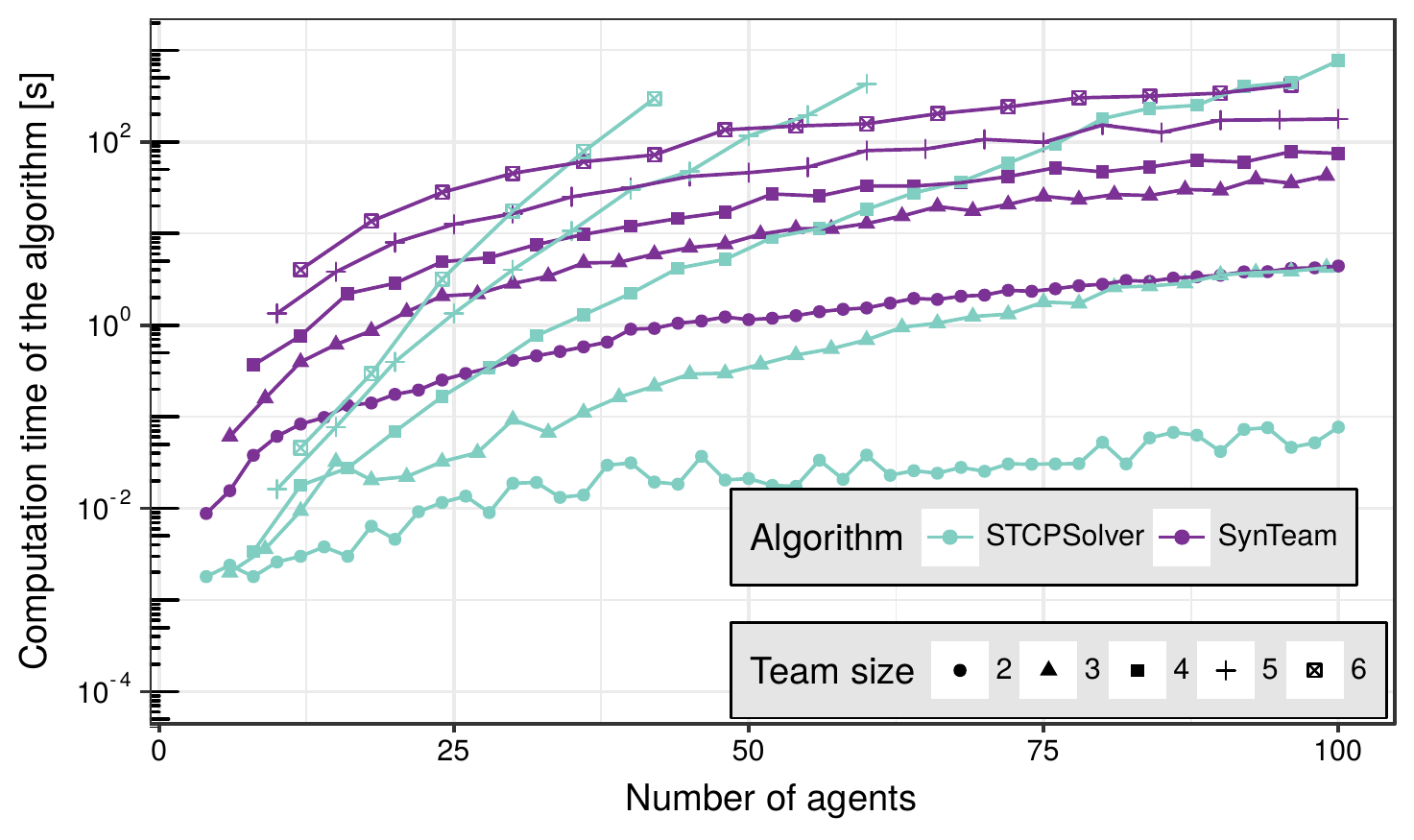}
		\caption{\texttt{English}, $\lambda = 0.8$.}
		\label{figure:SecondTimes:english:lambda08}
	\end{subfigure}
    \caption{Computation time needed by the two techniques (disregarding the data generation time of \cplex).}
    \label{figure:SecondTimes}
\end{figure}

To better understand this result, we compared the computation times of \cplex\ (disregarding its data generation time) with the computation times of \SynTeam. Figure \ref{figure:SecondTimes} provides this comparison in a graphical way. It can be observed that ---even in this case--- \SynTeam\ is more efficient for larger instances, that is, for team sizes of $m > 3$ and a growing number of students. \\


\begin{figure}[p] 
\centering
	\begin{subfigure}{0.49\textwidth}
		\includegraphics[width=\linewidth]{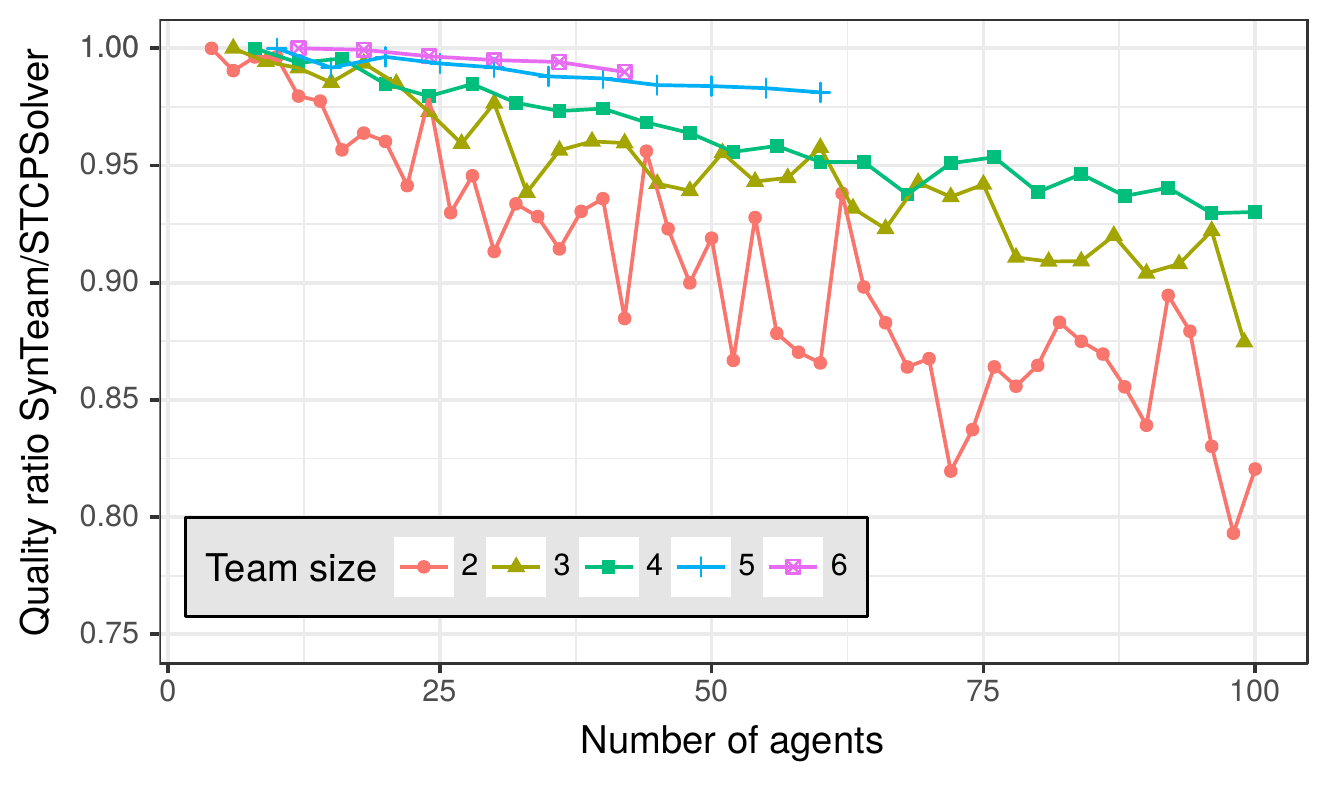}
		\caption{\texttt{body rythm}, $\lambda = 0.2$.}
		\label{figure:qRatio:body_rythm:lambda02}
	\end{subfigure}
	\begin{subfigure}{0.49\textwidth}
		\includegraphics[width=\linewidth]{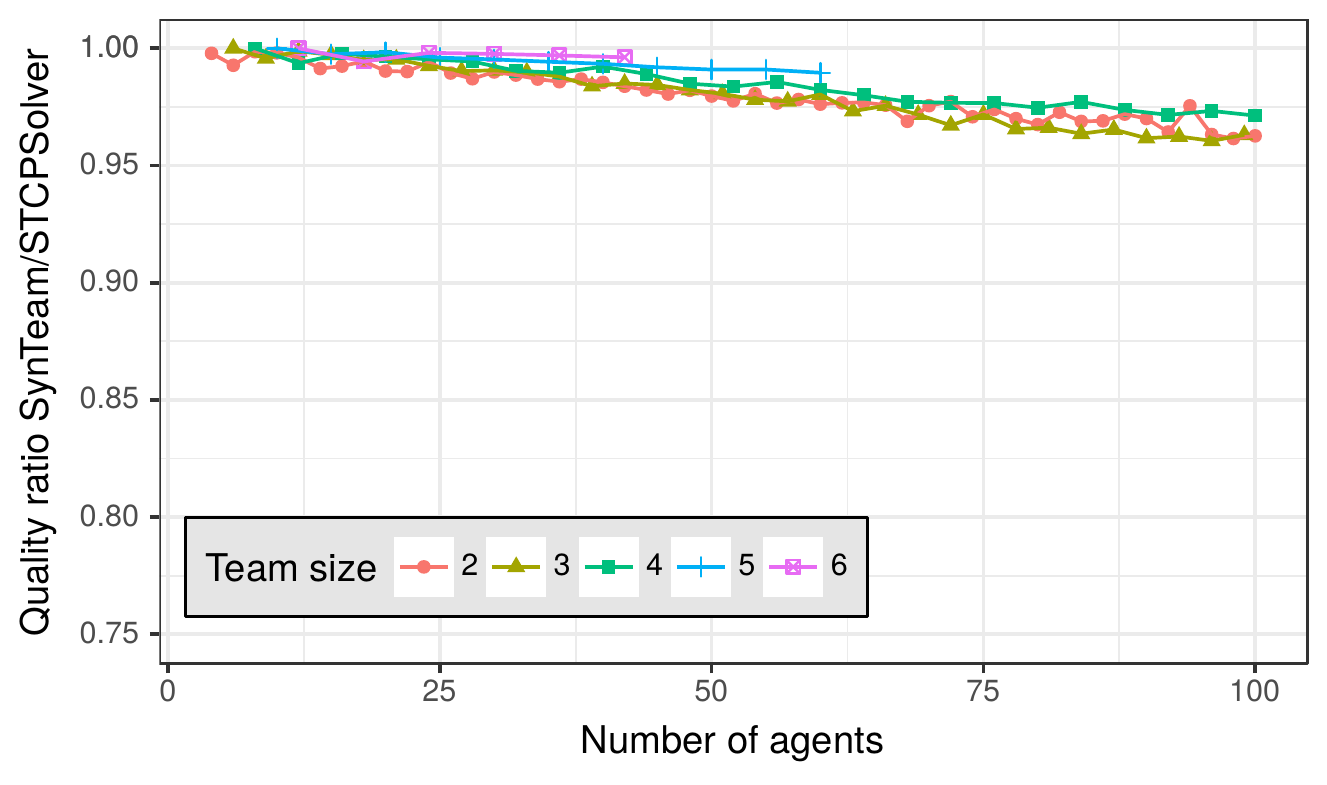}
		\caption{\texttt{body rythm}, $\lambda = 0.8$.}
		\label{figure:qRatio:body_rythm:lambda08}
	\end{subfigure}
	\begin{subfigure}{0.49\textwidth}
		\includegraphics[width=\linewidth]{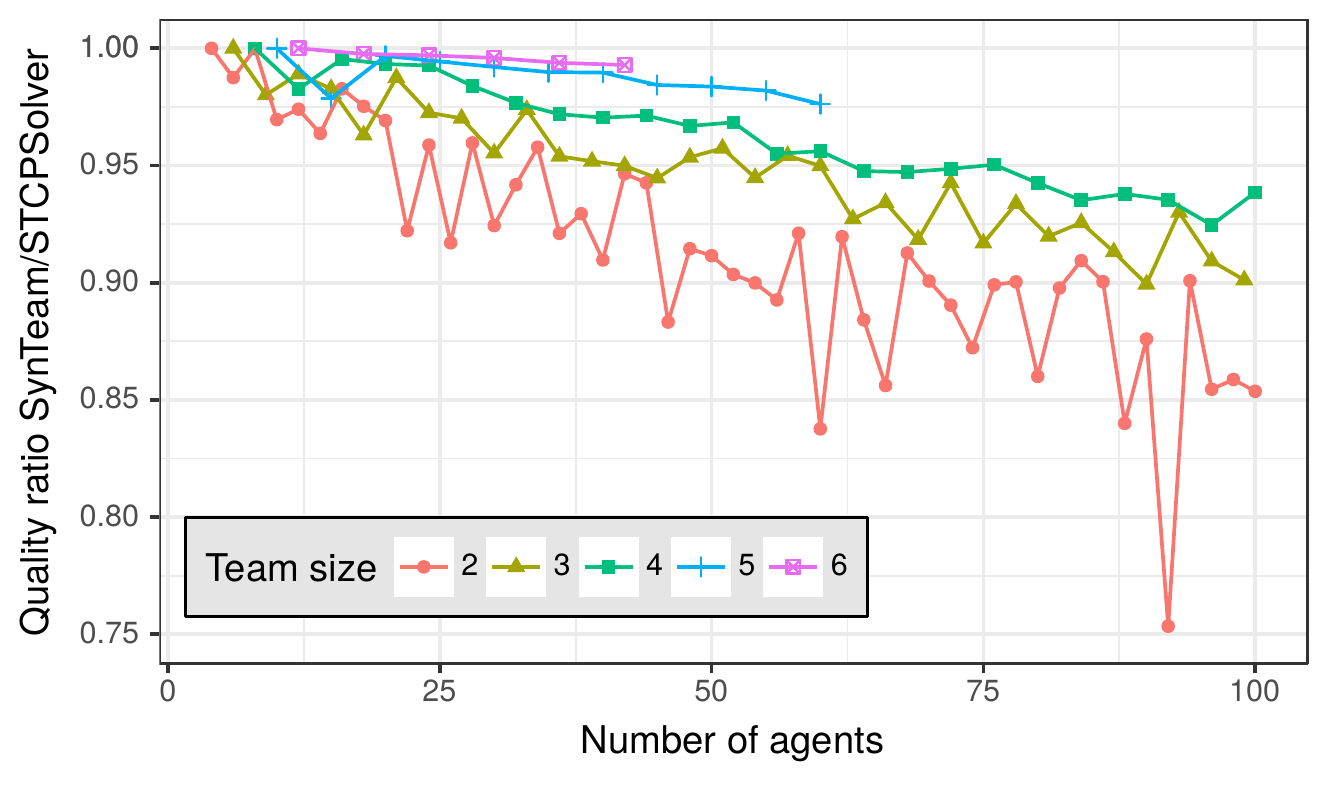}
		\caption{\texttt{entrepreneur}, $\lambda = 0.2$.}
		\label{figure:qRatio:entrepreneur:lambda02}
	\end{subfigure}
	\begin{subfigure}{0.49\textwidth}
		\includegraphics[width=\linewidth]{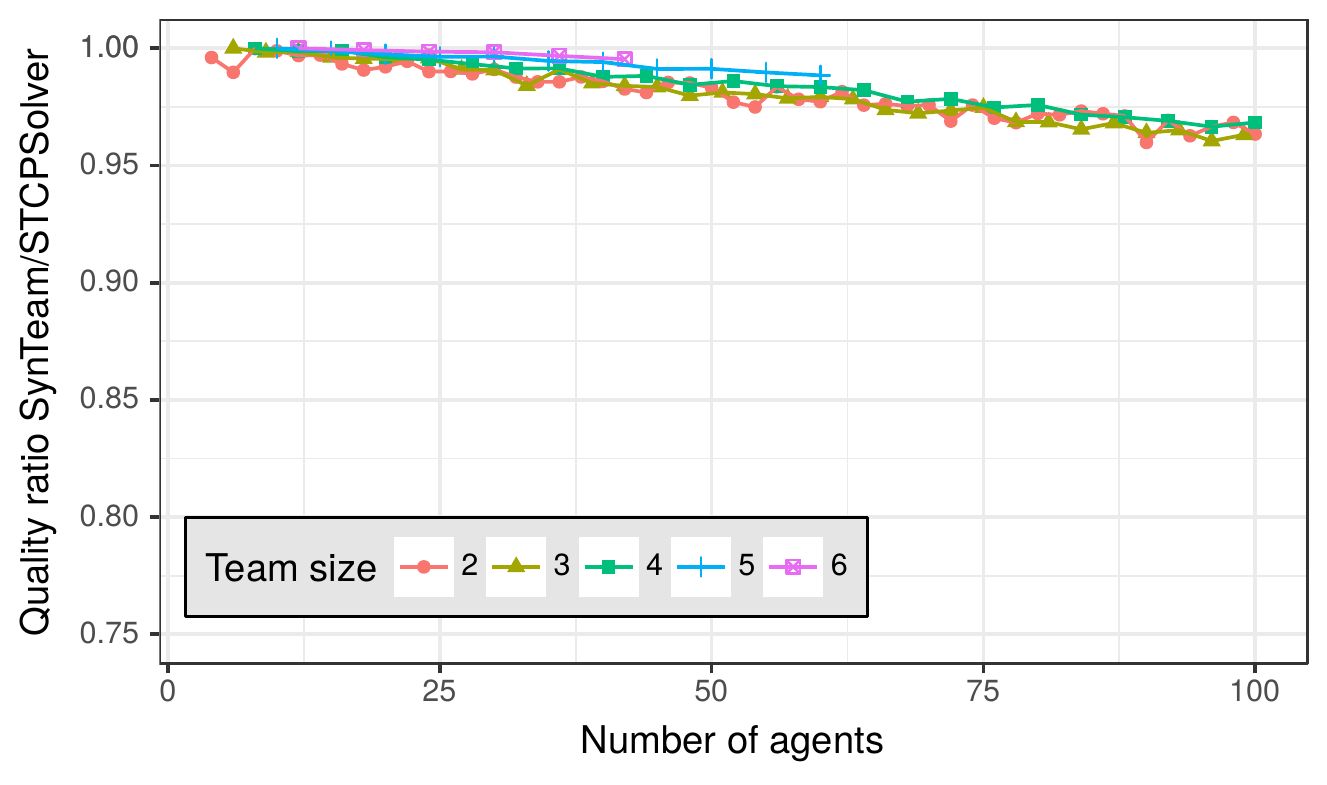}
		\caption{\texttt{entrepreneur}, $\lambda = 0.8$.}
		\label{figure:qRatio:entrepreneur:lambda08}
	\end{subfigure}
	\begin{subfigure}{0.49\textwidth}
		\includegraphics[width=\linewidth]{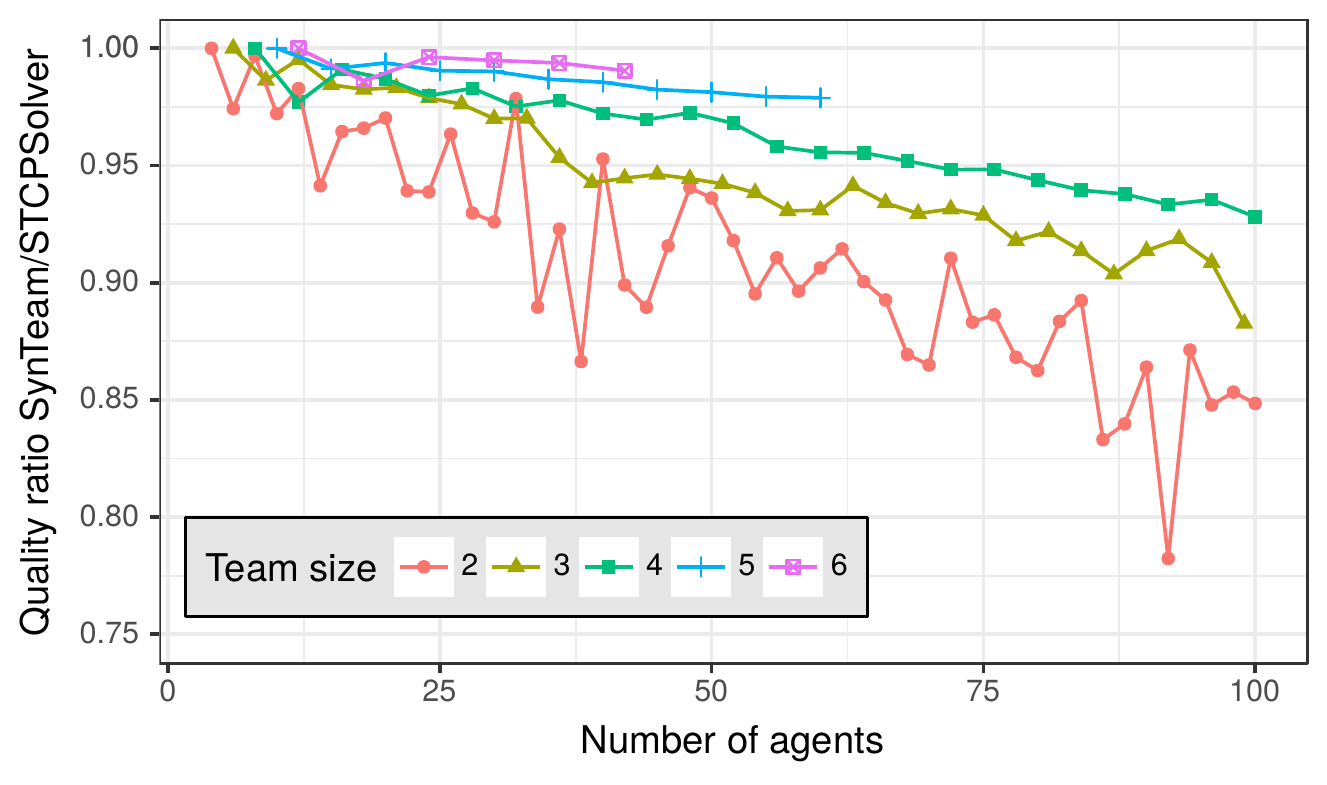}
		\caption{\texttt{arts design}, $\lambda = 0.2$.}
		\label{figure:qRatio:arts_design:lambda02}
	\end{subfigure}
	\begin{subfigure}{0.49\textwidth}
		\includegraphics[width=\linewidth]{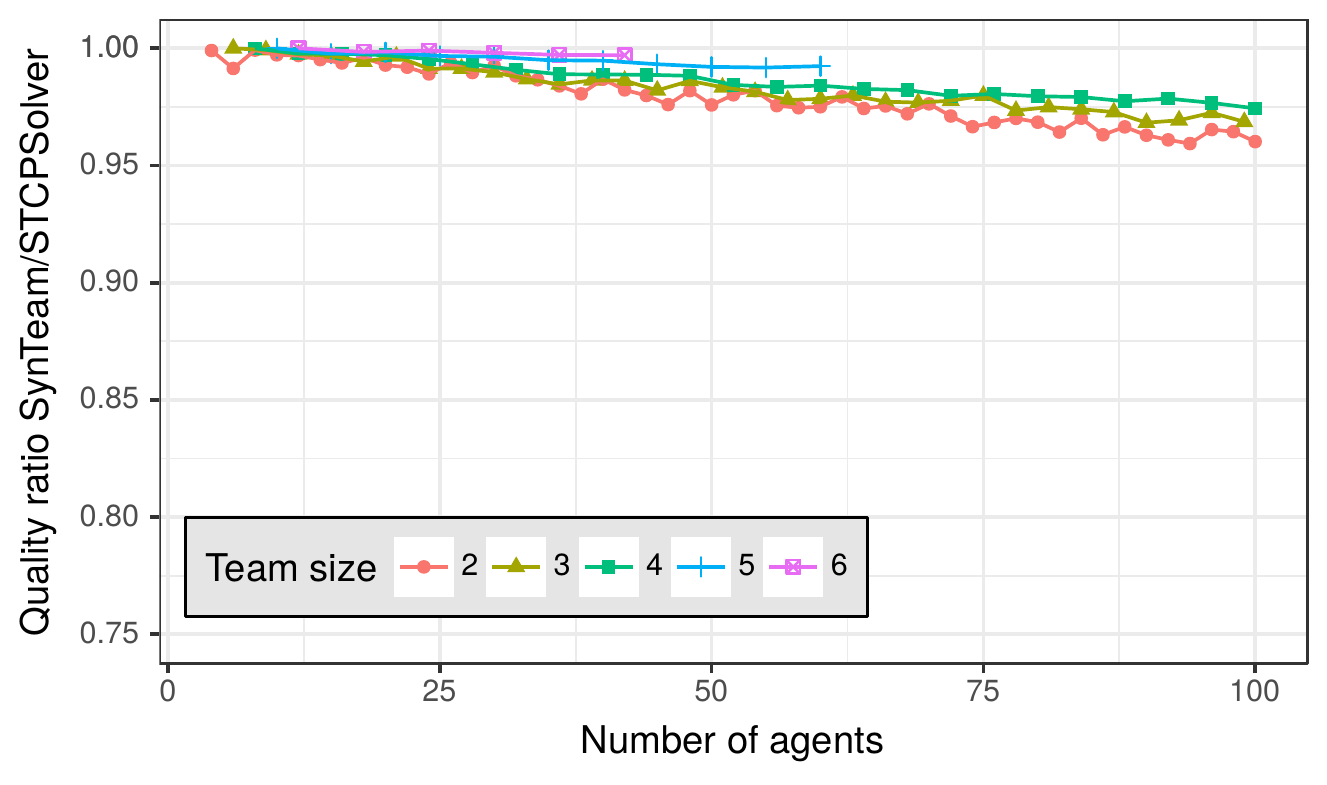}
		\caption{\texttt{arts design}, $\lambda = 0.8$.}
		\label{figure:qRatio:arts_design:lambda08}
	\end{subfigure}
	\begin{subfigure}{0.49\textwidth}
		\includegraphics[width=\linewidth]{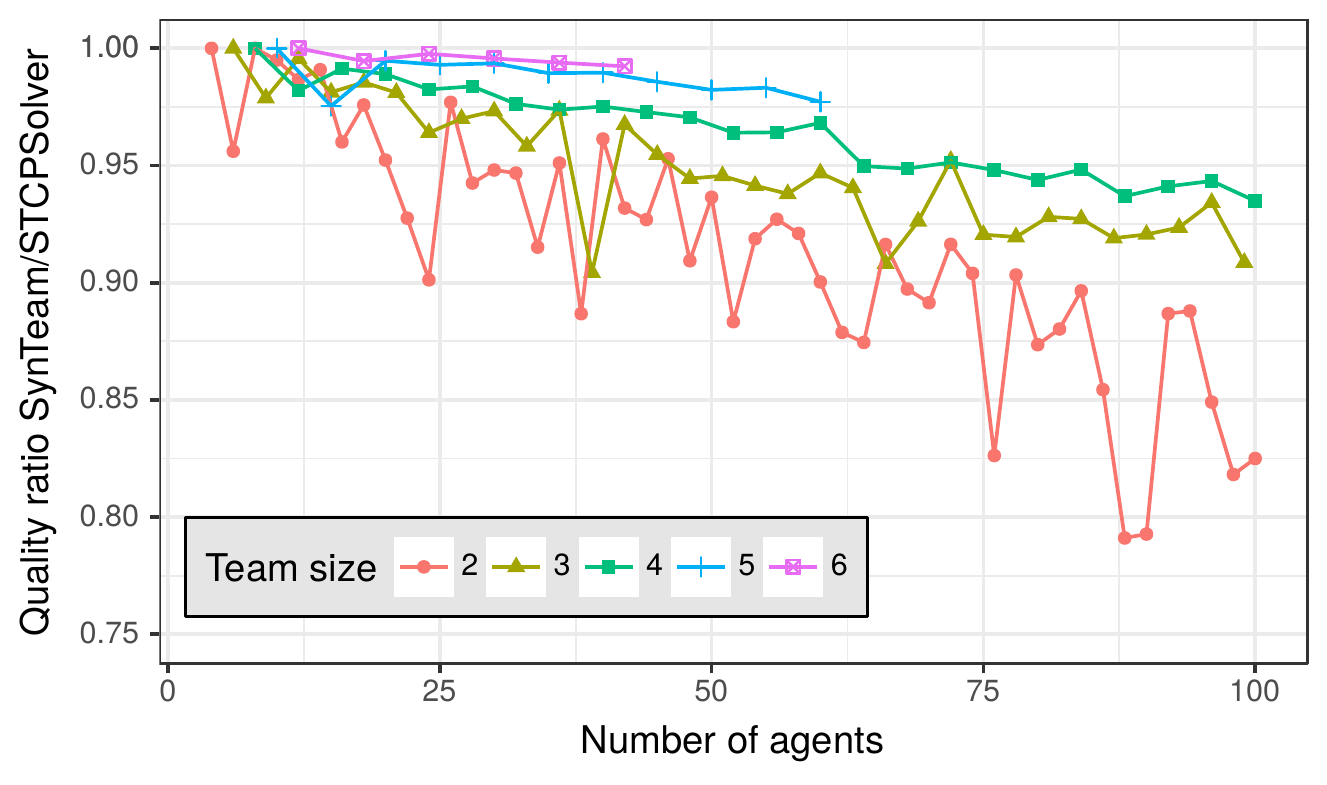}
		\caption{\texttt{English}, $\lambda = 0.2$.}
		\label{figure:qRatio:english:lambda02}
	\end{subfigure}
	\begin{subfigure}{0.49\textwidth}
		\includegraphics[width=\linewidth]{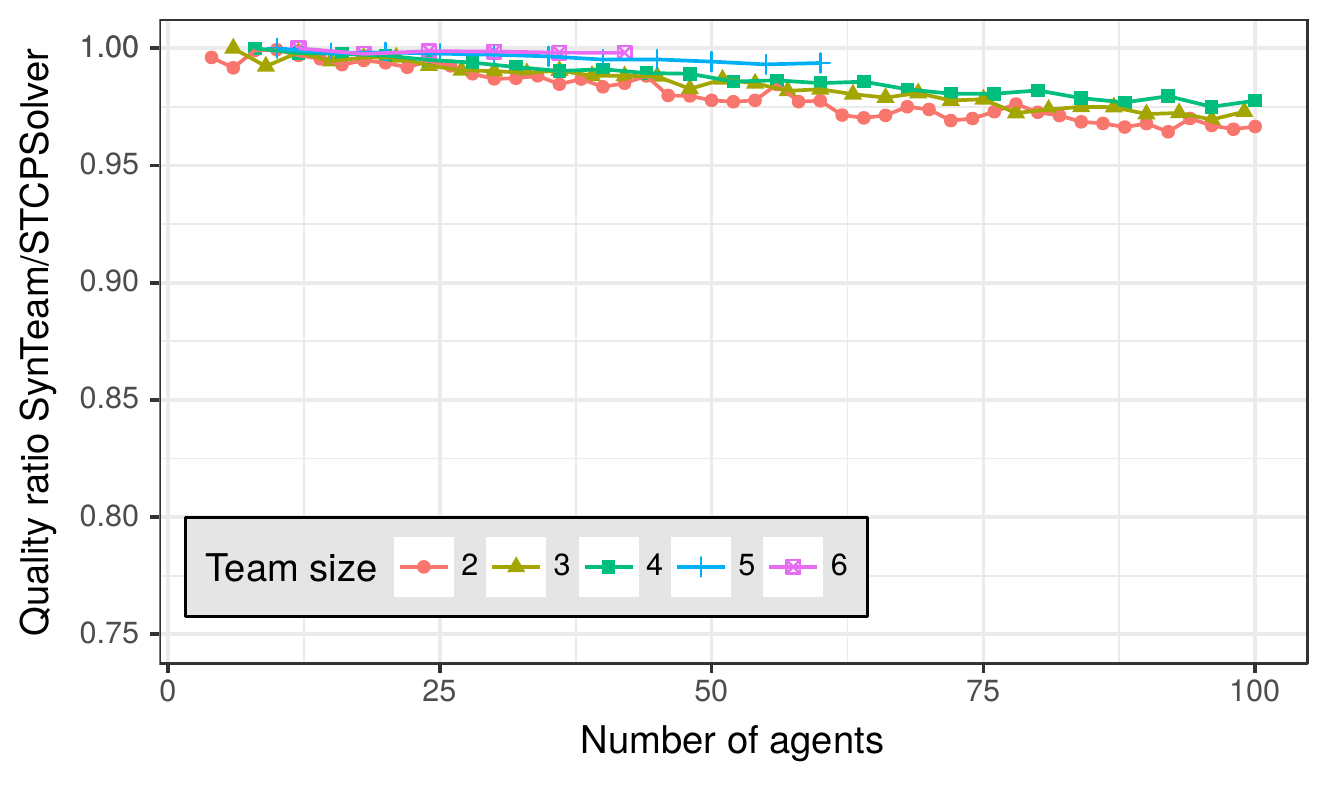}
		\caption{\texttt{English}, $\lambda = 0.8$.}
		\label{figure:qRatio:english:lambda08}
	\end{subfigure}
    \caption{\SynTeam\ relative quality plots.}
    \label{figure:qRatio}
\end{figure}

\noindent {\bf Quality Analysis.} 
The relative quality of the best solutions provided by \SynTeam\ are shown in Figure~\ref{figure:qRatio}. Note that the relative quality of \SynTeam\ is obtained by dividing the value of the best solution provided by \SynTeam\ by the value of an optimal solution calculated by \cplex. The following observations can be made. First, in all cases the relative quality of \SynTeam\ seems to decrease with decreasing team size. However, even in the case of the smallest considered team size ($m=2$), the quality ratio of \SynTeam\ is always above $95\%$ in the context of the experiments with $\lambda=0.8$, and above $75\%$ in the case of $\lambda=0.2$. The apparent difference in the quality ratios between $\lambda=0.2$ and $\lambda=0.8$ can be explained as follows. After an in-depth study, we noticed that the proficiency values of the different teams are nearly all from the range $[0.9,1.0]$, while the congeniality values are spread over the whole range $[0,1]$. Therefore, when preferring proficiency over congeniality ($\lambda=0.8$), there is a large concentration of teams with synergistic values of a rather high relative quality. This is not the case when $\lambda=0.2$. \\

\begin{figure}[p] 
\centering
	\begin{subfigure}{0.49\textwidth}
		\includegraphics[width=\linewidth]{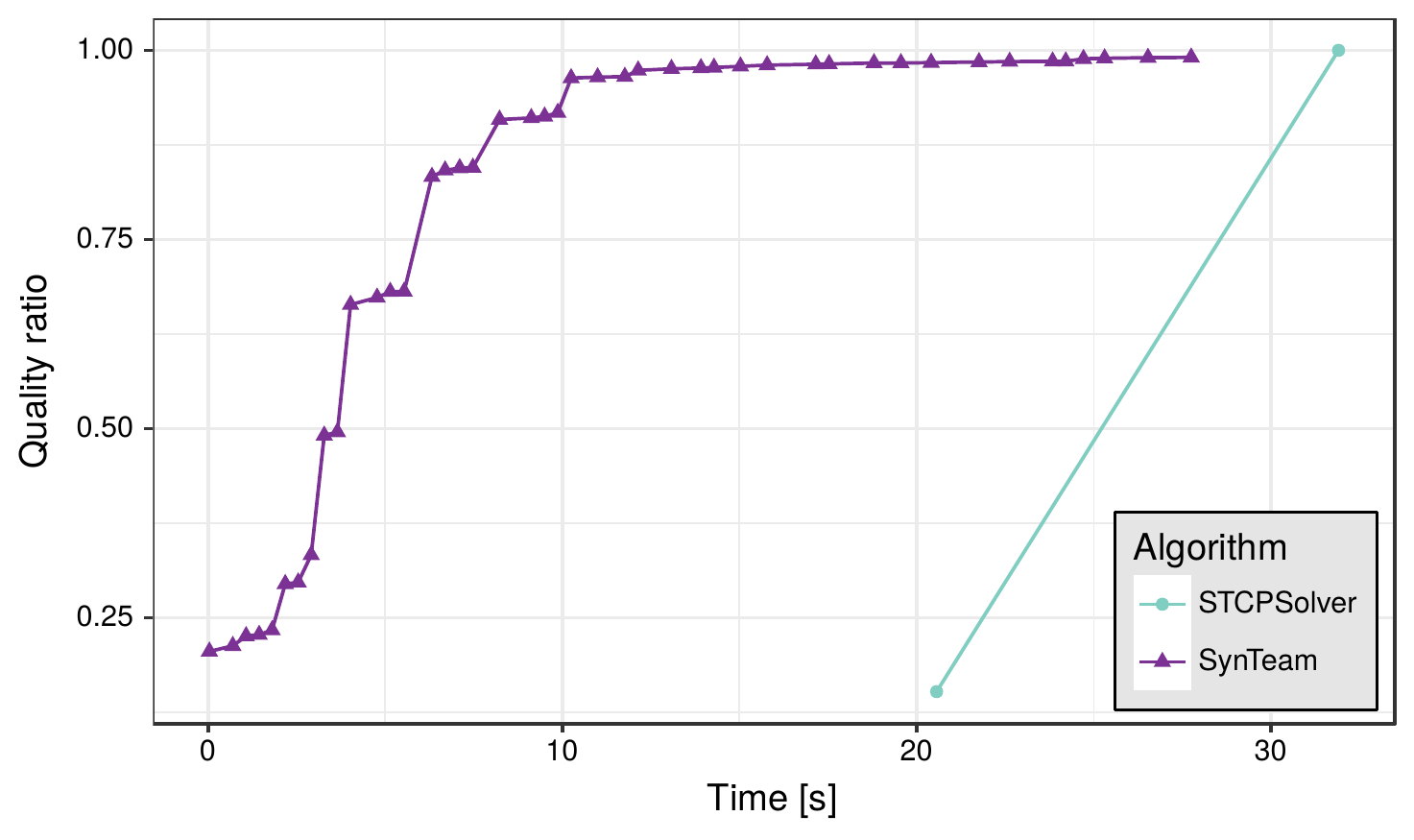}
		\caption{\texttt{body rythm}, $\lambda = 0.2$.}
		\label{figure:qTime:body_rythm:lambda02}
	\end{subfigure}
	\begin{subfigure}{0.49\textwidth}
		\includegraphics[width=\linewidth]{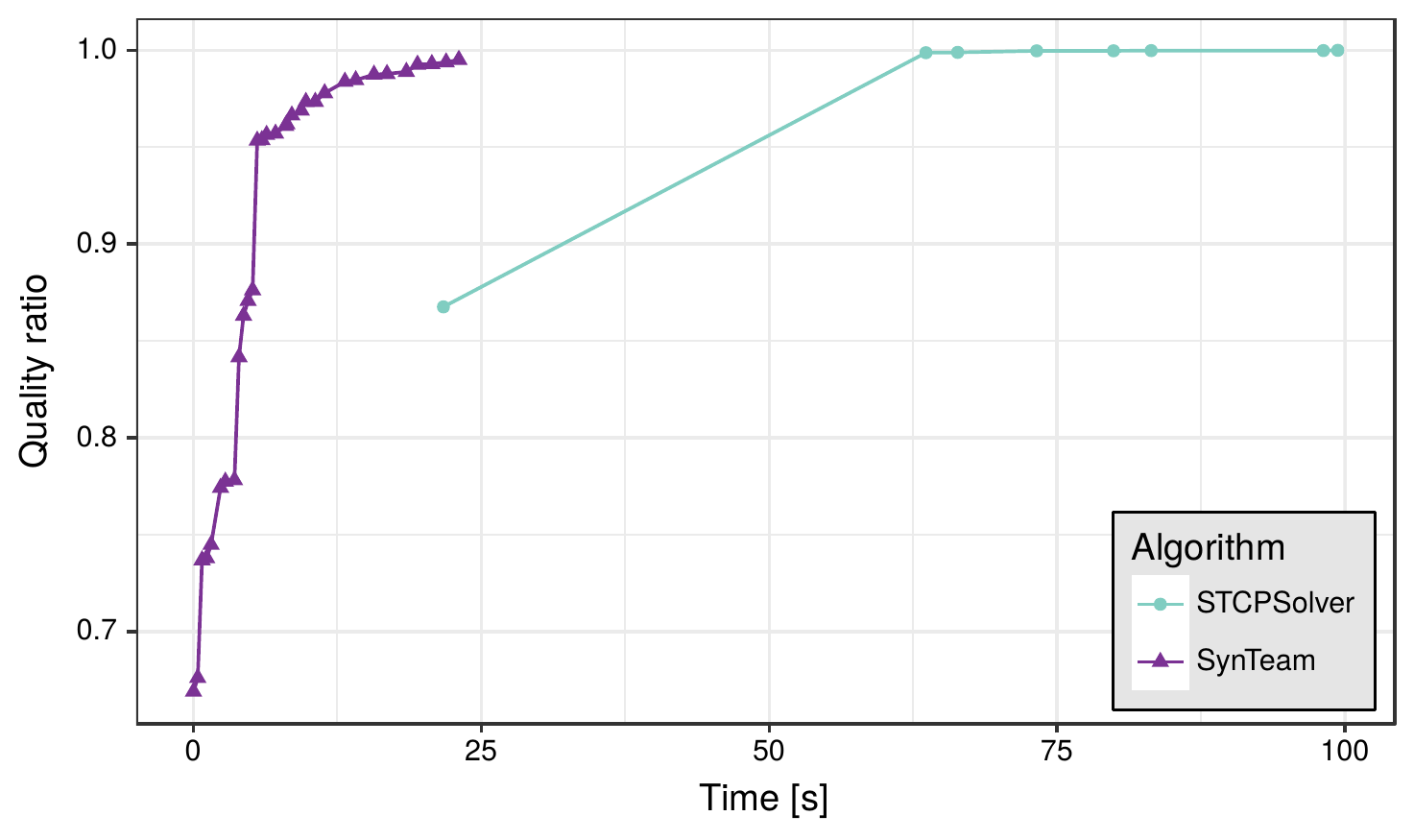}
		\caption{\texttt{body rythm}, $\lambda = 0.8$.}
		\label{figure:qTime:body_rythm:lambda08}
	\end{subfigure}
	\begin{subfigure}{0.49\textwidth}
		\includegraphics[width=\linewidth]{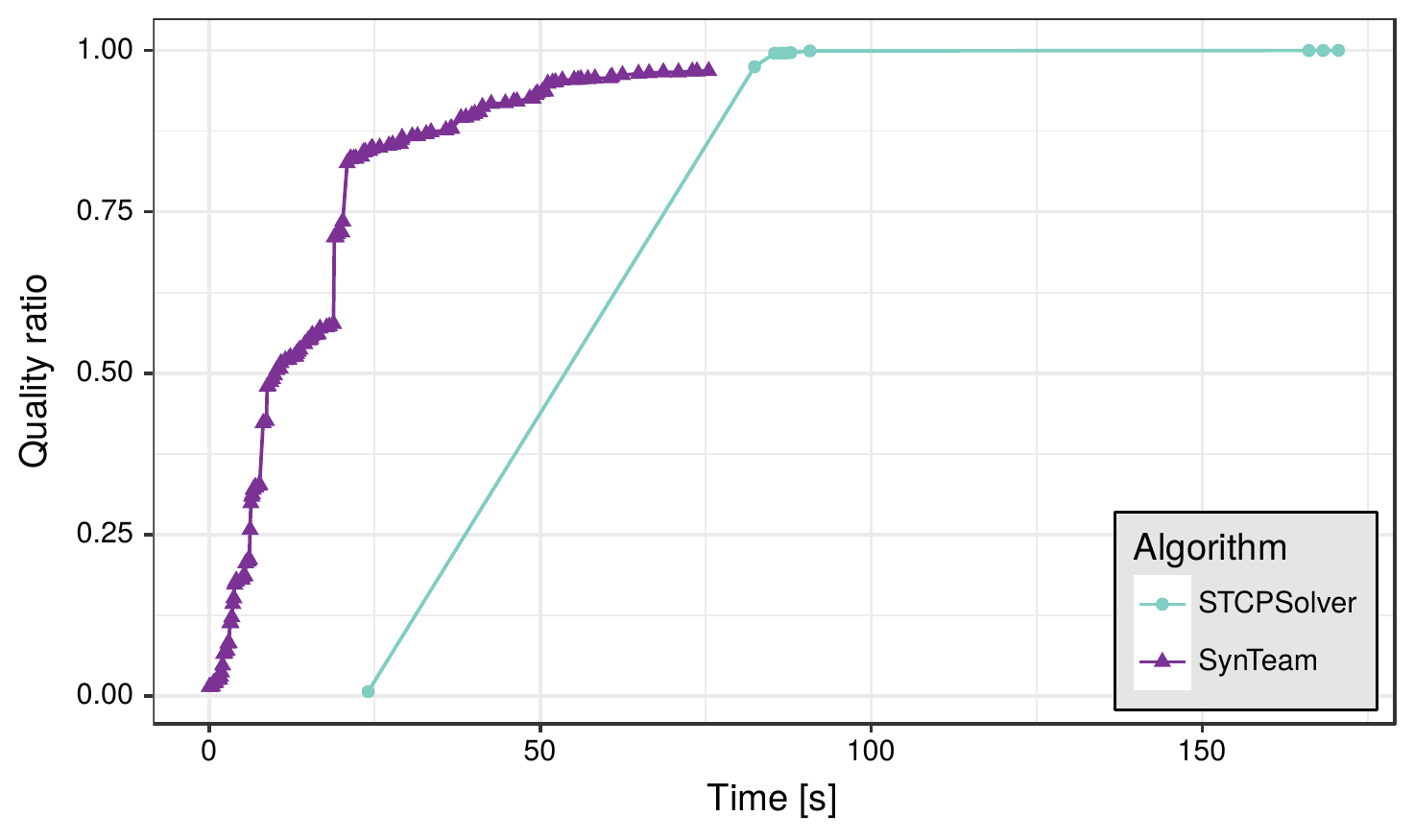}
		\caption{\texttt{entrepreneur}, $\lambda = 0.2$.}
		\label{figure:qTime:entrepreneur:lambda02}
	\end{subfigure}
	\begin{subfigure}{0.49\textwidth}
		\includegraphics[width=\linewidth]{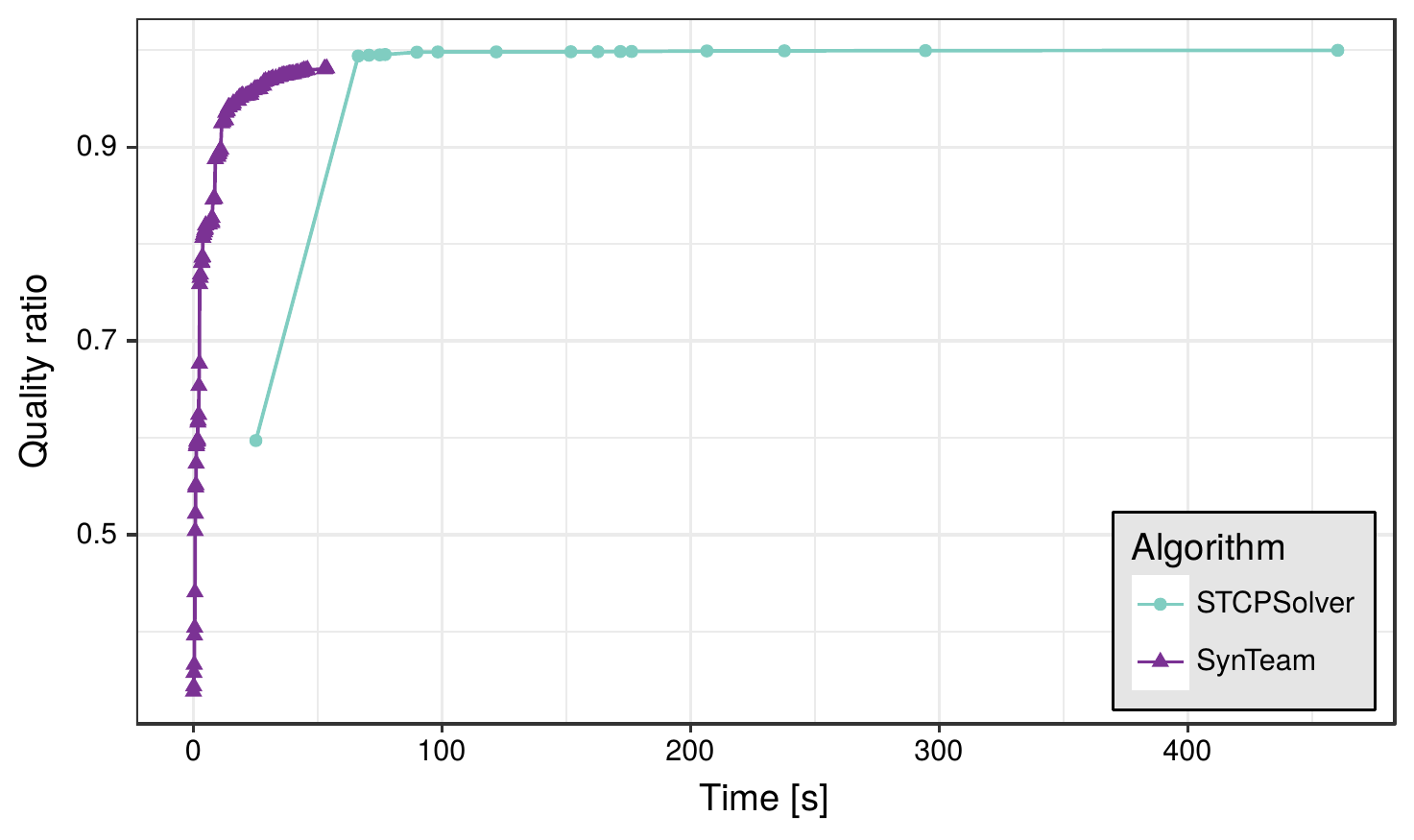}
		\caption{\texttt{entrepreneur}, $\lambda = 0.8$.}
		\label{figure:qTime:entrepreneur:lambda08}
	\end{subfigure}
	\begin{subfigure}{0.49\textwidth}
		\includegraphics[width=\linewidth]{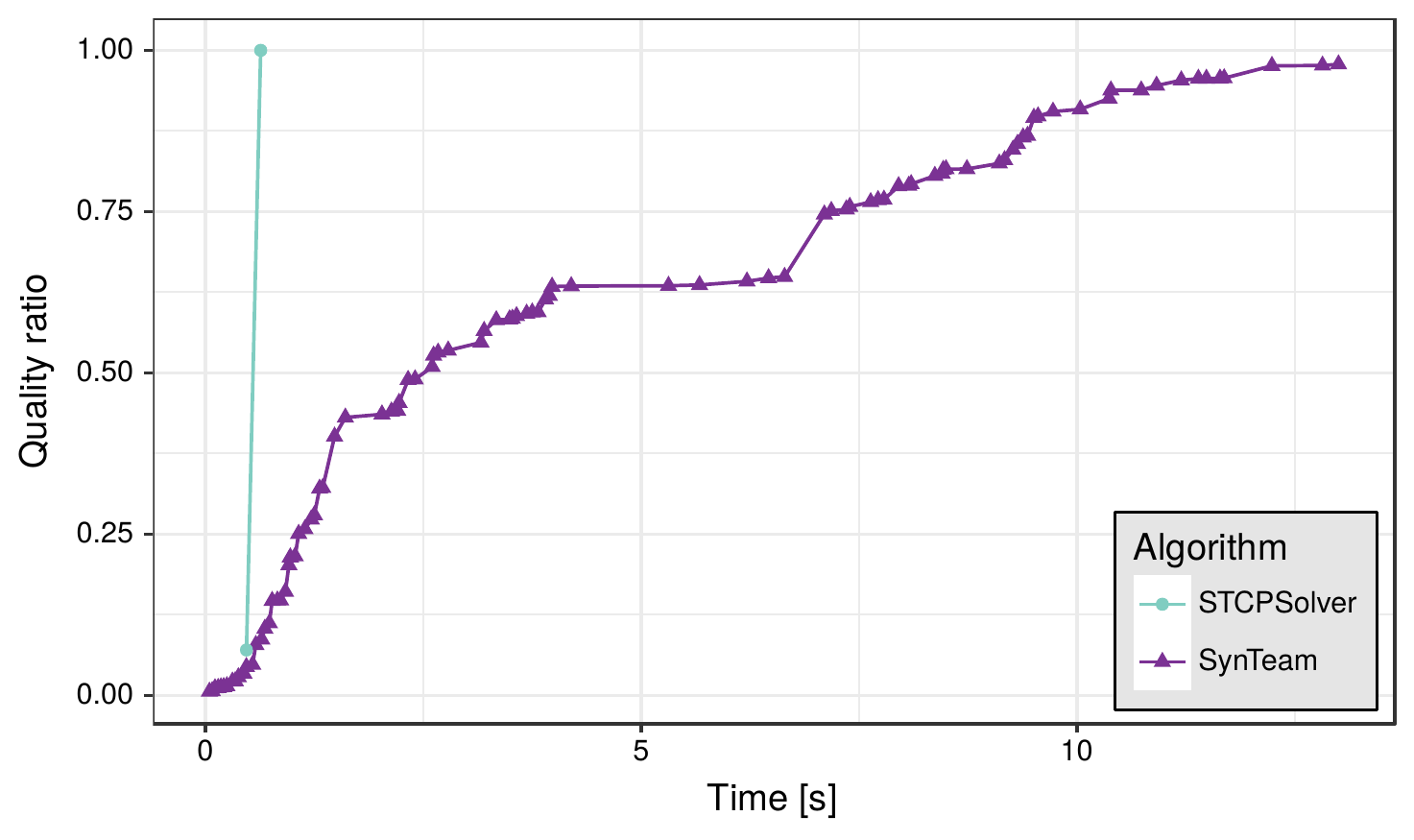}
		\caption{\texttt{arts design}, $\lambda = 0.2$.}
		\label{figure:qTime:arts_design:lambda02}
	\end{subfigure}
	\begin{subfigure}{0.49\textwidth}
		\includegraphics[width=\linewidth]{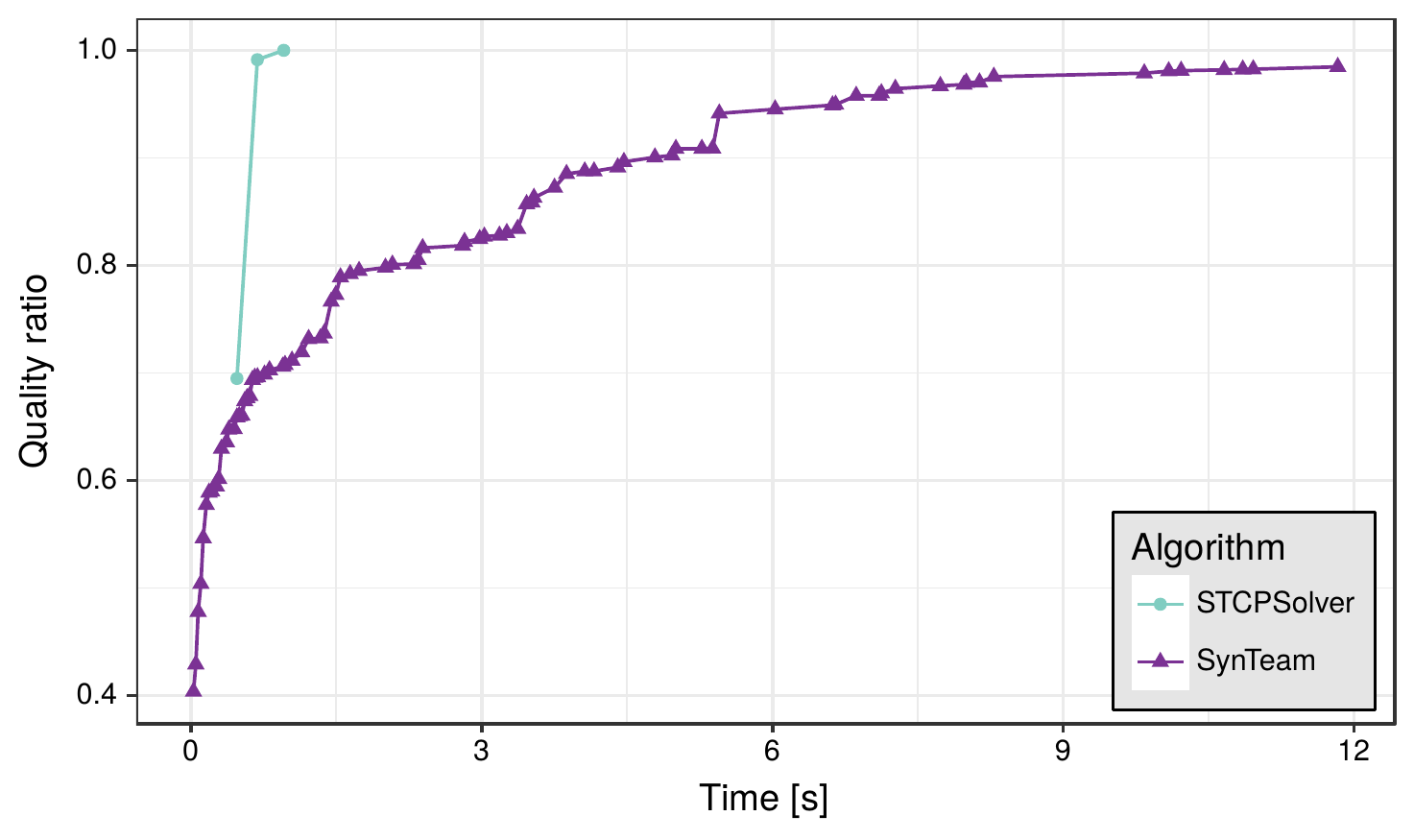}
		\caption{\texttt{arts design}, $\lambda = 0.8$.}
		\label{figure:qTime:arts_design:lambda08}
	\end{subfigure}
	\begin{subfigure}{0.49\textwidth}
		\includegraphics[width=\linewidth]{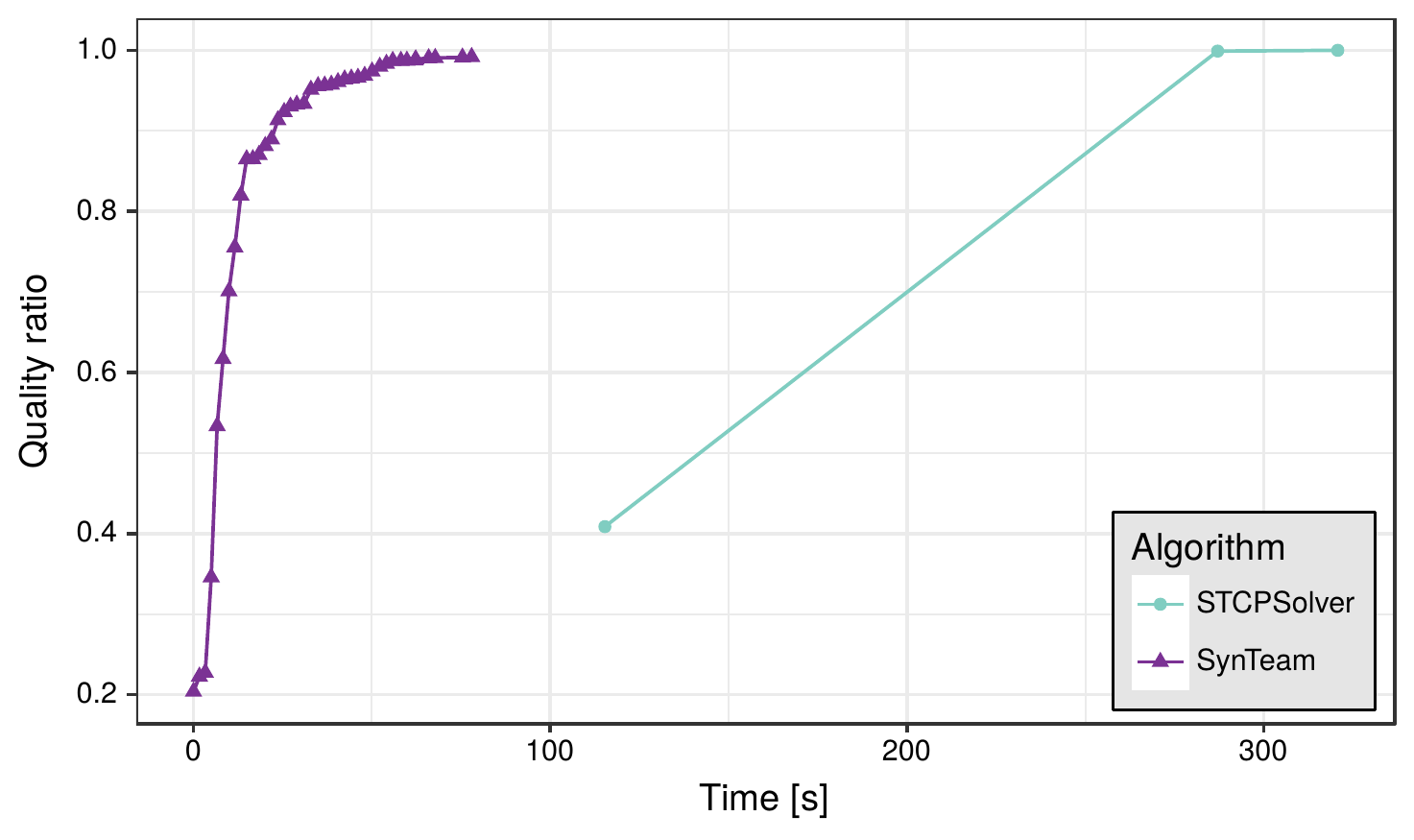}
		\caption{\texttt{English}, $\lambda = 0.2$}
		\label{figure:qTime:english:lambda02}
	\end{subfigure}
	\begin{subfigure}{0.49\textwidth}
		\includegraphics[width=\linewidth]{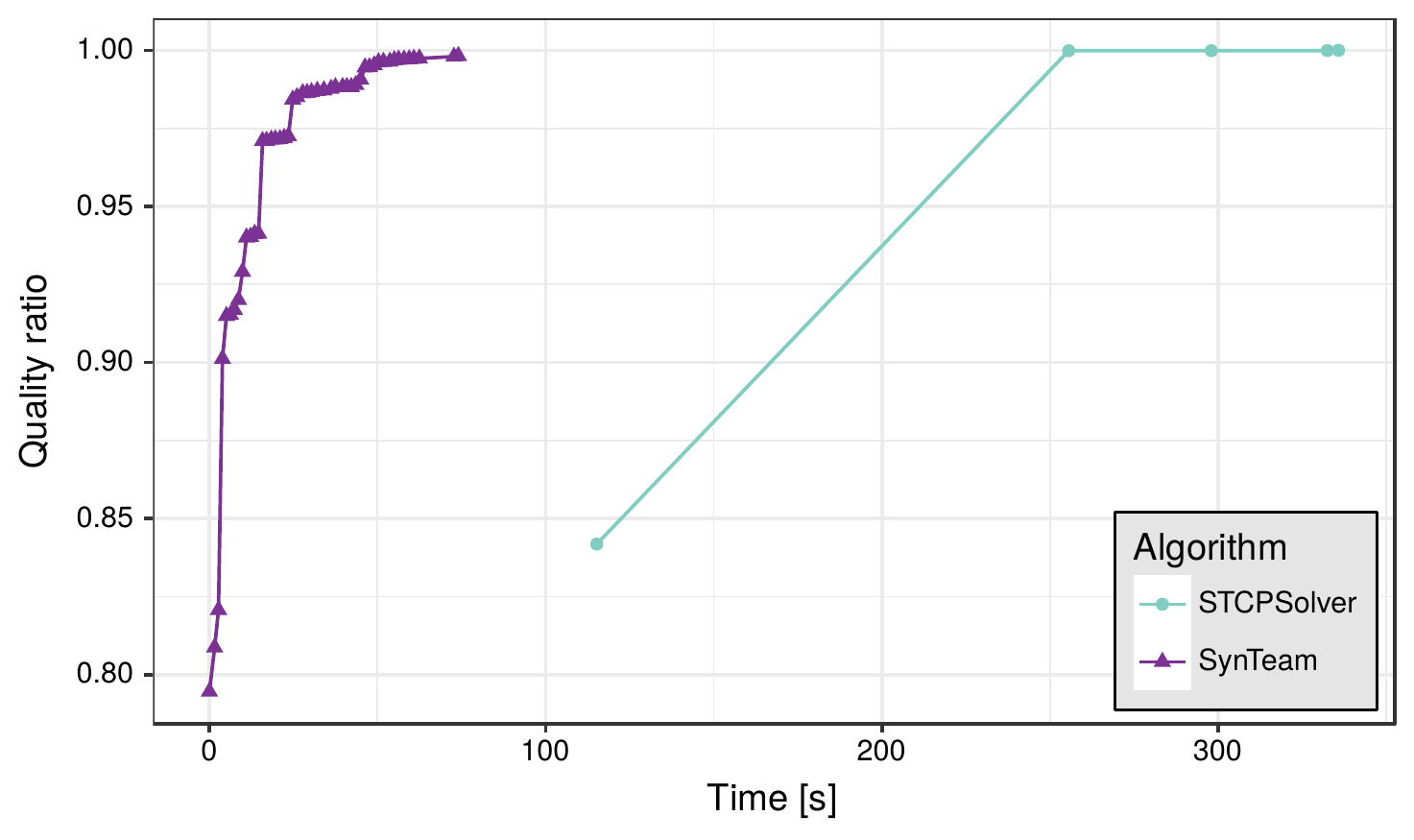}
		\caption{\texttt{English}, $\lambda = 0.8$}
		\label{figure:qTime:english:lambda08}
	\end{subfigure}
    \caption{Anytime performance of \SynTeam\ vs. \cplex. The curves show the quality ratio (with respect to the optimal solutions). Graphics (a) and (b) are for 45 students and a team size of 5. Graphics (c) and (d) are for 80 students and a team size of 4. Graphics (e) and (f) are for 60 students and a team size of 3. Finally, graphics (g) and (h) are for 42 students and a team size of 6.}
    \label{figure:qTime}
\end{figure}

\noindent {\bf Anytime performance.} 
We also decided to show examples of the anytime performance of the two different algorithms. For that purpose, we chose for each task type a different exemplary combination of $n$ and $m$. The corresponding graphics can be seen in Figure~\ref{figure:qTime}. Note that the considered combinations of $n$ and $m$ are detailed in the figure caption. Moreover, they do not include the data generation time of \cplex. 

Figure~\ref{figure:qTime} clearly shows that the anytime performance of \SynTeam\ is superior for team sizes with $m > 3$. In Figure~\ref{figure:qTime:body_rythm:lambda02}, for example, observe that \SynTeam\ provides very good solutions after approximately 11 seconds, while \cplex\ needs approximately $21$ seconds (in addition to more than 1000 seconds of data generation time) to come up with a first, low-quality solution. However, when team sizes are small (as in the case of Figure~\ref{figure:qTime:arts_design:lambda02} and~\ref{figure:qTime:arts_design:lambda08}), the anytime performance of \cplex\ is better. \\

\begin{figure}[t!] 
\centering
	\begin{subfigure}{0.49\textwidth}
		\includegraphics[width=\linewidth]{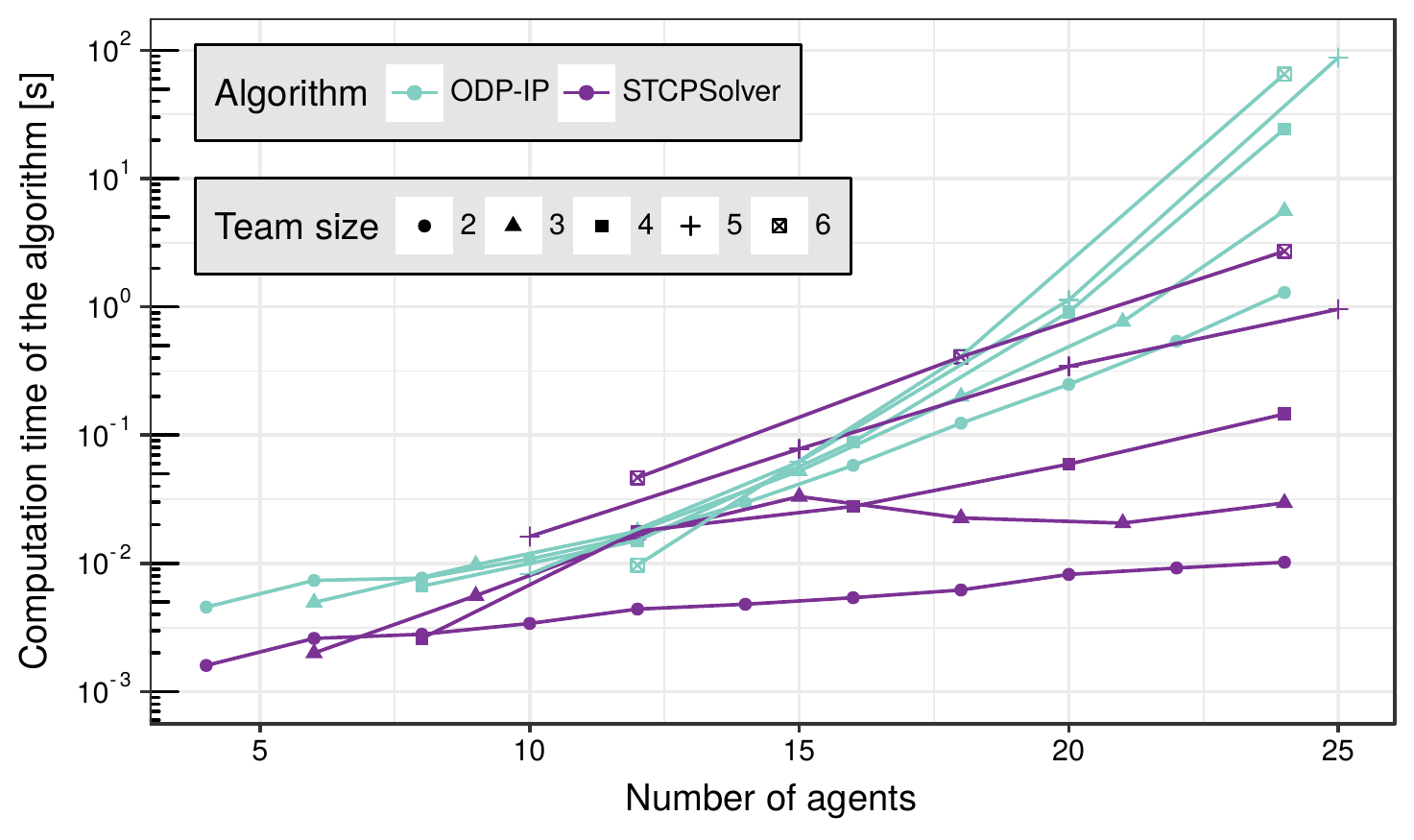}
		\caption{\texttt{arts design}, $\lambda = 0.2$.}
		\label{figure:odpip-comparison:arts_design:lambda02}
	\end{subfigure}
	\begin{subfigure}{0.49\textwidth}
		\includegraphics[width=\linewidth]{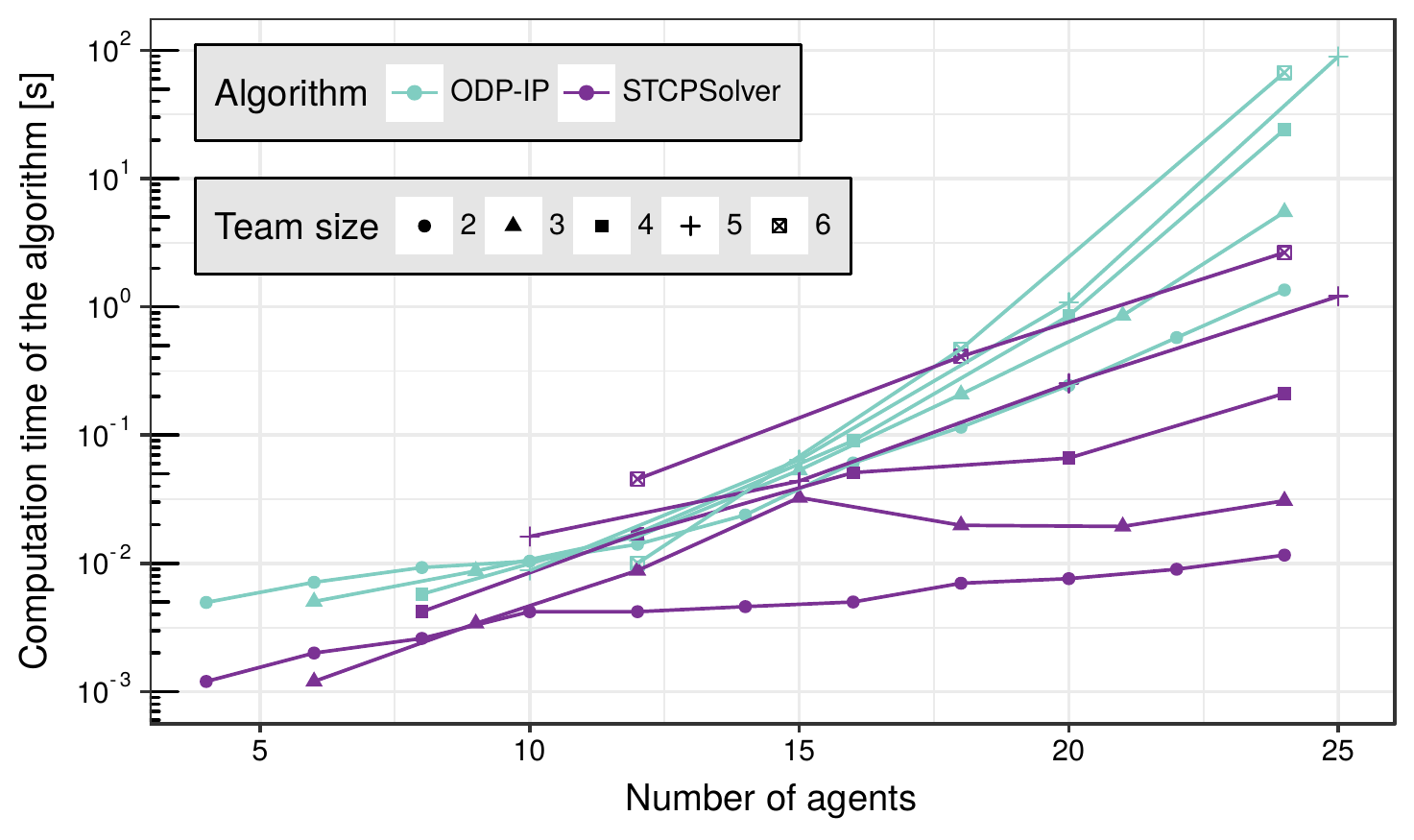}
		\caption{\texttt{arts design}, $\lambda = 0.8$.}
		\label{figure:odpip-comparison:arts_design:lambda08}
	\end{subfigure}
    \caption{Comparison of the computation times (disregarding the data gecmq neration time) between \cplex\ and ODP-IP in the context of task \texttt{arts design}. Note that ODP-IP is limited to a total number of 25 students.}
    \label{figure:odpip-comparison}
\end{figure}

\noindent {\bf Comparison to ODP-IP.} 
Furthermore, we compared \SynTeam{} to ODP-IP~\cite{michalak2016hybrid},
the state-of-the-art coalition structure generation algorithm that can be employed to solve the STCP, as discussed in Section \ref{sec:cfrel}. We employed it to solve all cases in which this was possible, that is, to all cases with $n \leq 25$.\footnote{Note that, due to exponential memory requirements, the ODP-IP algorithm can not be applied to cases with more than 25 students.} Figure~\ref{figure:odpip-comparison} shows a comparison of the average computation times of \cplex\ and ODP-IP, exemplary for task \texttt{arts design}. The main observation is that, even though ODP-IP seems to be slightly faster than \cplex\ for team sizes greater than 3 when $n$ is small, with a growing value of $n$ the computation time requirements of ODP-IP increase much more raplidly than those of \cplex. In fact, for the cases with $n \geq 24$, the computation time of ODP-IP is about two orders of magnitude higher than the one of \cplex. Moreover, the data generation time for ODP-IP, which is not taken into account in these graphics, is longer than for ODP-IP, due to the fact that \emph{every} subset of students (even unfeasible teams of size different from $m$) must be visited in the generation phase (see Section \ref{sec:cfrel}). \\

\textcolor{black}{
\noindent {\bf Comparison to a baseline metaheuristic: Simulated Annealing.} 
Finally, we compared \SynTeam{} to a standard implementation of simulated annealing (SA)~\cite{kirkpatrick1983optimization}, which is one of the baseline metaheuristics. The algorithm starts from a random initial solution. At each step, a random neighbour of the current solution is chosen by (1) randomly selecting two agents from two different teams, and by (2) swapping the two agents. This neighbouring solution is accepted as the new current solution (1) if it is at least as good as the current solution, or (2) with probability  
\begin{equation}
P_{accept}=\exp
\left(-\frac{\Delta }{T}\right)=\exp \left(-\frac{(S(P',\tau)-S(P,\tau))/S(P',\tau) }{T}\right),
\end{equation} 
where $(P,\tau)$ denotes the current solution, $(P',\tau)$ denotes the neighbouring solution, $\Delta$ is the percent change in synergistic value, and $T$ is an important control parameter called \emph{temperature}. At the beginning of the search process, the value of $T$ is relatively large so that even worsening moves are frequently accepted. During the optimization process, the temperature is gradually decreased so that fewer and fewer worsening moves are accepted.
}

\textcolor{black}{
We adopt the scheme for setting the value of $T$ at each iteration proposed by Samples et al.~\cite{SamBluMasRos02:ppsn}. The initial temperature is set such that the probability to accept a move with $\Delta=\delta =0.01$ is $P_{start}=0.9$. Moreover, at the end of the optimization process, we aim for a probability to accept a move with $\Delta =\delta =0.01$ of $P_{end}=0.1$. With these requirements, we define the temperature value at time $x$ to be $T:=r^{x}\tau _{\max }$, where $\tau _{\max }:=-\delta /\ln P_{start}$, $r:=\sqrt[t_{\max }]{\delta /(\ln (1/P_{end})\cdot {\tau_{\max}} )}$, and where $t_{\max }$ is the computation time limit of the run.
}

\begin{figure}[t!] 
\centering
	\begin{subfigure}{0.49\textwidth}
		\includegraphics[width=\linewidth]{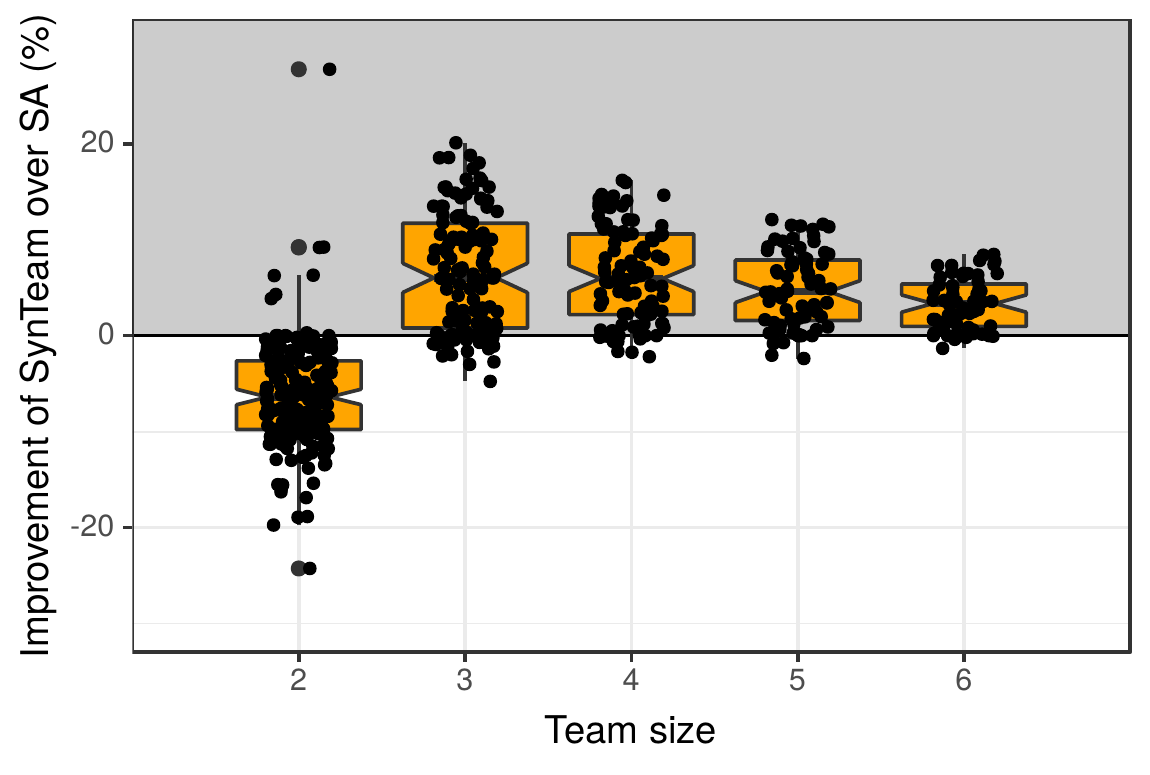}
		\caption{$\lambda = 0.2$.}
		\label{figure:sa-comparison:lambda02}
	\end{subfigure}
	\begin{subfigure}{0.49\textwidth}
		\includegraphics[width=\linewidth]{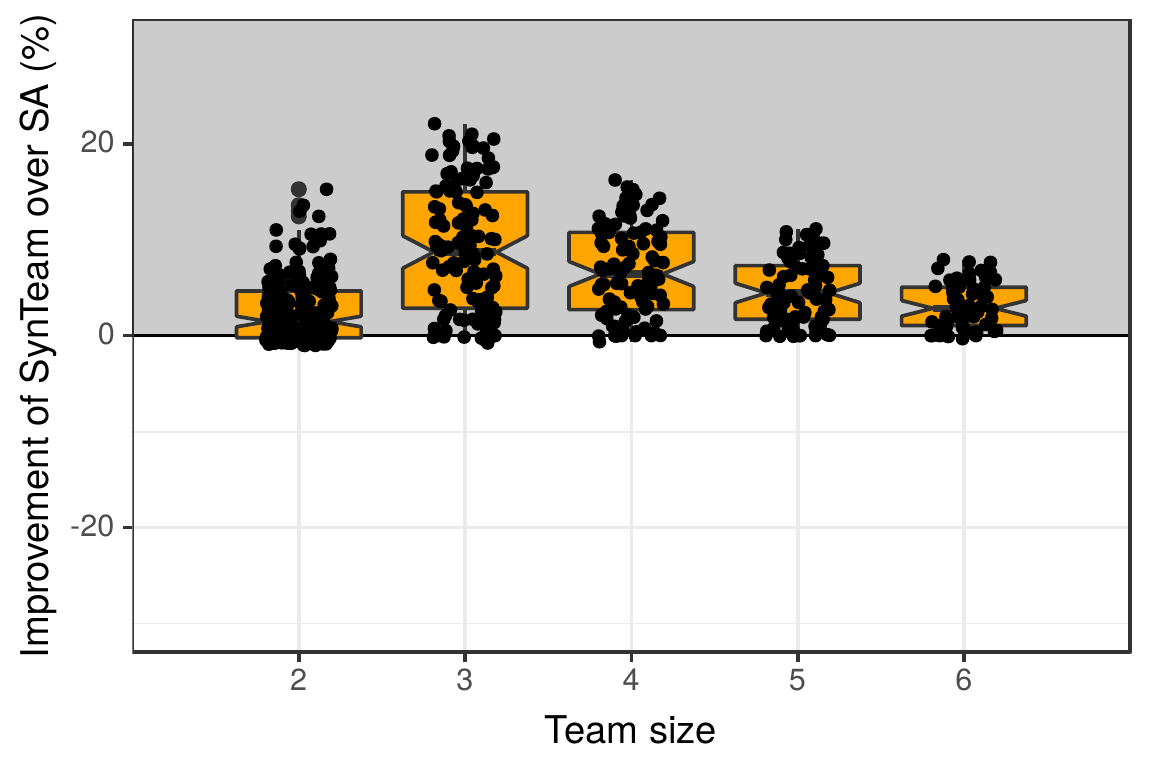}
		\caption{$\lambda = 0.8$.}
		\label{figure:sa-comparison:lambda08}
	\end{subfigure}
    \caption{\textcolor{black}{
The boxplots show the improvement (in percent) of \SynTeam{} over SA. Each box contains all comparisons concerning all different considered numbers of students ($n$) and all four tasks. Note that points in the area below zero (y-axis) indiacate that SA performed better than \SynTeam.}}
    \label{figure:sa-comparison}
\end{figure}

\textcolor{black}{
SA was applied to all problems to which \SynTeam{} was applied. Moreover, the computation time limit given to SA was the total time required by \SynTeam{} in each case. The results of the comparison between \SynTeam{} and SA are presented in Figure~\ref{figure:sa-comparison:lambda02} (for $\lambda=0.2$) and Figure~\ref{figure:sa-comparison:lambda08} (for $\lambda=0.8$) in terms of boxplots. The data points show, for each team size, the improvement of \SynTeam{} over SA (note that these graphics summarize over all numbers of students ($n$) and all four tasks). Hereby, values below zero (area with white background) indicate that SA performed better than \SynTeam{}. The  following conclusions can be drawn: first, there is just one single case ($\lambda=0.2$, team size 2) in which SA is generally better than \SynTeam{}. In all other cases, \SynTeam{} outperforms SA with statistical significance (as indicated by the notches of the boxplots). Interestingly, the relative performance of \SynTeam{} with respect to SA is better with a higher value of $\lambda$. \\
}
\textcolor{black}{
Summarizing, we would like to remark that we were not able to observe significant differences in (relative) algorithm performance when comparing between the different task types. This is a strong indication for the robustness of the developed techniques. Moreover, we obtained a confirmation of the well known principle that \emph{no algorithm is the best-performing one in all possible cases}. More specifically, when team sizes of $m \leq 3$ are concerned, we recommend the use of \cplex, while when $m > 3$ the use of \SynTeam\ is indicated.}


\section{Eduteams: a publicly-available application to compose teams}
\label{sec:eduteams}


\textcolor{black}{
We developed a web application, EduTeams, which is publicly available.\footnote{\url{http://eduteams.iiia.csic.es}} }

\begin{figure}
\centering
\includegraphics[max size={10cm}]{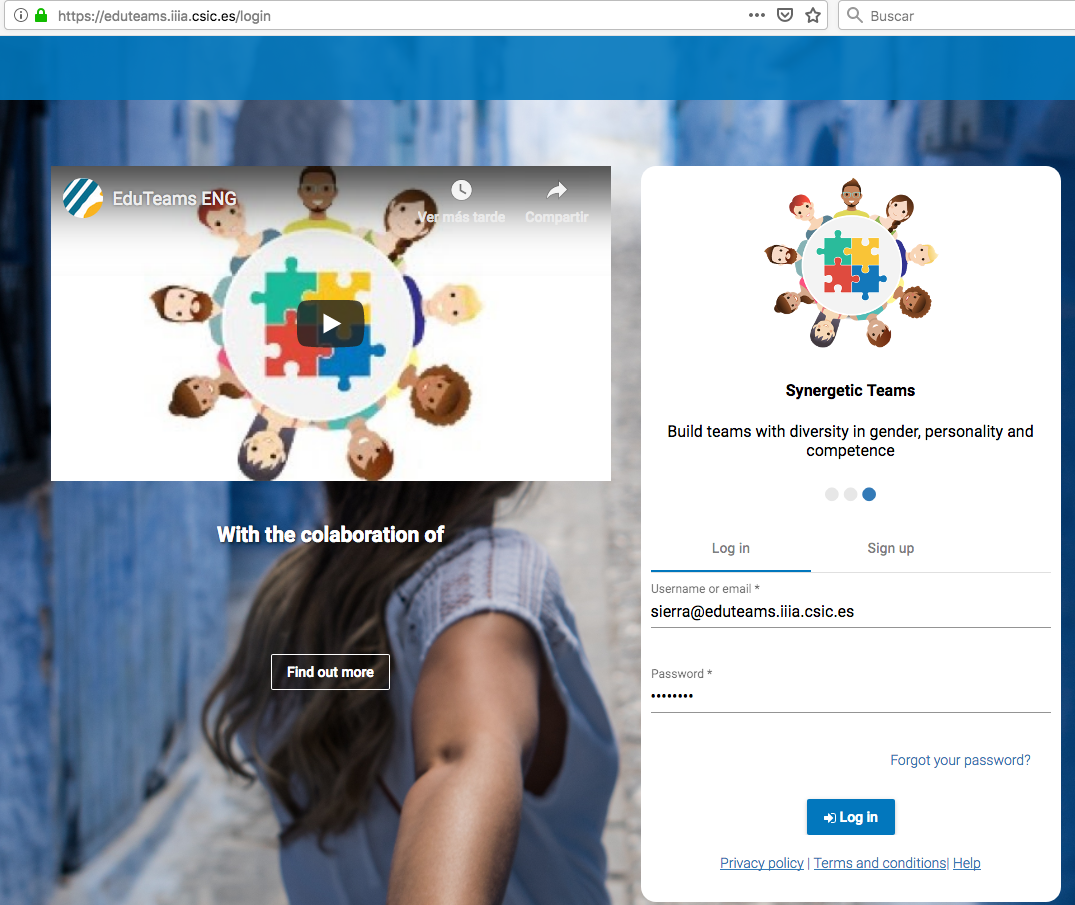}
\caption{EduTeams user registration page.}\label{edu1}
\end{figure}

\begin{figure}
\centering
\includegraphics[max size={10cm}]{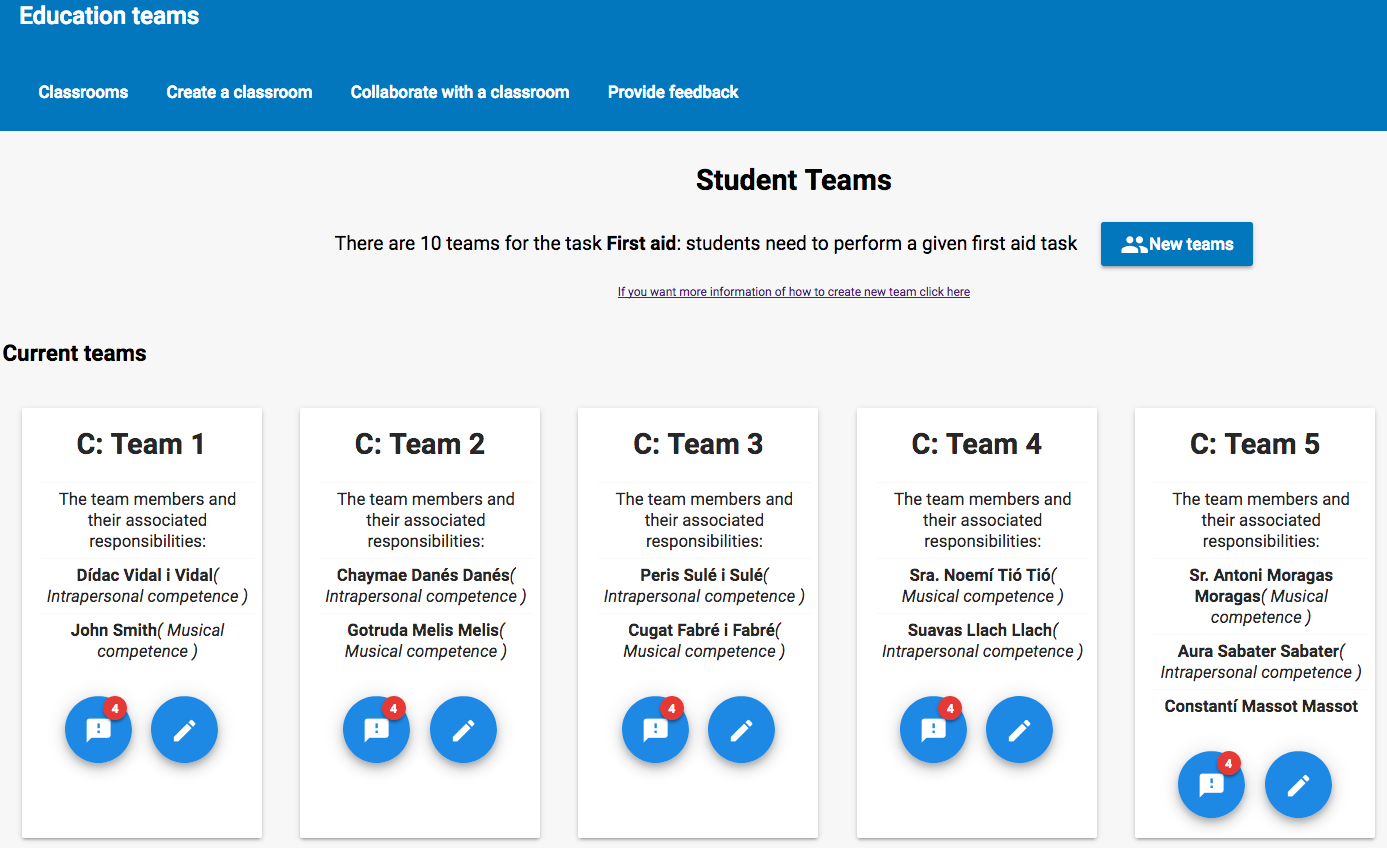}
\caption{Teams as shown in EduTeams.}\label{edu2}
\end{figure}

\textcolor{black}{
EduTeams allows teachers to employ \SynTeam\ to partition their classrooms into synergistic teams to perform collaborative tasks  requiring multiple competences. 
In short, after signing up in EduTeams, a teacher can create her own classrooms.
For a classroom, the teacher can specify tasks using the competences, competence requirements and importance levels specified in section \ref{subsec:empiricalSettings}. Several examples of task definitions are shown in Table \ref{tab:task-types}. Every classroom has a unique code that students need to set up their own accounts and thus join classrooms. After signing up, each student is asked to fill in one competence and one personality test. Once all students in a classroom complete both tests, the teacher owning the classroom can proceed with team composition. Figures \ref{edu1} and \ref{edu2} display a couple of snapshots of EduTeams using artificial data so that no privacy is breached.
}

\section{Conclusions and future work}\label{sec:discuss}

In this paper, we defined the Synergistic Team Composition Problem (STCP) in the domain of student team composition. We introduced two different solutions to solve our problem, that is, to partition students' classrooms into teams that are balanced in size, competences, personality and gender. First, we proposed an algorithm called \cplex\ to optimally solve our problem. Second, we proposed an algorithm called \SynTeam, a heuristic that yields close to optimum, but not necessarily optimal solutions. Our computational results show that the benefits of \SynTeam\ with respect to \cplex\ grow with the increasing number of students and team sizes. 
Moreover, \SynTeam\ gives good quality approximate solutions depending on the trade-off between proficiency and congeniality. 
Thus, even in the case of the smallest considered team size ($m = 2$), the quality ratio of SynTeam
is always above 95\% when preferring proficiency over congeniality ($\lambda = 0.8$), and above
75\% when preferring congeniality over proficiency ($\lambda = 0.2$).

This paper identified and formalised an interesting real-world case as a new type of constrained coalition formation problem. This case requires a \emph{balanced} coalition structure in terms of both coalitional values and coalition sizes. The computational comparison of both algorithms offers the guidance for their use by any institution that is in need for automatic team composition (e.g. classrooms, research units, private companies). Note that the algorithms compose teams in a completely automated way without experts knowledge, which is a tremendous advantage for settings where there are no guidelines or expertise available.


The STCP problem opens new research paths. 
First, there is the need for considering more general and richer models to better express the different determinants of the team performance.
For instance, we plan to extend our approach so as to consider preferences and constraints coming from human experts (e.g., conflicts of interests).
Furthermore, we plan to extend our STCP model to deal with multiple task types, along the lines of the work by Pr{\"a}ntare and Heintz~\cite{10.1007/978-3-030-03098-8_10}.
Finally, we aim at investigating how to exploit paralellism so that our search process can benefit from the capabilities offered by modern computer architectures.


\section*{Acknowledgements}
\noindent
This work was supported by projects CIMBVAL (MINECO, TIN2017-89758-R), AppPhil (RecerCaixa 2017), 2017 SGR 172, Collectiveware (MINECO/FEDER, TIN2015-66863-C2-1-R) and AI4EU (H2020-825619). Bistaffa was supported by the H2020-MSCA-IF-2016 HPA4CF project. The research was partially supported by the ST Engineering - NTU corporate Lab through the NRF corporate lab@university scheme. The first author thanks Enzyme Advising Group for partial sponsorship of the Industrial PhD programme to develop ideas introduced in this paper.

\section*{References}
\bibliography{mybibfile}

\end{document}